\documentclass[times,twocolumn,final]{elsarticle}

\usepackage{medima}
\usepackage{framed,multirow}
\usepackage{multicol}
\usepackage{makecell}

\usepackage{amssymb}
\usepackage{latexsym}

\usepackage{url}
\usepackage[table]{xcolor}

\usepackage{subfigure}
\usepackage{bm}
\usepackage{amsmath,amsfonts}
\usepackage{algorithmic}
\usepackage{graphicx}
\usepackage{textcomp}
\usepackage{mathrsfs}
\usepackage{pifont}
\usepackage{stfloats}
\usepackage{booktabs}

\usepackage{tabularx}

\usepackage[colorlinks,
linkcolor=blue,
anchorcolor=blue,
citecolor=blue,
urlcolor=blue]{hyperref}

\definecolor{newcolor}{rgb}{.8,.349,.1}

\def\blue{\textcolor{blue}}
\def\red{\textcolor{red}}

\journal{Medical Image Analysis}

\begin{document}

\verso{Yu Cai \textit{et~al.}}

\begin{frontmatter}

\title{MedIAnomaly: A comparative study of anomaly detection in medical images}

\author[1]{Yu \snm{Cai}}
\author[2]{Weiwen \snm{Zhang}}
\author[2,3,4]{Hao \snm{Chen}\corref{cor1}}
\cortext[cor1]{Corresponding author.}
\ead{jhc@cse.ust.hk}
\author[1,2]{Kwang-Ting \snm{Cheng}}

\address[1]{Department of Electronic and Computer Engineering, The Hong Kong University of Science and Technology, Hong Kong, China}
\address[2]{Department of Computer Science and Engineering, The Hong Kong University of Science and Technology, Hong Kong, China}
\address[3]{Department of Chemical and Biological Engineering, The Hong Kong University of Science and Technology, Hong Kong, China}
\address[4]{Division of Life Science, The Hong Kong University of Science and Technology, Hong Kong, China}

\received{5 April 2024}
\finalform{5 February 2025}
\accepted{6 February 2025}
\availableonline{17 February 2025}

\begin{abstract}
Anomaly detection (AD) aims at detecting abnormal samples that deviate from the expected normal patterns. Generally, it can be trained merely on normal data, without a requirement for abnormal samples, and thereby plays an important role in rare disease recognition and health screening in the medical domain. Despite the emergence of numerous methods for medical AD, the lack of a fair and comprehensive evaluation causes ambiguous conclusions and hinders the development of this field. To address this problem, this paper builds a benchmark with unified comparison. Seven medical datasets with five image modalities, including chest X-rays, brain MRIs, retinal fundus images, dermatoscopic images, and histopathology images, are curated for extensive evaluation. Thirty typical AD methods, including reconstruction and self-supervised learning-based methods, are involved in comparison of image-level anomaly classification and pixel-level anomaly segmentation. Furthermore, for the first time, we systematically investigate the effect of key components in existing methods, revealing unresolved challenges and potential future directions. The datasets and code are available at \url{https://github.com/caiyu6666/MedIAnomaly}.
\end{abstract}


\end{frontmatter}


\section{Introduction}  \label{sec:introduction}

Anomaly detection (AD) is a fundamental machine learning problem~\citep{chandola2009anomaly,pang2021deep}, which aims to detect abnormal samples that deviate from expected normal patterns. It works under the assumption that a large number of normal samples with similar patterns are readily available, while abnormal samples, with diverse and unknown patterns, are difficult to collect comprehensively. Therefore, existing AD methods generally train the model merely on normal data. This alleviates the requirement for abnormal samples, facilitating the application of AD in the medical domain, especially in rare disease recognition and health screening where abnormal samples are usually rare and diverse. Take the case of the United States, where over 7,000 rare diseases affect more than 30 million people, though each disease alone affects fewer than 200,000 people (about 0.06\% of the population).\footnote{\url{https://www.fda.gov/patients/rare-diseases-fda}} Their rarity and diversity make it infeasible to collect enough samples for each disease to train a typical supervised model. This fact makes AD an appropriate solution to assist radiologists in recognizing potential abnormalities from medical images.

\begin{table}[ht]
\centering
\caption{Performance (AUC) of our implemented AE and prevailing works on Hyper-Kvasir and OCT2017 datasets. The details of AE (Our impl.) are shown in Section~\ref{sec:impl_rec}. Results$^*$ are taken from \citet{tian2021constrained,tian2023self}, results$^\dag$ are taken from \citet{zhao2022ae}, and results$^\ddag$ are taken from \citet{zhao2021anomaly}.}
\label{tab:indistinguishable_data}


\resizebox{\linewidth}{!}{
\begin{tabular}{lllll}
\toprule
Method & Hyper-Kvasir  & OCT2017   \\ \midrule
AE                                  & -             & $83.02^\dag$   \\
f-AnoGAN \citep{schlegl2019f}       & $90.7^*$      & $85.09^\dag$   \\
GANomaly \citep{akcay2019ganomaly}  & -             & $90.52^\dag$   \\
OCGAN \citep{perera2019ocgan}       & $81.3^*$      & -              \\
IGD \citep{chen2022deep}            & $93.9^*$      & -              \\
CCD-IGD \citep{tian2021constrained} & $97.2^*$      & -              \\
DifferNet \citep{rudolph2021same}   & -             & $94.27^\dag$   \\
Fastflow \citep{yu2021fastflow}     & -             & $94.08^\dag$   \\
AE-FLOW \citep{zhao2022ae}          & -             & $98.15^\dag$   \\
SALAD \citep{zhao2021anomaly}       & -             & $96.42^\ddag$  \\ 
PMSACL-PaDiM \citep{tian2023self}   & $99.6^*$      & -              \\ \midrule
\rowcolor{green!20} AE (Our impl.)  & $99.5_{\pm 0.0}$  & $96.4_{\pm 0.4}$  \\ \bottomrule
\end{tabular}
}

\end{table}

Numerous studies \citep{schlegl2017unsupervised,schlegl2019f,mao2020abnormality,tan2021detecting,schluter2022natural,cai2022dual} have been dedicated to medical AD. However, the lack of a comprehensive and fair comparison makes some conclusions unclear and hinders the development of this area. 
On the one hand, the use of different datasets or partitions in existing studies impedes reproducibility and comparability. As a result, some studies underestimate the baseline and draw biased conclusions.
For instance, several papers \citep{tian2021constrained,tian2023self,zhao2021anomaly,zhao2022ae} performed experiments on datasets such as Hyper-Kvasir \citep{borgli2020hyperkvasir} or OCT2017 \citep{kermany2018identifying}. However, as shown in Table~\ref{tab:indistinguishable_data}, a simple auto-encoder (AE) implemented by us, with default settings described in Section~\ref{sec:impl_rec}, achieves nearly perfect performance on these datasets. This suggests that these datasets are not challenging for AD, and therefore evaluating AD methods on them is not suitable.
On the other hand, despite many methods belonging to the same paradigm, the absence of a unified implementation leads to different network architectures and training tricks, resulting in unfair comparison. Take the case of reconstruction-based methods, where f-AnoGAN \citep{schlegl2019f} adopts residual blocks \citep{he2016deep} with multiple convolutional layers for the encoder and decoder, AE-Flow \citep{zhao2022ae} chooses Wide ResNet-50-2 \citep{zagoruyko2016wide} as the encoder, while AE-U \citep{mao2020abnormality} uses plain convolutional layers without any specific designs. These differences could interfere with the comparison of methodologies and make ambiguous results. Overall, the comparisons based on improper datasets or inconsistent implementations are unreasonable. To address these issues, comprehensive datasets and a unified implementation of methodologies for medical AD are in high demand.

Although some surveys and benchmarks have been made for medical AD \citep{fernando2021deep,baur2021autoencoders,lagogiannis2023unsupervised,bao2023bmad,cai2023dual}, they do not provide a comprehensive and fair comparison of methods. \citet{fernando2021deep} provided a systematic review of machine learning–based medical AD techniques with their applications and limitations, while no experiments were conducted. \citet{baur2021autoencoders} compared a series of medical reconstruction-based AD methods on unified datasets. However, the comparison was limited to reconstruction methods only and the experiments were performed on a single image modality (brain MRI). These could lead to incomplete conclusions. Also, their utilization of private datasets impedes reproducibility. 
\citet{cai2023dual} assessed various medical AD methods on reorganized datasets, but did not dive into a detailed analysis of these methods. \citet{bao2023bmad} evaluated a range of AD methods on reorganized public medical datasets. However, most of these methods were originally designed for industrial applications rather than the medical domain.

It is important to note that \citet{lagogiannis2023unsupervised} recently offered a detailed analysis of diverse SOTA AD methods across multiple datasets. They extensively investigated the influence of anomaly size and intensity, the effects of limited training data, and the performance of self-supervised pre-training on weight initialization and backbone pre-training. 
However, their analysis did not thoroughly encompass some non-latest yet representative methods, such as various variants of AE-based methods and certain self-supervised methods utilizing specialized synthetic anomalies. Additionally, they did not employ unified network architectures for analyzing methods within the same paradigm. As a result, the inherent properties of the components and network architectures in typical methods remain unexplored.
Unlike \citet{lagogiannis2023unsupervised}, we aim to further investigate the effects and inherent properties of key components in typical methods by conducting comparisons across comprehensive datasets and employing as fair as possible network configurations.
The main contribution of this work is summarized as follows:
\begin{itemize}
    \item We present a taxonomy of common anomaly detection methods, encompassing reconstruction-based methods (image-reconstruction and feature-reconstruction), self-supervised learning-based methods (one-stage and two-stage), and feature reference-based methods (knowledge distillation and feature modeling). A thorough literature review is conducted accordingly.
    \item We collect seven medical datasets, comprising five image modalities, to facilitate the evaluation of anomaly detection.
    \item We conduct a comprehensive comparison of thirty representative methods using the collected datasets. Furthermore, the effects of key components employed in these methods are analyzed. 
    \item Through experiments and analysis, we highlight unresolved challenges and potential future directions in the field of medical anomaly detection.
\end{itemize}
\section{Literature review}  \label{sec:literature}
AD is mostly formulated as one-class classification (OCC) \citep{ruff2018deep}. Given a normal training set $\mathcal{D}_{train}=\{\mathbf{x}_i\}_{i=1}^{N}$ containing $N$ normal samples, the objective is to learn an anomaly scoring function $\mathcal{A}({~\cdot~; \boldsymbol{\theta}})$ parameterized by $\boldsymbol{\theta}$ that assigns a score to an input sample. The score should satisfy the condition $\mathcal{A}(\mathbf{x}_n) < \mathcal{A}(\mathbf{x}_a)$ for any normal sample $\mathbf{x}_n$ and abnormal sample $\mathbf{x}_a$. In the context of medical image analysis, the anomaly score can be assigned to both images and pixels, enabling image-level anomaly classification (AnoCls) and pixel-level anomaly segmentation (AnoSeg). 

In this section, we provide a systematic review of typical methods for AD, categorized into reconstruction-, self-supervised learning-, and feature reference-based methods.

\subsection{Reconstruction-based anomaly detection}

\begin{figure*}
\centering
\includegraphics[width=0.65\linewidth]{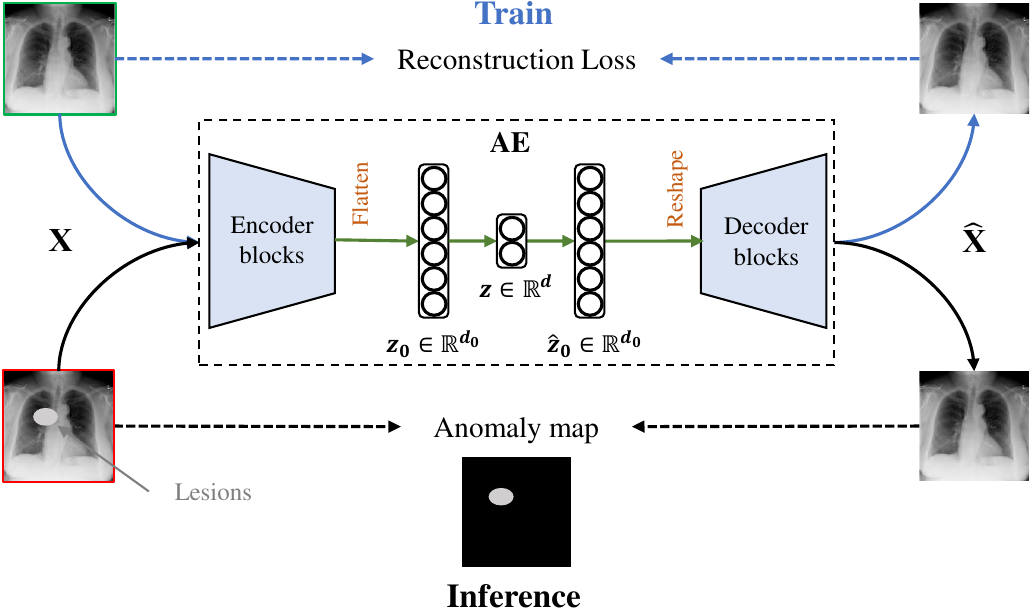}
\caption{Overview of reconstruction-based anomaly detection. The reconstruction model is trained to minimize reconstruction loss on normal images. During inference, lesions in abnormal images are assumed unable to be reconstructed by the trained model, and in turn yield a high reconstruction error.}
\label{fig:overview_rec}
\end{figure*}

Reconstruction-based methods are a dominant strategy for medical AD \citep{baur2021autoencoders}. These methods typically rely on generative models such as generative adversarial networks (GANs) \citep{goodfellow2014generative}, auto-encoders (AEs) or their variants. As depicted in Fig.~\ref{fig:overview_rec}, they train a reconstruction model to reconstruct only normal images, and then utilize the reconstruction error as an anomaly score during inference. The underlying assumption is that a model trained solely on normal images will struggle to accurately reconstruct unseen abnormal regions, and in turn yield a high reconstruction error on these regions. We categorize existing reconstruction-based methods into two groups: image-reconstruction and feature-reconstruction approaches.

\subsubsection{Image-reconstruction}
Image-reconstruction approaches directly perform reconstruction and anomaly score calculation in the image space. Typically, \citet{schlegl2017unsupervised,schlegl2019f} trained GANs to learn the manifold of normal images and computed reconstruction errors from both reconstructed images and intermediate discriminator features. However, GAN-based models are intractable to train and suffer from memorization pitfalls. 

Numerous works \citep{baur2021autoencoders,bergmann2018improving,zimmerer2019unsupervised,chen2018unsupervised,marimont2021anomaly} have explored vanilla AE and its variants, including Variational AE (VAE), \citep{kingma2013auto}, Adversarial AE (AAE) \citep{makhzani2015adversarial}, and Vector Quantized VAE (VQ-VAE) \citep{van2017neural}. While these methods have a stable training process, the reconstruction quality is often suboptimal. To mitigate this issue, several works \citep{baur2019deep,akcay2019ganomaly,bercea2023generalizing} have introduced adversarial training to AE-based models to enhance the reconstruction quality. Additionally, various studies have explored the utilization of different distance functions for reconstruction, such as Structural Similarity Index Measure (SSIM) \citep{bergmann2018improving,behrendt2024diffusion} and perceptual loss (PL) \citep{shvetsova2021anomaly,bercea2023generalizing}. These approaches aim to enhance the reconstruction quality by improving image-space fidelity and perceptual consistency.

However, the assumption of reconstruction-based AD that the model cannot reconstruct unseen abnormal regions does not always hold. This issue is still unresolved even though some efforts have been made. We discuss this point from the perspective of two distinct problems as follows.

The first problem is that models can reconstruct some unseen abnormal regions due to their generalization ability, resulting in false negatives. To tackle this problem, \citet{gong2019memorizing} proposed to record prototypes of normal training patterns, which serve as input of the decoder to facilitate anomaly-free reconstruction. \citet{zimmerer2018context} introduced an inpainting task \citep{pathak2016context} which could enable the reconstruction model to repair abnormal regions. \citet{kascenas2022denoising,kascenas2023role} explored denoising tasks with different noises for training reconstruction AD models, demonstrating the superiority of their proposed coarse noise compared to Gaussian and Simplex noises \citep{Wyatt_2022_CVPR}.

The second problem is that intrinsic reconstruction errors on normal regions are inevitable due to the finite network capacity, resulting in false positives. \citet{meissen2022pitfalls} demonstrated that these intrinsic errors could overshadow the reconstruction errors on abnormal regions that have moderate pixel intensity, even if the reconstruction model successfully repairs these abnormal regions. \citet{mao2020abnormality} observed that high error often appears around high-frequency regions like boundaries. To suppress these errors, they proposed to automatically estimate the pixel-level uncertainty of reconstruction using AE, which is then used to normalize the reconstruction error. Recent studies \citep{bercea2023aes,liu2023diversity} have explored the use of deformation fields \citep{balakrishnan2019voxelmorph} to address this issue. \citet{bercea2023aes} proposed to refine the reconstruction using estimated dense deformation fields. \citet{liu2023diversity} introduced deformation as a measure of reconstruction diversity, complementing the original reconstruction error.

Moreover, the emergence of DDPMs \citep{ho2020denoising} has introduced a new pipeline for reconstruction-based AD. This pipeline includes a partial diffusion process applied to input images to corrupt abnormal regions, followed by a denoising process to reconstruct the corresponding healthy approximation.
\citet{Wyatt_2022_CVPR} observed that the original Gaussian noise in DDPMs cannot effectively corrupt abnormal regions and proposed to use Simplex noise as an alternative. \citet{bercea2023mask} highlighted the limitation of DDPMs in controlling noise granularity and proposed a solution by integrating automatic masking, stitching, and re-sampling techniques based on DDPMs. To achieve accurate reconstruction of complex brain structures, \citet{behrendt2024patched} proposed to apply the diffusion process to only a patch of the input image at a time, and then apply the denoising process to this result to recover the noised patch.

In addition to the above methods that utilize image reconstruction errors, several studies have investigated the popular visual attention techniques, Class Activation Mapping (CAM) \citep{zhou2016learning} and its gradient-weighted counterpart \citep{selvaraju2017grad}, within the image-reconstruction framework. These techniques are commonly used to highlight important regions in an image and have thus been explored for localizing abnormal regions. For instance, \citet{zimmerer2019unsupervised} introduced this technique into VAE, computing the gradients of reconstruction errors and Kullback-Leibler (KL)-divergence w.r.t. the input image to localize abnormal regions. Several subsequent works \citep{venkataramanan2020attention,liu2020towards,silva2022constrained} have explored similar frameworks.

\subsubsection{Feature-reconstruction}
Different from image-reconstruction approaches, feature-reconstruction approaches reconstruct high-level feature maps instead of image intensity. They adopt a pre-trained deep neural network to map images into the feature space, where the reconstruction is performed. \citet{shi2021unsupervised} and \citet{meissen2022unsupervised} used ImageNet \citep{krizhevsky2012imagenet} pre-trained networks to produce multi-scale feature maps, based on which reconstruction methods are applied. \citet{you2022adtr,you2022unified} argued for the existence of an ``identical shortcut" in reconstruction networks, and proposed a transformer \citep{vaswani2017attention} architecture with a layer-wise query decoder and neighbor masked attention to reconstruct input features. \citet{deng2022anomaly} proposed to aggregate the multi-scale feature maps of the pre-trained encoder into a dense embedding. Taking this embedding as input, the decoder is trained to reconstruct the original multi-scale feature maps. \citet{guo2023recontrast} argued that using frozen encoders pre-trained on natural images is suboptimal due to the semantic gap. They introduced some elements of contrastive learning \citep{chen2021exploring} to jointly optimize the pre-trained encoder for feature reconstruction on the target domain, achieving domain-specific anomaly detection.

The feature-reconstruction approaches are able to achieve superior performance to the image-reconstruction approaches, which can be attributed to the stronger discriminative capability of features in comparison to images. In the feature space, anomalies can deviate more significantly from normal samples, preventing their reconstruction. Consequently, the problem of reconstructing unseen abnormal regions in image-reconstruction approaches is somewhat mitigated by these methods.

Despite their advantages, feature-reconstruction approaches may encounter challenges in accurately localizing small lesions due to the inherent loss of low-level information when utilizing intermediate feature maps. While feature maps with richer semantic information can highlight certain abnormal regions that may appear subtle in the intensity space, they do not fully retain low-level details such as the shapes and boundaries of abnormal regions. As a compromise, existing methods often aggregate feature maps from multiple intermediate layers \citep{shi2021unsupervised,meissen2022unsupervised,you2022unified,deng2022anomaly}. However, there is still a bias in the selection of the number and depth of these layers, which tends to capture abnormal patterns of specific sizes. This bias can ultimately lead to sub-optimal performance when dealing with complex real-world scenarios.

\subsection{Self-supervised learning-based anomaly detection}
\begin{figure*}
\centering
\includegraphics[width=\linewidth]{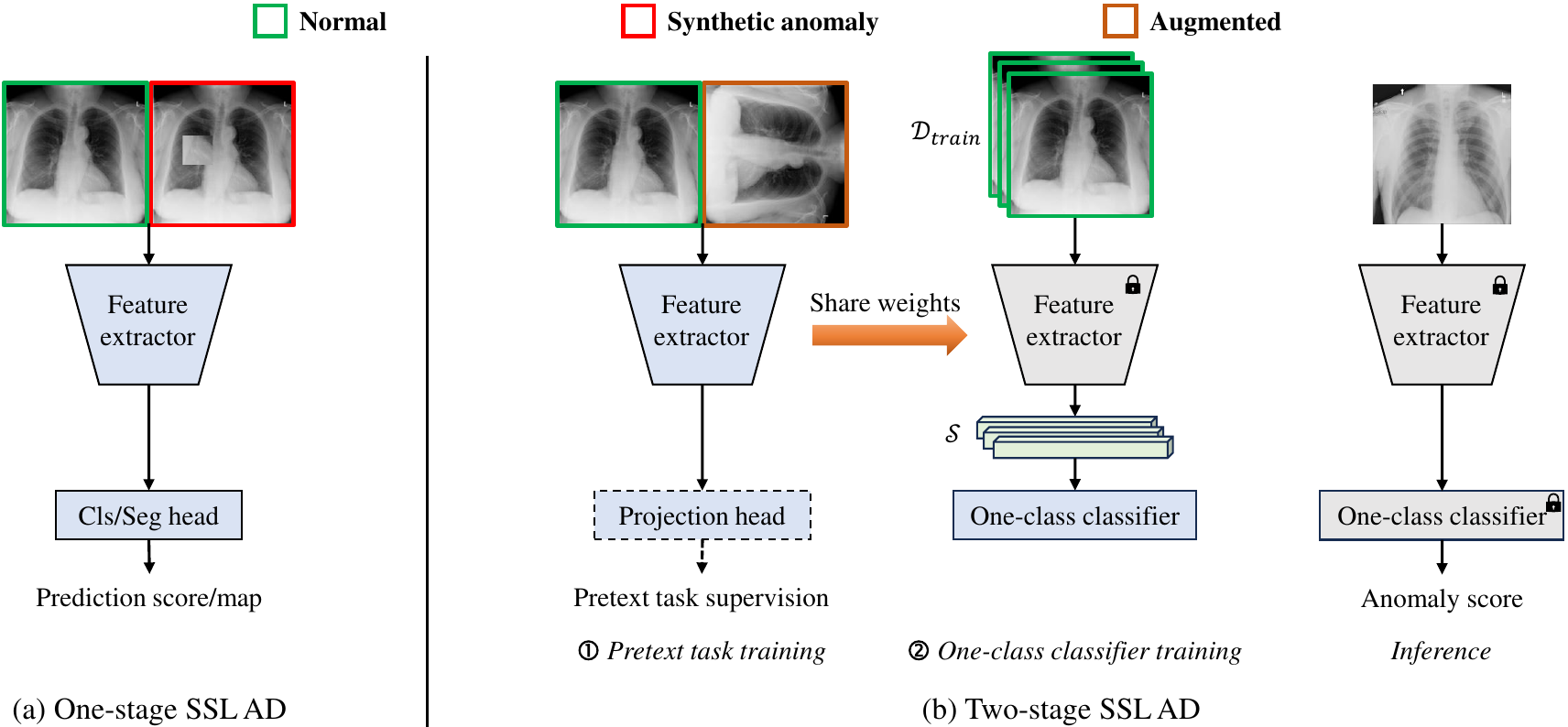}
\caption{Overview of the two paradigms for self-supervised anomaly detection. (a) The one-stage approach trains a model to detect manually synthetic anomalies, and directly applies this model to detect real anomalies. (b) The two-stage approach firstly learns self-supervised representations through a pretext task on the normal training data, and then builds a one-class classifier on the learned representations.}
    \label{fig:overview_ssl}
\end{figure*}

Self-supervised learning (SSL) \citep{jing2020self} is a paradigm of learning methods in which networks are explicitly trained using pretext tasks with generated pseudo labels. This technique has been extensively explored in various methods through different fashions. To provide a clear taxonomy of AD, we specifically focus on methods that directly utilize SSL to train anomaly discrimination networks or learn anomaly-related representations, categorizing them as SSL-based AD methods. The existing SSL-based AD methods have two paradigms: one-stage and two-stage approaches, which are depicted in Fig.~\ref{fig:overview_ssl}. 

\subsubsection{The one-stage approach}
As shown in Fig.~\ref{fig:overview_ssl}(a), the one-stage approach trains a model to detect manually synthetic anomalies, and directly applies the trained model to detect real anomalies. The principle of this approach is to design realistic pseudo anomalies, so that the model trained on these synthetic abnormal patterns can generalize well to the unseen real abnormal samples. 

\citet{tan2020detecting} proposed Foreign Patch Interpolation (FPI), which synthesizes abnormal regions by interpolating a patch in the current normal image using another randomly sampled normal training image. However, the synthesized results are significantly inconsistent with the rest of the image, which may provide no incentive for the model to recognize real anomalies, a much harder task. To enhance the realism of synthesized results, \citet{tan2021detecting} further proposed Poisson image interpolation (PII), which blends the gradient of image patches instead of the raw intensity value \citep{poisson2003}, producing seamless defects. Building upon PII, various efforts have been made to improve performance within this framework. For instance, \citet{schluter2022natural} integrated more augmentation, a patch shape sampling strategy, and background constraints into PII for more diverse and task-relevant pseudo anomalies. 
\citet{p2023confidence} introduced the probabilistic feature into PII by generating circular anomalies with the radius and location sampled from normal distributions. 
Additionally, \citet{baugh2023many} identified a lack of structured validation for these approaches, which could result in overfitting on synthetic anomalies. To tackle this issue, they proposed to use multiple synthetic self-supervision tasks for both training and validation. This approach allows the model's performance on unseen anomalies to be approximated and monitored.

\subsubsection{The two-stage approach}
As shown in Fig.~\ref{fig:overview_ssl}(b), the two-stage approach firstly learns self-supervised representations through a pretext task on the normal training data, and then builds a one-class classifier on the learned representations. This framework was formally presented by \citet{sohn2021learning}. The principle of this approach is to design pretext tasks using the normal training data to learn discriminative representations capable of distinguishing normal and abnormal samples.

One of the popular pretext tasks is contrastive learning. \citet{sohn2021learning} introduced  distribution augmentation \citep{jun2020distribution} for one-class contrastive learning to enhance the compactness of inlier distribution. They further demonstrated that one-class classifiers like one-class SVM (OC-SVM) \citep{scholkopf1999support} and kernel density estimation (KDE) \citep{breunig2000lof} are better than surrogate classifiers based on the simulated outliers \citep{Bergman2020Classification-Based}. \citet{tian2021constrained,tian2023self} combined this approach with designed augmentations to learn medical task-related knowledge. Another commonly used pretext task is to classify normal samples versus synthetic anomalies. \citet{li2021cutpaste} proposed CutPaste, which cuts an image patch and pastes at a random location of the image, to synthesize anomalies. They learned representations by classifying normal data from CutPaste, and then built a Gaussian density estimator (GDE) on the learned representations. Based on CutPaste, \citet{sato2023anatomy} proposed an anatomy-aware pasting (AnatPaste) strategy that utilizes a threshold-based lung segmentation mask to guide the synthesis of anomalies in chest X-rays, taking anatomical structures of the lung into account. 

In addition, \citet{reiss2021panda} emphasized the use and adaptation of ImageNet pre-trained features, which we demonstrate to have impressive performance on medical AD in Section~\ref{sec:exp}. To address the catastrophic collapse \citep{ruff2018deep} during adaptation, they proposed sample-wise early stopping to select checkpoints based on the normalized distances of anomalous samples to the normal center, and used elastic weight consolidation (EWC) \citep{kirkpatrick2017overcoming} inspired by continual learning as a regularization term. \citet{reiss2023mean} explored the standard contrastive loss \citep{chen2020simple} for fine-tuning pre-trained representations for OCC and showed that it achieves poor performance. They attributed this phenomenon to the increase of feature uniformity caused by the standard contrastive loss, which makes anomalies harder to detect. As an alternative, they proposed Mean-Shifted Contrastive Loss which computes the angles between the features of images with respect to the center of the normal features rather than the origin.

\subsubsection{Other SSL-related approaches}
In addition to the aforementioned typical applications, SSL has been designed in various other ways to aid in AD. For example, DRAEM \citep{zavrtanik2021draem} jointly trains a reconstruction network with a segmentation network via SSL. The reconstruction network is trained to repair synthetic abnormal regions in input images, while the segmentation network is expected to generate the segmentation map by learning a distance function between the input and the reconstruction. The authors claimed that it can alleviate the overfitting to synthetic appearance. However, \citet{cai2023dual} argued that the segmentation network in DRAEM tends to learn a shortcut to directly segment the synthetic appearance in the input. They proposed to train a refinement network via SSL based on only anomaly score maps, so that the network never sees synthetic regions, thereby avoiding overfitting. Some recent works \citep{zhang2023destseg,cai2023discrepancy} also have similar designs to refine raw anomaly score maps using SSL.

Moreover, \citet{ristea2022self} proposed the self-supervised predictive convolutional attentive block (SSPCAB), which is an attention block \citep{hu2018squeeze} equipped with a center-masked dilated convolution layer. The block is plugged into anomaly classification or segmentation CNNs, introducing reconstruction of the masked area as a self-supervised task. They claimed that this strategy can help grasp the global arrangement of local features. \citet{madan2023self} further extended the SSPCAB with a 3D masked convolutional layer, a transformer for channel-wise attention, and a novel self-supervised objective based on Huber loss. They demonstrated the broader applicability of this approach across various scenarios.

It is noteworthy that many AD methods in other taxonomies also incorporate SSL designs to improve their performance. However, we do not categorize these methods as SSL-based AD methods due to their distinct working principles, which differ from our definition of SSL-based AD methods but align with other taxonomies. For example, \citet{zimmerer2018context} introduced an inpainting task to AE, and \citet{kascenas2022denoising,kascenas2023role} introduced a denoising task to UNet. Although these methods leverage self-supervised tasks during training, their objective is to facilitate pseudo-healthy reconstruction, which is the goal of reconstruction-based AD methods. Therefore, we categorize these methods as reconstruction-based rather than SSL-based methods\footnote{Note that DAE \citep{kascenas2023role} is also considered a SSL-based AD methods by some studies \citep{lagogiannis2023unsupervised}, as it trains a UNet with skip connections using a self-supervised denoising task. The categorization of DAE may still be controversial.}.

\subsection{Feature reference-based anomaly detection}
Feature reference-based methods perform AD based on the disparity between the current feature and reference ones, and are particularly popular in industrial defect detection. These methods can be categorized into two groups: knowledge distillation and feature modeling approaches. Knowledge distillation approaches utilize feature maps of the current input, extracted by a teacher network, as the reference; while feature modeling approaches employ prototypes of normal training data, extracted by a model pre-trained on ImageNet \citep{krizhevsky2012imagenet}, as the reference.

Specifically, knowledge distillation approaches \citep{bergmann2020uninformed,salehi2021multiresolution,deng2022anomaly,tien2023revisiting,zhang2023destseg} train student networks to regress the normal feature maps of a pre-trained teacher network. It is expected that feature maps of the student network, without knowledge about anomalies, will differ from those of the teacher network on abnormal regions. This discrepancy between the feature maps of the teacher and student is designed as an anomaly score. Feature modeling methods \citep{roth2022towards,jiang2022softpatch,lee2022cfa} utilize intermediate features of a pre-trained network to model the local features of normal image patches, and then select a set of representative prototypes to store in a memory bank. During inference, they measure the distance between the testing patch features and the stored prototypes as the anomaly score.

\section{Benchmark} \label{sec:benchmark}
In this section, we introduce our benchmark from the perspectives of datasets and implementation to facilitate comprehensive and fair comparison of methods for medical AD. Section~\ref{subsec:data} begins by analyzing previous datasets that are particularly easy for AD, and then introduces our curated datasets, followed by an introduction to the evaluation metrics employed. Section~\ref{subsec:impl} introduces our unified implementation with default setups for typical reconstruction- and SSL-based AD methods. 

\subsection{Datasets and metrics}  \label{subsec:data}

\begin{figure}[!b]
\centering
\includegraphics[width=1\linewidth]{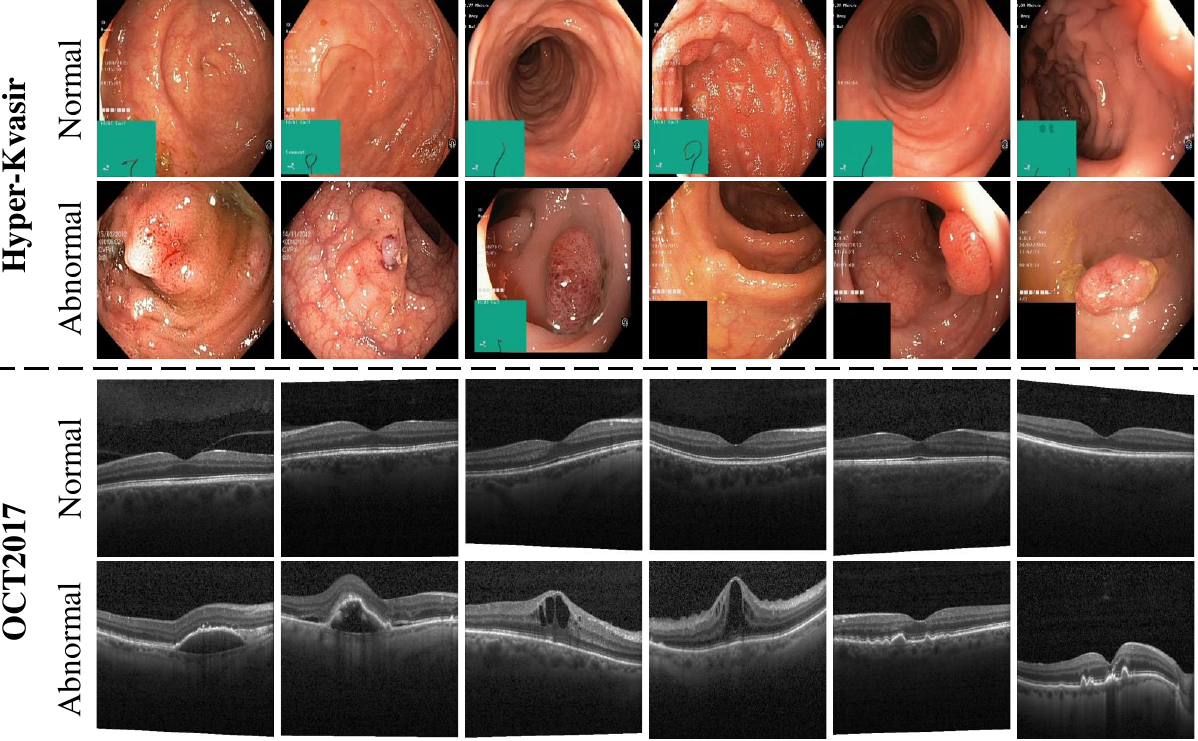}
\caption{Examples of datasets deemed too simple for AD, including the Hyper-Kvasir and OCT2017 datasets.}
\label{fig:simple_data}
\end{figure}

\subsubsection{Datasets not suitable for anomaly detection}
As depicted in Table~\ref{tab:indistinguishable_data}, datasets such as Hyper-Kvasir \citep{borgli2020hyperkvasir} and OCT2017 \citep{kermany2018identifying} are remarkably easy for anomaly detection, rendering them unsuitable for evaluation purposes. This realization prompts us to explore the underlying reasons and aid in the selection of more appropriate datasets for evaluating AD.

In Fig.~\ref{fig:simple_data}, we present examples of these two datasets, from which we identify two key factors. Firstly, the presence of anomaly-unrelated bias in abnormal samples creates an undesirable shortcut for AD. Normal images in the Hyper-Kvasir dataset (following the split from \citet{tian2021constrained}) consistently feature green thumbnails in the lower left corner. In contrast, most abnormal images in the dataset exhibit black thumbnails or no thumbnails in that location. Consequently, these samples can be easily distinguished solely based on the lower left corner, even though this characteristic has no direct correlation with the disease itself. Secondly, distinct low-level differences can make abnormal images conspicuously distinguishable from normal ones. Polyps in the Hyper-Kvasir dataset exhibit clear boundaries and occupies substantial regions within the image, while ophthalmic diseases in OCT2017 dataset display distinct patterns of distorted texture and shape in retinal layers. These abnormal patterns deviate significantly from the corresponding simpler normal patterns, making them easily recognizable.

Therefore, to ensure effective evaluation, it is crucial to select datasets where abnormal samples and normal samples do not exhibit anomaly-unrelated bias, and where abnormal patterns are not excessively different from normal patterns.

\begin{table*}[!t]
\centering
\caption{Summary of the datasets.}   \label{data_repartition}
\begin{tabular}{lllll}
\toprule
\multicolumn{1}{l}{\multirow{2}{*}{\textbf{Dataset}}}       & \multicolumn{1}{l}{\multirow{2}{*}{\textbf{Modality}}}   & \multicolumn{1}{l}{\multirow{2}{*}{\textbf{Task}}}    &\multicolumn{2}{l}{\textbf{Repartition}}      \\ \cmidrule(l){4-5} 
\multicolumn{1}{c}{}                                        & \multicolumn{1}{c}{}          & \multicolumn{1}{c}{}  & $\mathcal{D}_{train}$ & $\mathcal{D}_{test}$ (Normal+Abnormal)   \\ \midrule
RSNA\textsuperscript{\ref{1}}                               & Chest X-ray                   & AnoCls                & 3851                  & 1000+1000     \\
VinDr-CXR\textsuperscript{\ref{2}} \citep{nguyen2022vindr}  & Chest X-ray                   & AnoCls                & 4000                  & 1000+1000     \\
Brain Tumor\textsuperscript{\ref{3}}                        & Brain MRI                     & AnoCls                & 1000                  & 600+600       \\
LAG \citep{li2019attention}                                 & Retinal fundus image          & AnoCls                & 1500                  & 811+811       \\
ISIC2018 \citep{codella2019skin}                            & Dermatoscopic image           & AnoCls                & 6705                  & 909+603       \\
Camelyon16 \citep{bejnordi2017diagnostic}                   & Histopathology image          & AnoCls                & 5088                  & 1120+1113     \\
BraTS2021 \citep{baid2021rsna}                              & Brain MRI                     & AnoCls \& AnoSeg      & 4211                  & 828+1948      \\
\bottomrule
\end{tabular}
\end{table*}

\subsubsection{Image-level anomaly classification datasets}
Six datasets are collected for image-level AnoCls: \\
\textbf{RSNA Dataset}\footnote{\url{https://www.kaggle.com/c/rsna-pneumonia-detection-challenge}\label{1}}. The dataset contains 8851 normal and 6012 lung opacity chest X-rays (CXRs). In experiments, we use 3851 normal images as the normal training dataset $\mathcal{D}_{train}$, and 1000 normal and 1000 abnormal (lung opacity) images as the test dataset $\mathcal{D}_{test}$. \\
\textbf{VinDr-CXR Dataset}\footnote{\url{https://www.kaggle.com/c/vinbigdata-chest-xray-abnormalities-detection}\label{2}} \citep{nguyen2022vindr}. The dataset contains 10606 normal and 4394 abnormal CXRs that include 14 categories of anomalies in total. In experiments, we use 4000 normal images as $\mathcal{D}_{train}$, and 1000 normal and 1000 abnormal images as $\mathcal{D}_{test}$. \\
\textbf{Brain Tumor Dataset}\footnote{\url{https://www.kaggle.com/datasets/masoudnickparvar/brain-tumor-mri-dataset}\label{3}}. The dataset contains 2000 MRI slices with no tumors, 1621 with glioma, and 1645 with meningioma. The glioma and meningioma are regarded as anomalies. The images with no tumor come from Br35H\footnote{\url{https://www.kaggle.com/datasets/ahmedhamada0/brain-tumor-detection}\label{4}} and \citet{saleh2020brain}, while the images with glioma and meningioma come from \citet{saleh2020brain} and \citet{cheng2015enhanced}. In experiments, we use 1000 normal images (with no tumors) as $\mathcal{D}_{train}$, and 600 normal and 600 abnormal images (300 with glioma and 300 with meningioma) as $\mathcal{D}_{test}$. \\
\textbf{LAG Dataset} \citep{li2019attention}. The dataset contains 3143 normal retinal fundus images and 1711 abnormal retinal fundus images with glaucoma. In experiments, we use 1500 normal images as $\mathcal{D}_{train}$, and 811 normal and 811 abnormal images as $\mathcal{D}_{test}$. \\
\textbf{ISIC2018 Dataset} \citep{codella2019skin}. The dataset from Task 3 of the ISIC2018 challenge is used. It contains seven categories, from which we take the nevus (NV) as the normal category, following previous works \citep{lu2018anomaly,guo2023encoder}. In experiments, we use 6705 normal (NV) images from the official training set as $\mathcal{D}_{train}$, and 909 normal (NV) images and 603 abnormal (other six categories) images from the official testing set as $\mathcal{D}_{test}$. \\
\textbf{Camelyon16 Dataset} \citep{bejnordi2017diagnostic}. We use the well-processed version of Camelyon16 for AD provided by \citet{bao2023bmad}, and merge their validation and testing set as our $\mathcal{D}_{test}$. The whole slide images (WSIs) are cropped into $256\times256$ patches at 40$\times$ magnification, which are randomly sampled to build the image dataset for AD. In experiments, we use 5088 normal images as $\mathcal{D}_{train}$, and 1120 normal and 1113 abnormal images as $\mathcal{D}_{test}$.

\subsubsection{Pixel-level anomaly segmentation datasets}
\begin{figure*}[!t]
\centering
\subfigure[Train Normal]{
\begin{minipage}[t]{0.315\linewidth}
\centering
\includegraphics[width=\linewidth]{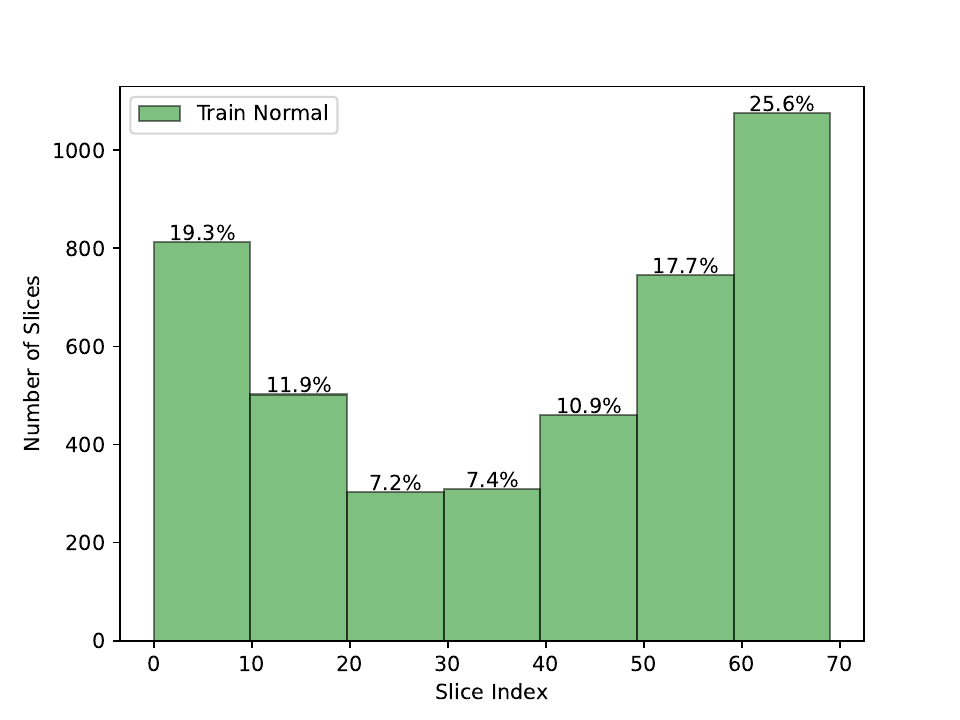}
\end{minipage}
}
\subfigure[Test Normal]{
\begin{minipage}[t]{0.315\linewidth}
\centering
\includegraphics[width=\linewidth]{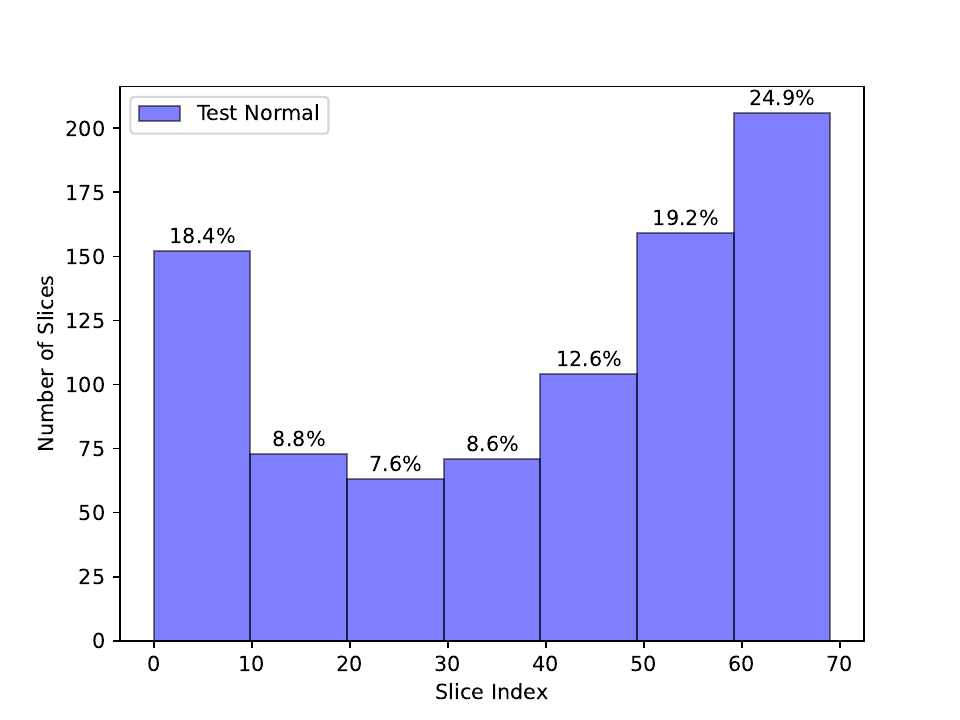}
\end{minipage}
}
\subfigure[Test Abnormal]{
\begin{minipage}[t]{0.315\linewidth}
\centering
\includegraphics[width=\linewidth]{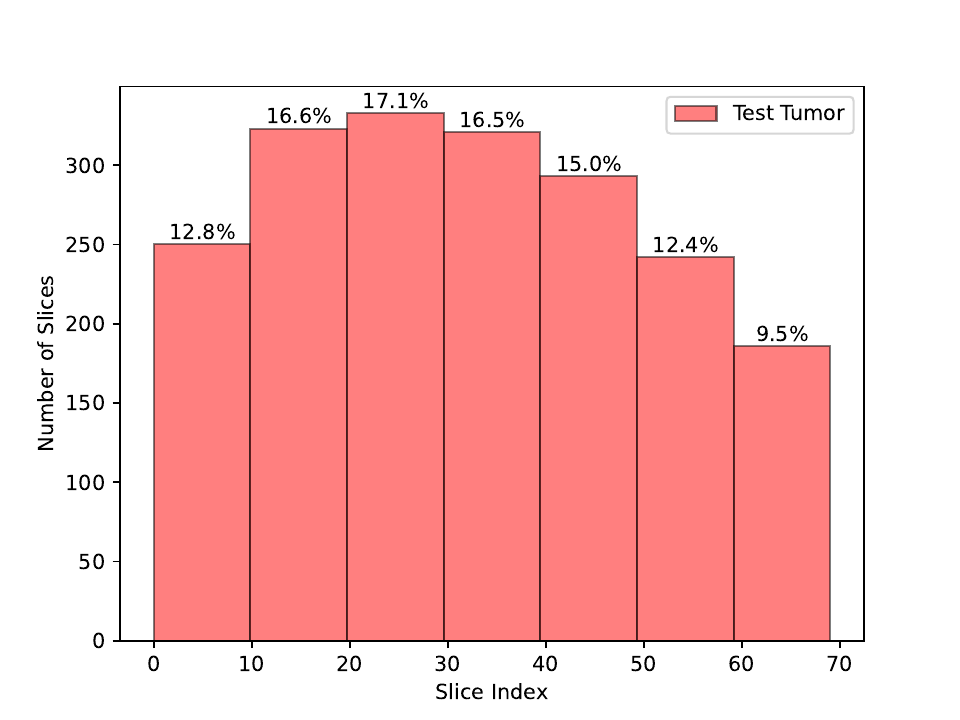}
\end{minipage}
}

\caption{Statistics of slice indices in our processed BraTS2021 dataset. A central crop was performed before slice extraction, resulting in index 0 corresponding to index 50 of the original volume and index 69 corresponding to index 119 of the original volume.} \label{fig:brats_stat}
\end{figure*}

\begin{figure*}[!b]
\centering
\includegraphics[width=0.74\linewidth]{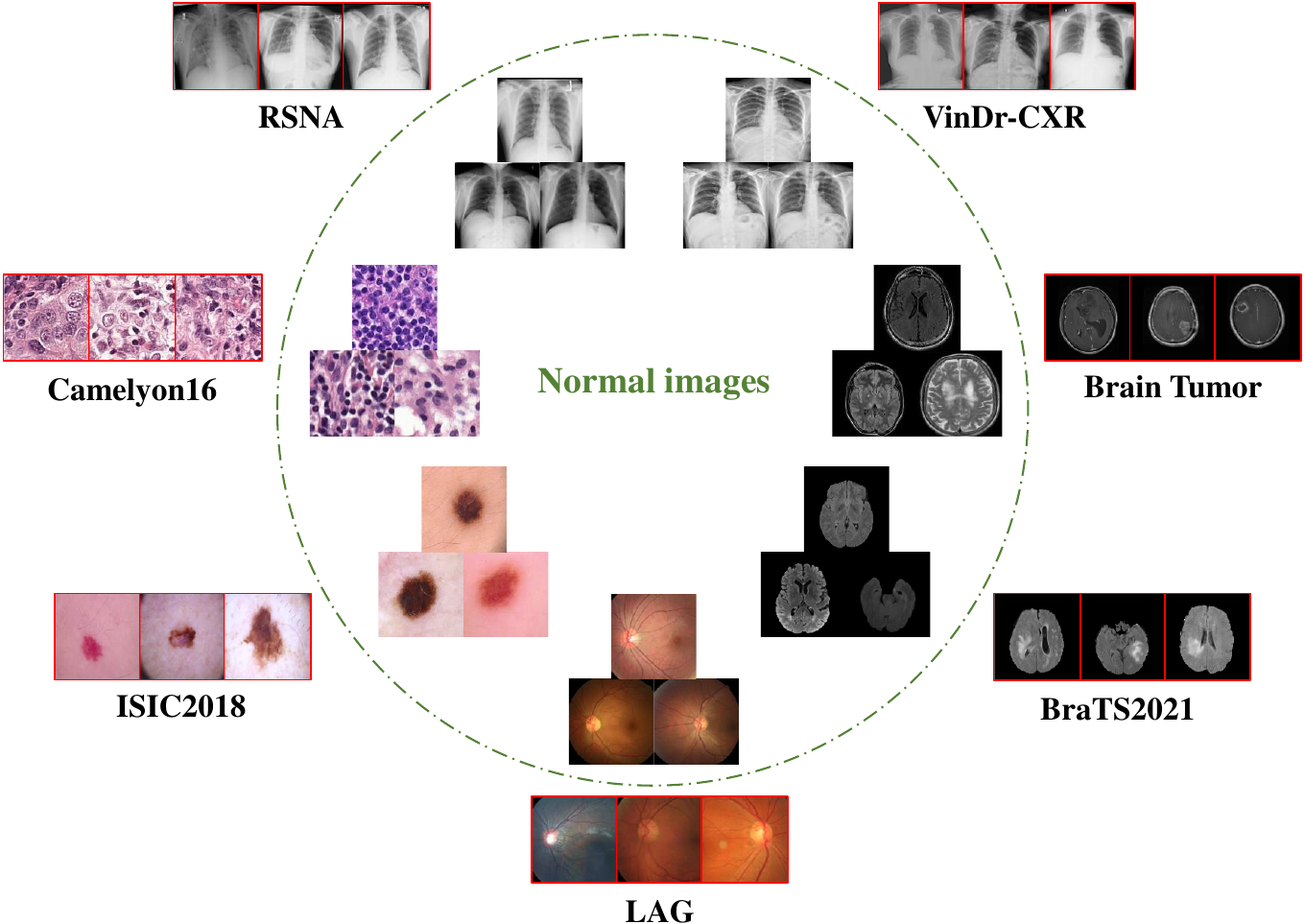}
\caption{Examples of the collected datasets for AD. Samples enclosed by the green dashed circle are normal, while others are abnormal.}
\label{fig:examples}
\end{figure*}

\textbf{BraTS2021} \citep{baid2021rsna} is reorganized for pixel-level anomaly segmentation in this paper. It provides 1251 MRI cases with the resolution $155 \times 240 \times 240$, as well as the corresponding voxel-level annotation for tumor regions. Each case has multiple modalities: T1, T1ce, T2, and FLAIR. We use only the FLAIR in experiments as it is the most sensitive to tumor regions. In preprocessing, the scans undergo a central cropping operation to remove the empty corners, resulting in dimensions of $70 \times 208 \times 208$. For data construction, we utilize 1051 scans for training slice extraction, and the remaining 200 scans for testing slice extraction. To construct $\mathcal{D}_{train}$, we uniformly sample slices from the original normal axial slices without tumors in the scans, ensuring a balanced representation. Additionally, we discard redundant slices to enforce a minimum spacing of five slices between each sampled slice, guaranteeing sufficient content variation among neighboring slices. As a result, a total of 4211 normal 2D axial slices are extracted from 1051 FLAIR scans. Following the same principle, we further extract 828 normal and 1948 tumor slices from the remaining 200 FLAIR scans to build $\mathcal{D}_{test}$.

Due to the intrinsic characteristics of brain MRIs, central slices typically exhibit more relevant content and anomalies, as illustrated in Fig.~\ref{fig:brats_stat}. This imbalance could introduce biases to slice-level anomaly classification. Consequently, this dataset is primarily intended for evaluating pixel-level anomaly segmentation, while we also provide image-level metrics for reference.

\subsubsection{Summary of the datasets} \label{subsec:data_summary}

Fig.~\ref{fig:examples} presents normal and abnormal examples from the seven datasets. Table~\ref{data_repartition} summarizes details of the image modality, supported task, and repartition in the datasets. Regarding the repartition, there is no validation set available for model selection due to the unavailability of abnormal samples and labels during the training of AD models.

The datasets display distinct abnormal patterns. The chest X-ray, brain MRI, and retinal fundus datasets (i.e., RSNA, VinDr-CXR, Brain Tumor, BraTS2021 and LAG) showcase local anomalies. This implies that abnormal images contain the healthy version, with the addition of lesions in some areas. In contrast, Camelyon16 and ISIC2018 present global semantic anomalies, wherein abnormal images belong to completely different categories compared to normal images.

\subsubsection{Evaluation metrics}
Unsupervised AD methods usually generate continuous predictions. However, determining an appropriate threshold for binarizing these predictions remains an unresolved problem \citep{baur2021autoencoders}. Consequently, we evaluate the performance using metrics that are independent of thresholds. For image-level AnoCls, we assess the area under the ROC curve (AUC) and average precision (AP). For pixel-level AnoSeg, we utilize the pixel-level AP ($\text{AP}_\text{pix}$) and the best possible Dice score ($\lceil$Dice$\rceil$). Note that $\lceil$Dice$\rceil$ is a dataset-wise Dice score calculated at the optimal operating point on the testing set, eliminating the need to select a threshold and mitigating the associated effects. We do not employ the pixel-level AUC since it is insensitive to false positives, which is particularly relevant in medical images where the majority of pixels are negative. 

All of our experiments were conducted three times using different random seeds, enabling us to report the mean and standard deviation of the metrics.

\subsection{Implementation}  \label{subsec:impl}

\subsubsection{Unified implementation of the reconstruction methods}  \label{sec:impl_rec}

\begin{table}[ht]
\centering
\caption{Default configuration for AE in reconstruction methods.} \label{tab:default_recon}
\begin{tabular}{ll}
\toprule
Components                  & Configuration         \\  \midrule
Input size $H \times W$     & 64 $\times$ 64        \\  
\#Encoder blocks            & 4                     \\
Block depth $D$             & 1                     \\
Basic width $C_0$           & 16                    \\
Channels of encoder blocks  & $C_0$-2$C_0$-4$C_0$-4$C_0$  \\ 
Latent size $d$             & 16                    \\  
\bottomrule
\end{tabular}
\end{table}

As depicted in Fig.~\ref{fig:overview_rec}, AE for reconstruction AD comprises an encoder that compresses the input into a compact latent feature representation, $\mathbf{z}=f_{e}(\mathbf{x}) \in \mathbb{R}^d$, and a decoder that reconstructs the original input by decoding the representation, $\hat{\mathbf{x}} = f_{d}(\mathbf{z}) \in \mathbb{R}^{C \times H \times W}$.

To ensure fairness, we employ the same AE for all reconstruction-based methods whenever possible. The encoder consists of four convolutional blocks, each downsampling the resolution to $\frac{H_{in}}{2} \times \frac{W_{in}}{2}$. The encoder blocks are then followed by a flattening operation and two fully-connected layers, which compress the feature to a vector of dimension $d$. The decoder exhibits an architecture symmetric to the encoder. Table~\ref{tab:default_recon} summarizes our default configuration for this AE.

\subsubsection{Unified implementation of the self-supervised methods}

\begin{table}[!t]
\centering
\caption{Default configuration for SSL methods. Note that the output layer is a single layer at the end of the cls/seg head. It is omitted in this table for simplicity.} \label{tab:default_ssl}
\begin{tabular}{ll}
\toprule
Components                  & Configuration                                 \\  \midrule
Input size $H \times W$     & 224 $\times$ 224                              \\
Feature extractor $f$       & ResNet18 (Conv layers)                        \\  
Classification head $g_c$   & One linear layer. [512]                       \\
Segmentation head  $g_s$    & Three residual blocks. [64, 32, 16]           \\
One-class classifier $\psi$ & Gaussian density estimator (GDE)              \\  \bottomrule
\end{tabular}
\end{table}

As shown in Fig.~\ref{fig:overview_ssl}, the one-stage SSL approach comprises a feature extractor $f$ and a classification/segmentation head $g_c$/$g_s$, while the two-stage approach comprises a feature extractor $f$, a projection head $g$, and a one-class classifier $\psi$. 

To ensure fairness, we employ ResNet18 \citep{he2016deep}, excluding classification layers, as the feature extractor for all SSL-based methods whenever possible. The input resolution is set to $224 \times 224$. For additional specifics, we use the unified implementation from \citet{schluter2022natural} as the framework for all one-stage segmentation approaches, and the implementation from \url{https://github.com/Runinho/pytorch-cutpaste} as the foundation for all two-stage approaches. Since most two-stage approaches utilize the classification task as the pretext task, we implement the one-stage classification approaches based on the same framework, but retain the classification head for evaluation. Table~\ref{tab:default_ssl} summarizes our default configuration for SSL methods.

\section{Experiments and analysis} \label{sec:exp}
\begin{table*}[!t]
\centering
\caption{Summary of AD methods involved in the experiments. img-rec: image-reconstruction; feat-rec: feature-reconstruction; 1-stage-c: one-stage classification; 1-stage-s: one-stage segmentation. IN: ImageNet pre-trained. $^\dag$Methods only support sample-wise anomaly score.}  \label{tab:method_list}

\resizebox{\linewidth}{!}{
\begin{tabular}{llllll}
\toprule
& Method & Backbone & Loss Function & Anomaly score at $i$-th px & Remarks \\ \midrule
& Intensity & -- & -- & \makecell[l]{$\mathbf{x}_i$ or $-\mathbf{x}_i$ \\ (subject to the dataset)} & \makecell[l]{$-\mathbf{x}_i$ for the Brain Tumor dataset \\ $\mathbf{x}_i$ for others} \\ \midrule \midrule

\multicolumn{3}{l}{\textbf{Reconstruction AD}} \\ \midrule
\multirow{26}{*}{\rotatebox{90}{img-rec}}   
& AE                                    & AE        & $\mathcal{L}_0=\ell_2(\mathbf{x}, \hat{\mathbf{x}})$                       & $(\mathbf{x}_i - \hat{\mathbf{x}}_i)^2$  &  \\ \cmidrule(l){2-6}
& AE-$\ell_1$                           & AE        & $\ell_1(\mathbf{x}, \hat{\mathbf{x}})$                        & $\vert \mathbf{x}_i - \hat{\mathbf{x}}_i \vert$   &      \\ \cmidrule(l){2-6}
& AE-SSIM \citep{bergmann2018improving} & AE        & \texttt{(1-SSIM($\mathbf{x}, \hat{\mathbf{x}}$)).mean()}  & \texttt{Up(1-SSIM$(\mathbf{x}, \hat{\mathbf{x}})$)}$_i$  
    & \makecell[l]{\texttt{Up}: resize to $H\times W$}  \\ \cmidrule(l){2-6}
& AE-PL \citep{shvetsova2021anomaly}    & AE+VGG19-IN        & \texttt{PL($\mathbf{x}, \hat{\mathbf{x}}$).mean()}        & \texttt{Up(PL$(\mathbf{x}, \hat{\mathbf{x}})$)}$_i$  
    & \texttt{PL} at \texttt{vgg19.conv4\_2} \\ \cmidrule(l){2-6}
& VAE \citep{kingma2013auto}            & VAE       & \makecell[l]{$\mathcal{L}_0+$ \\ $\lambda \ell_{KL}(\mathcal{N}(\mu(\mathbf{x}), \sigma^2(\mathbf{x}))) \| \mathcal{N}(\mathbf{0},\mathbf{I}))$} & $(\mathbf{x}_i - \hat{\mathbf{x}}_i)^2$ & $\lambda=0.005$ \\ \cmidrule(l){2-6}
& Constrained AE \citep{chen2018unsupervised} & AE  & $\mathcal{L}_0 + \ell_2(f_e(\mathbf{x}), f_e(\hat{\mathbf{x}}))$ & $(\mathbf{x}_i - \hat{\mathbf{x}}_i)^2$ & \\ \cmidrule(l){2-6}
& MemAE \citep{gong2019memorizing}       & AE+Memory & $\mathcal{L}_0$ + $\lambda \mathcal{L}_{entropy}(f_e(\mathbf{x}), \mathbf{M})$ & $(\mathbf{x}_i - \hat{\mathbf{x}}_i)^2$ 
    & \makecell[l]{$\mathbf{M}$: Memory bank \\ $\lambda=0.0002$}  \\ \cmidrule(l){2-6}
& CeAE \citep{zimmerer2018context}      & AE        & $\ell_2(\mathbf{x}, \hat{\mathbf{x}}\textsuperscript{ce})$                  & $(\mathbf{x}_i - \hat{\mathbf{x}}_i)^2$           & $\mathbf{x}\textsuperscript{ce}$: $\mathbf{x}$ masked with 10 $\times$ 10 px \\ \cmidrule(l){2-6}

& AnoDDPM \citep{Wyatt_2022_CVPR} & UNet [DDPM] & $\mathbb{E}_{t \sim[1,T], x \sim q(x)}\left[\left\|\epsilon-\epsilon_\theta\left(x_t, t\right)\right\|^2\right]$ &  $(\mathbf{x}_i - \hat{\mathbf{x}}_i)^2$   & \makecell[l]{$\epsilon$: Simplex noise  \\ $\epsilon_\theta$: UNet  \\  t: Time step  \\ Input size: $256\times256$}   \\ \cmidrule(l){2-6}
& AutoDDPM \citep{bercea2023mask} & UNet [DDPM] & $\mathbb{E}_{t \sim[1,T], x \sim q(x), \epsilon \sim \mathcal{N}(0,1)}\left[\left\|\epsilon-\epsilon_\theta\left(x_t, t\right)\right\|^2\right]$ & $\vert \hat{\mathbf{x}}-\mathbf{x} \vert * \hat{m}$ & \makecell[l]{Customized inference step: \\ Mask, Stitch, and Re-Sample. \\ $\hat{m}$: The initial mask} \\ \cmidrule(l){2-6}

& f-AnoGAN \citep{schlegl2019f}         & \makecell[l]{Customized \\ GAN + Encoder}  & \makecell[l]{Stage 1 ($f_{gen} \& f_{dis}$): \\ ~~$\ell_{WGAN}$ \citep{arjovsky2017wasserstein}; \\ Stage 2 ($f_e$): \\ ~~$\ell_2(\mathbf{x}, \hat{\mathbf{x}})+\ell_2(f_{dis}(\mathbf{x}), f_{dis}(\hat{\mathbf{x}}))$.}  & $\ell_2(\mathbf{x}, \hat{\mathbf{x}})+\ell_2(f_{dis}(\mathbf{x}), f_{dis}(\hat{\mathbf{x}})) ^\dag$  & \makecell[l]{$f_e$: Encoder; $f_{gen}$: Generator \\ $f_{dis}$: Discriminator \\ $\hat{\mathbf{x}}=f_{gen}(f_e(\mathbf{x}))$ \\ AnoSeg score: $(\mathbf{x}_i - \hat{\mathbf{x}}_i)^2$}  \\ \cmidrule(l){2-6}
& GANomaly \citep{akcay2019ganomaly}    & \makecell[l]{AE+Encoder \vspace{4pt} \\ + Discriminator} & \makecell[l]{G: $\alpha \mathcal{L}_0 + \beta \ell_2(f_{dis}(\mathbf{x}), f_{dis}(\hat{\mathbf{x}}))$ \\$~~+ \gamma \ell_2(f_e(\mathbf{x}), f_{e2}(\hat{\mathbf{x}}))$ \vspace{4pt} \\ D: $\ell_{bce}(c(f_{dis}(\mathbf{x})), 1)$ \\ $~~+\ell_{bce}(c(f_{dis}(\hat{\mathbf{x}})), 0)$}  & \makecell[l]{\texttt{model.train()} \\ $\ell_2(f_e(\mathbf{x}), f_{e2}(\hat{\mathbf{x}})) ^\dag$}
& \makecell[l]{$\alpha, \beta, \gamma = 50, 1, 1$ \\ $f_{e2}$: the extra encoder \\ $f_{dis}$: Discriminator \\ $c$: the classifier after $f_{dis}$ \\ Inference under \texttt{model.train()}} \\ \cmidrule(l){2-6}
& AE-U \citep{mao2020abnormality}       & AE (\texttt{out\_channels}$\times 2$) & $\frac{(\mathbf{x}_i-\hat{\mathbf{x}}_i)^2}{\sigma_i^2}+\log \sigma_i^2$   & $\frac{(\mathbf{x}_i-\hat{\mathbf{x}}_i)^2}{\sigma_i^2}$  & $\sigma$: predicted pixel uncertainty \\ \cmidrule(l){2-6}
& DAE \citep{kascenas2023role}          & Customized UNet   & $\ell_2(\mathbf{x}, \hat{\mathbf{x}}\textsuperscript{cn})$ & $(\mathbf{x}_i - \hat{\mathbf{x}}_i)^2$ & \makecell[l]{$\mathbf{x}\textsuperscript{cn}$: $\mathbf{x}$ + coarse noise  \\  Input size: $128\times128$} \\ \cmidrule(l){2-6}
& AE-Grad \citep{zimmerer2019unsupervised} & AE & $\mathcal{L}_0$ & $\vert\frac{\partial \ell_2(\mathbf{x}, \hat{\mathbf{x}})}{\partial \mathbf{x}}\vert_i$ & \\ \cmidrule(l){2-6}
& VAE-Grad{\scriptsize rec} \citep{zimmerer2019unsupervised} & VAE & $\mathcal{L}_{VAE}$ & $\vert\frac{\partial \ell_2(\mathbf{x}, \hat{\mathbf{x}})}{\partial \mathbf{x}}\vert_i$ & \\ \cmidrule(l){2-6}
& VAE-Grad{\scriptsize combi} \citep{zimmerer2019unsupervised} & VAE & $\mathcal{L}_{VAE}$ & $(\mathbf{x}_i - \hat{\mathbf{x}}_i)^2 \cdot \vert \frac{\partial \ell_{KL}}{\partial \mathbf{x}}\vert_i$ & \\ \midrule
\multirow{8}{*}{\rotatebox{90}{feat-rec}}
& FAE-SSIM \citep{meissen2022unsupervised}  & \makecell[l]{ResNet18 (2 blocks) \\ Customized AE} & \texttt{(1-SSIM($\mathbf{F}, \hat{\mathbf{F}}$)).mean()} & \texttt{Up(1-SSIM$(\mathbf{F}, \hat{\mathbf{F}})$).mean(0)}$_i$ & $\mathbf{F} \in \mathbb{R}^{C_F \times H_F \times W_F}$: feature maps \\ \cmidrule(l){2-6}
& FAE-MSE \citep{meissen2022unsupervised}   & \makecell[l]{ResNet18 (2 blocks) \\ Customized AE} & $\ell_2(\mathbf{F}, \hat{\mathbf{F}})$ & \texttt{Up$(\mathbf{F} - \hat{\mathbf{F}})^2$.mean(0)$_i$} &  \\  \cmidrule(l){2-6}
& RD \citep{deng2022anomaly} & \multicolumn{3}{l}{[Reference to the original paper]}  \\  \cmidrule(l){2-6}
& RD++ \citep{tien2023revisiting} & \multicolumn{3}{l}{[Reference to the original paper]}  \\  \cmidrule(l){2-6}
& ReContrast \citep{guo2023recontrast} & \multicolumn{3}{l}{[Reference to the original paper]}  \\  \midrule  \midrule
\multicolumn{3}{l}{\textbf{Self-supervised learning AD}} \\ \midrule
\multirow{6}{*}{\rotatebox{90}{2-stage}}
& PANDA \citep{reiss2021panda}          & ResNet152-IN      & $\ell_{center} + \ell_{ewc}$  & $\psi(f(\mathbf{x})) ^\dag$ &    \\ \cmidrule(l){2-6}
& MSC \citep{reiss2023mean}             & ResNet18-IN       & $\ell_{msc}$                  & $\psi(f(\mathbf{x})) ^\dag$ & \\ \cmidrule(l){2-6}
& CutPaste \citep{li2021cutpaste}       & ResNet18          & $\ell_{ce}(g_c(f(\mathbf{x})), 0)+ \ell_{ce}(g_c(f(\mathbf{x}\textsuperscript{cp})), 1)$ & $\psi(f(\mathbf{x})) ^\dag$ & $\mathbf{x}\textsuperscript{cp}$: CutPaste samples  \\ \cmidrule(l){2-6}
& CutPaste-3way \citep{li2021cutpaste}  & ResNet18          & $\ell_{cutpaste}+\ell_{ce}(g_c(f(\mathbf{x}\textsuperscript{scar})), 1)$ & $\psi(f(\mathbf{x})) ^\dag$ & scar: scar-shape CutPaste  \\ \cmidrule(l){2-6}
& AnatPaste \citep{sato2023anatomy}     & ResNet18          & $\ell_{ce}(g_c(f(\mathbf{x})), 0)+ \ell_{ce}(g_c(f(\mathbf{x}\textsuperscript{ap})), 1)$ & $\psi(f(\mathbf{x})) ^\dag$ & $\mathbf{x}\textsuperscript{ap}$: AnatPaste samples  \\ \midrule
\multirow{4}{*}{\rotatebox{90}{1-stage-c}}
& CutPaste \citep{li2021cutpaste}       & ResNet18          & Same as the 2-stage version & $g_c(f(\mathbf{x})) ^\dag$ &  \\ \cmidrule(l){2-6}
& CutPaste-3way \citep{li2021cutpaste}  & ResNet18          & Same as the 2-stage version & $g_c(f(\mathbf{x})) ^\dag$ &  \\ \cmidrule(l){2-6}
& AnatPaste \citep{sato2023anatomy}     & ResNet18          & Same as the 2-stage version & $g_c(f(\mathbf{x})) ^\dag$ &  \\ \midrule
\multirow{5}{*}{\rotatebox{90}{1-stage-s}}
& CutPaste \citep{schluter2022natural}  & ResNet18          & $\ell_{bce}(g_s(f(\mathbf{x}\textsuperscript{cp})), \mathbf{y}_{pseudo})$ & $g_s(f(\mathbf{x}))_i$ & $\mathbf{y}_{pseudo}$: pixel-wise pseudo label \\ \cmidrule(l){2-6}
& FPI \citep{tan2020detecting}          & ResNet18          & $\ell_{bce}(g_s(f(\mathbf{x}\textsuperscript{fpi})), \mathbf{y}_{pseudo})$ & $g_s(f(\mathbf{x}))_i$ & \\ \cmidrule(l){2-6}
& PII \citep{tan2021detecting}          & ResNet18          & $\ell_{bce}(g_s(f(\mathbf{x}\textsuperscript{pii})), \mathbf{y}_{pseudo})$ & $g_s(f(\mathbf{x}))_i$ & \\ \cmidrule(l){2-6}
& NSA \citep{schluter2022natural}       & ResNet18          & $\ell_{bce}(g_s(f(\mathbf{x}\textsuperscript{nsa})), \mathbf{y}_{pseudo})$ & $g_s(f(\mathbf{x}))_i$ & \\  \bottomrule
\end{tabular}
}

\end{table*}

To provide a comprehensive evaluation and analysis for medical AD methods, we assess a total of thirty AD methods on five different image modalities. This extensive coverage guarantees the generalizability of our results and findings. In addition, we evaluate the performance of directly using pixel intensity as the anomaly score \citep{meissen2021challenging}, denoted as `Intensity', which reflects the distinguishability between normal and abnormal pixel intensity within datasets. Table~\ref{tab:method_list} summarizes the methods in our experiments. 

\begin{figure*}[!t]
\centering
\includegraphics[width=\linewidth]{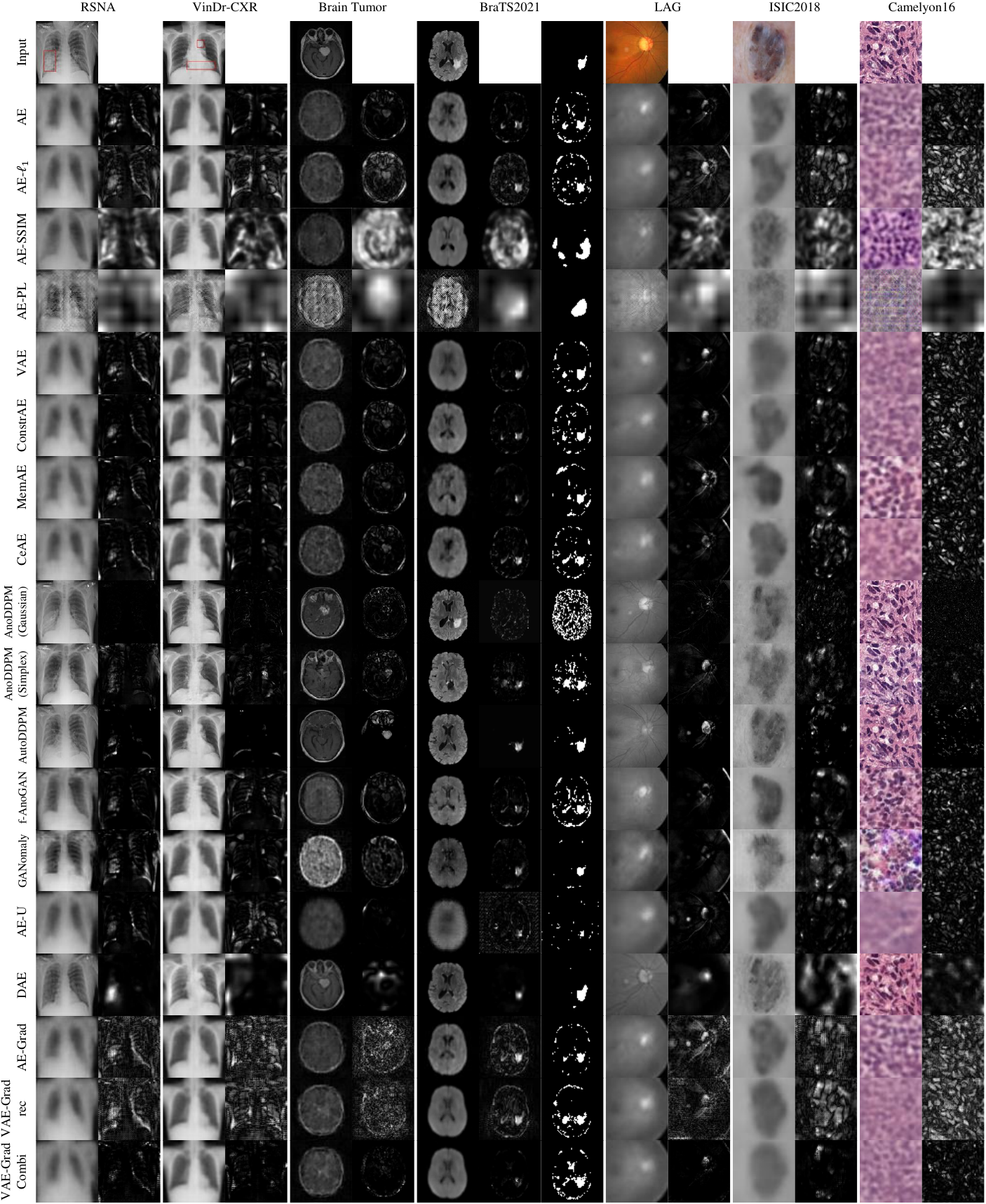}
\caption{Visualization of the compared image-reconstruction methods on typical abnormal images. The first row displays input abnormal images along with the abnormal regions (if available). The second rows depict reconstructed images and anomaly maps generated by each method. The third column of BraTS2021 dataset depicts binarized predictions at operating points of $\lceil$Dice$\rceil$.}
\label{fig:visualization}
\end{figure*}

\begin{figure}[!t]
\centering
\includegraphics[width=\linewidth]{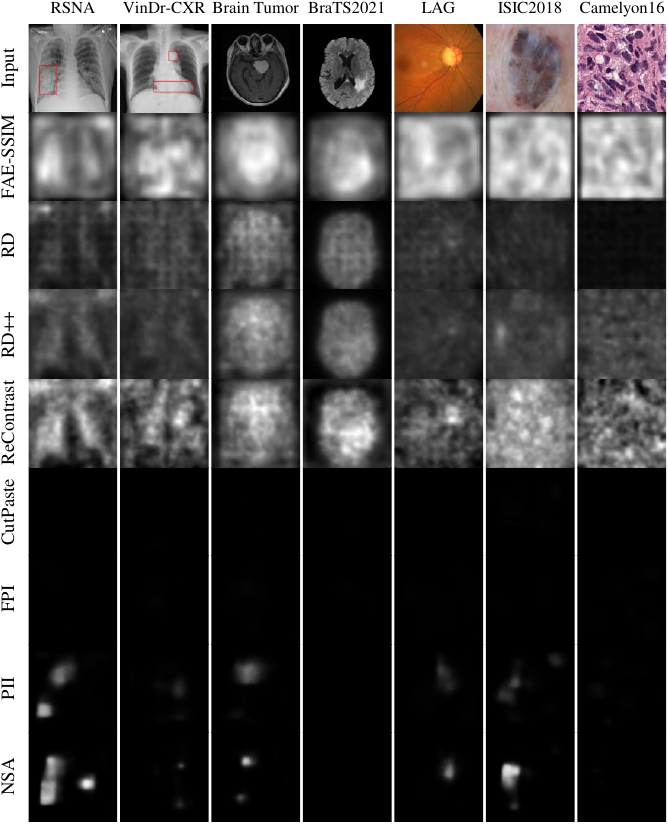}
\caption{Visualization of the compared methods (excluding image-reconstruction methods) on typical abnormal images. The first row presents the input abnormal images, while the subsequent rows depict the anomaly maps generated by each method.}
\label{fig:visualization_2}
\end{figure}

Tables~\ref{tab:ano_cls} and \ref{tab:ano_seg} present the performance of the methods on AnoCls and AnoSeg, respectively. For qualitative analysis, we visualize reconstructed images and prediction maps generated by image-reconstruction methods in Fig.~\ref{fig:visualization}, and visualize prediction maps generated by other methods in Fig.~\ref{fig:visualization_2}. In the subsequent sections, we will first analyze the state-of-the-arts, and then delve into a detailed discussion of important findings from both the reconstruction and SSL methods.

\begin{table*}[t]
\centering
\caption{Performance on image-level AnoCls. In each paradigm, the \red{best} results are highlighted in red, while the \blue{second-} and \blue{third-} best results are highlighted in blue. img-rec: image-reconstruction; feat-rec: feature-reconstruction; 1-stage-c: one-stage classification; 1-stage-s: one-stage segmentation. $^\dag$Methods utilize ImageNet pre-trained weights.}
\label{tab:ano_cls}
\resizebox{\linewidth}{!}{
\begin{tabular}{lllcccccccccccccc}
\toprule
& \multirow{2}{*}{Method} & \multirow{2}{*}{\#Params/FLOPs} & \multicolumn{2}{c}{ RSNA } & \multicolumn{2}{c}{ VinDr-CXR } & \multicolumn{2}{c}{ Brain Tumor } & \multicolumn{2}{c}{ LAG } & \multicolumn{2}{c}{ ISIC2018 } & \multicolumn{2}{c}{ Camelyon16 } & \multicolumn{2}{c}{ BraTS2021 } \\
\cmidrule(l){4-5}\cmidrule(l){6-7} \cmidrule(l){8-9} \cmidrule(l){10-11} \cmidrule(l){12-13} \cmidrule(l){14-15} \cmidrule(l){16-17}
& & & AUC & AP & AUC & AP & AUC & AP & AUC & AP & AUC & AP & AUC & AP & AUC & AP \\  \midrule 
\rowcolor{gray!20} & Intensity & -- & 45.8 & 46.4 & 46.3 & 46.7 & 80.4 & 78.1 & 41.3 & 42.6 & 51.2 & 40.1 & 52.1 & 53.1 & 66.8 & 81.5 \\ \midrule \midrule

\multicolumn{3}{l}{\textbf{Reconstruction AD}} \\ \midrule
\multirow{17}{*}{\rotatebox{90}{img-rec}} 
&AE        & 2.35M/29.9M & 67.5\scriptsize{$\pm$0.9} & 66.7\scriptsize{$\pm$0.5} & 56.4\scriptsize{$\pm$0.4} & 60.2\scriptsize{$\pm$0.5} & 85.9\scriptsize{$\pm$0.3} & 76.9\scriptsize{$\pm$0.1} & 79.0\scriptsize{$\pm$1.0} & 75.7\scriptsize{$\pm$1.4} & 74.5\scriptsize{$\pm$0.8} & 63.0\scriptsize{$\pm$0.6} & 36.8\scriptsize{$\pm$1.2} & 40.3\scriptsize{$\pm$0.6} & 82.6\scriptsize{$\pm$0.1} & 92.0\scriptsize{$\pm$0.1} \\
&AE-$\ell_1$   & 2.35M/29.9M & 68.1\scriptsize{$\pm$0.4} & 67.9\scriptsize{$\pm$0.4} & 57.6\scriptsize{$\pm$0.4} & 61.1\scriptsize{$\pm$0.5} & 80.1\scriptsize{$\pm$0.5} & 72.4\scriptsize{$\pm$0.4} & 76.1\scriptsize{$\pm$2.4} & 72.1\scriptsize{$\pm$1.7} & 75.1\scriptsize{$\pm$0.4} & 64.3\scriptsize{$\pm$0.6} & 35.3\scriptsize{$\pm$0.3} & 39.4\scriptsize{$\pm$0.2} & 81.6\scriptsize{$\pm$0.4} & 89.9\scriptsize{$\pm$0.5} \\
&AE-SSIM   & 2.35M/29.9M & 80.9\scriptsize{$\pm$0.3} & 78.6\scriptsize{$\pm$0.3} & 54.5\scriptsize{$\pm$0.2} & 55.2\scriptsize{$\pm$0.2} & 92.7\scriptsize{$\pm$0.2} & 85.4\scriptsize{$\pm$0.3} & 67.6\scriptsize{$\pm$0.5} & 64.3\scriptsize{$\pm$0.5} & \blue{77.0}\scriptsize{$\pm$0.1} & \blue{65.4}\scriptsize{$\pm$0.2} & 38.0\scriptsize{$\pm$0.9} & 41.2\scriptsize{$\pm$0.3} & 80.3\scriptsize{$\pm$0.2} & 86.0\scriptsize{$\pm$0.2} \\
&AE-PL$^\dag$  & 2.35M/29.9M & \blue{87.5}\scriptsize{$\pm$0.2} & \blue{84.8}\scriptsize{$\pm$0.6} & \blue{75.3}\scriptsize{$\pm$0.1} & \blue{73.7}\scriptsize{$\pm$0.3} & \red{95.7}\scriptsize{$\pm$0.0} & \blue{92.1}\scriptsize{$\pm$0.1} & \red{85.6}\scriptsize{$\pm$0.4} & \red{81.6}\scriptsize{$\pm$0.5} & 68.4\scriptsize{$\pm$1.1} & 51.9\scriptsize{$\pm$1.8} & \blue{76.1}\scriptsize{$\pm$0.1} & \blue{67.6}\scriptsize{$\pm$0.4} & \blue{85.1}\scriptsize{$\pm$0.3} & 92.4\scriptsize{$\pm$0.2} \vspace{4pt}\\
&AE[$d=4$] & 2.33M/29.9M & 72.9\scriptsize{$\pm$2.1} & 70.3\scriptsize{$\pm$0.9} & 61.3\scriptsize{$\pm$0.6} & 64.1\scriptsize{$\pm$0.8} & 70.8\scriptsize{$\pm$1.4} & 64.1\scriptsize{$\pm$1.0} & \blue{83.3}\scriptsize{$\pm$1.0} & \blue{78.9}\scriptsize{$\pm$0.7} & 67.8\scriptsize{$\pm$0.2} & 57.1\scriptsize{$\pm$0.5} & 34.5\scriptsize{$\pm$0.1} & 39.7\scriptsize{$\pm$0.1} & 79.4\scriptsize{$\pm$0.1} & 89.6\scriptsize{$\pm$0.3} \\
&VAE       & 2.37M/29.9M & 67.9\scriptsize{$\pm$0.8} & 67.0\scriptsize{$\pm$0.8} & 56.4\scriptsize{$\pm$0.4} & 60.1\scriptsize{$\pm$0.5} & 85.3\scriptsize{$\pm$0.4} & 76.5\scriptsize{$\pm$0.3} & 79.4\scriptsize{$\pm$0.3} & 76.4\scriptsize{$\pm$0.7} & 74.5\scriptsize{$\pm$0.3} & 63.9\scriptsize{$\pm$0.3} & 36.2\scriptsize{$\pm$0.3} & 40.1\scriptsize{$\pm$0.1} & 80.6\scriptsize{$\pm$0.8} & 90.9\scriptsize{$\pm$0.5} \\
&Constrained AE    & 2.35M/29.9M & 67.5\scriptsize{$\pm$0.7} & 66.7\scriptsize{$\pm$0.6} & 54.6\scriptsize{$\pm$1.3} & 58.8\scriptsize{$\pm$0.8} & 82.3\scriptsize{$\pm$0.5} & 73.7\scriptsize{$\pm$0.6} & 70.0\scriptsize{$\pm$3.3} & 61.4\scriptsize{$\pm$1.9} & 74.7\scriptsize{$\pm$0.2} & 63.3\scriptsize{$\pm$0.1} & 36.8\scriptsize{$\pm$0.4} & 40.3\scriptsize{$\pm$0.2} & 82.5\scriptsize{$\pm$0.3} & 91.3\scriptsize{$\pm$0.9} \\
&MemAE     & 2.35M/29.9M & 69.6\scriptsize{$\pm$0.4} & 68.1\scriptsize{$\pm$0.4} & 56.9\scriptsize{$\pm$0.9} & 60.9\scriptsize{$\pm$0.9} & 82.6\scriptsize{$\pm$0.6} & 74.3\scriptsize{$\pm$0.2} & 79.5\scriptsize{$\pm$1.4} & 76.2\scriptsize{$\pm$0.5} & 69.1\scriptsize{$\pm$3.0} & 59.2\scriptsize{$\pm$2.7} & 39.4\scriptsize{$\pm$1.3} & 41.4\scriptsize{$\pm$0.7} & 63.9\scriptsize{$\pm$7.7} & 78.5\scriptsize{$\pm$6.9} \vspace{4pt} \\
&CeAE      & 2.35M/29.9M & 68.0\scriptsize{$\pm$0.5} & 66.8\scriptsize{$\pm$0.6} & 56.1\scriptsize{$\pm$0.2} & 60.1\scriptsize{$\pm$0.3} & 85.3\scriptsize{$\pm$1.5} & 76.0\scriptsize{$\pm$1.5} & 79.3\scriptsize{$\pm$1.4} & 76.2\scriptsize{$\pm$1.2} & 75.0\scriptsize{$\pm$0.0} & 63.4\scriptsize{$\pm$0.2} & 36.5\scriptsize{$\pm$0.4} & 40.2\scriptsize{$\pm$0.2} & 81.7\scriptsize{$\pm$1.5} & 91.5\scriptsize{$\pm$0.7} \\
&AnoDDPM\tiny{Gaussian} & 132M/278G{\tiny$\times200$} & 52.3\scriptsize{$\pm$1.3} & 57.6\scriptsize{$\pm$0.7} & 41.2\scriptsize{$\pm$1.0} & 42.6\scriptsize{$\pm$0.3} & 39.9\scriptsize{$\pm$2.4} & 41.1\scriptsize{$\pm$0.9} & 41.2\scriptsize{$\pm$2.8} & 45.3\scriptsize{$\pm$2.0} & 73.1\scriptsize{$\pm$0.6} & 60.4\scriptsize{$\pm$0.4} & 43.7\scriptsize{$\pm$0.4} & 50.0\scriptsize{$\pm$0.1} & 68.8\scriptsize{$\pm$0.4} & 79.9\scriptsize{$\pm$0.1} \\
&AnoDDPM\tiny{Simplex} & 132M/278G{\tiny$\times200$} & 79.4\scriptsize{$\pm$0.9} & 78.9\scriptsize{$\pm$0.7} & 55.5\scriptsize{$\pm$0.7} & 57.2\scriptsize{$\pm$0.5} & 67.2\scriptsize{$\pm$0.1} & 65.6\scriptsize{$\pm$0.7} & 58.5\scriptsize{$\pm$1.5} & 57.9\scriptsize{$\pm$0.9} & 66.7\scriptsize{$\pm$0.5} & 56.7\scriptsize{$\pm$0.9} & \blue{70.6}\scriptsize{$\pm$0.4} & \blue{70.4}\scriptsize{$\pm$0.1} & 82.7\scriptsize{$\pm$0.8} & 91.7\scriptsize{$\pm$0.7} \\
&AutoDDPM & 18.0M/65.1G{\tiny$\times200$} & 83.6\scriptsize{$\pm$0.2} & 81.4\scriptsize{$\pm$0.1} & 69.4\scriptsize{$\pm$0.2} & 68.6\scriptsize{$\pm$0.2} & 92.7\scriptsize{$\pm$0.2} & 87.8\scriptsize{$\pm$0.2} & 72.4\scriptsize{$\pm$0.7} & 71.2\scriptsize{$\pm$1.0} & 73.7\scriptsize{$\pm$0.4} & 59.7\scriptsize{$\pm$0.0} & \red{80.7}\scriptsize{$\pm$0.3} & \red{76.7}\scriptsize{$\pm$0.2} & 84.5\scriptsize{$\pm$0.9} & \blue{93.1}\scriptsize{$\pm$0.4} \\
&f-AnoGAN  & 28.5M/6.88G & 80.8\scriptsize{$\pm$0.6} & 77.1\scriptsize{$\pm$0.4} & \blue{74.7}\scriptsize{$\pm$0.7} & \blue{72.8}\scriptsize{$\pm$0.1} & 84.4\scriptsize{$\pm$2.0} & 79.7\scriptsize{$\pm$2.0} & 80.8\scriptsize{$\pm$0.2} & 74.0\scriptsize{$\pm$0.5} & 73.9\scriptsize{$\pm$0.5} & 61.7\scriptsize{$\pm$0.5} & 42.5\scriptsize{$\pm$0.7} & 46.6\scriptsize{$\pm$0.4} & \blue{85.1}\scriptsize{$\pm$0.7} & \red{93.5}\scriptsize{$\pm$0.3} \\
& GANomaly  & 3.52M/36.6M & 80.0\scriptsize{$\pm$0.8} & 76.6\scriptsize{$\pm$1.1} & 62.9\scriptsize{$\pm$2.5} & 62.1\scriptsize{$\pm$3.4} & 89.6\scriptsize{$\pm$2.7} & 84.3\scriptsize{$\pm$4.6} & 74.4\scriptsize{$\pm$1.3} & 68.6\scriptsize{$\pm$1.7} & 63.4\scriptsize{$\pm$0.8} & 53.6\scriptsize{$\pm$0.8} & 50.2\scriptsize{$\pm$1.6} & 50.2\scriptsize{$\pm$1.2} & 59.9\scriptsize{$\pm$4.5} & 76.2\scriptsize{$\pm$3.1} \\
&AE-U      & 2.35M/30.9M & \blue{86.5}\scriptsize{$\pm$0.9} & \blue{84.4}\scriptsize{$\pm$1.1} & \red{76.9}\scriptsize{$\pm$0.7} & \red{75.4}\scriptsize{$\pm$0.3} & \blue{94.4}\scriptsize{$\pm$1.0} & \blue{89.4}\scriptsize{$\pm$1.8} & 80.8\scriptsize{$\pm$2.9} & 77.3\scriptsize{$\pm$4.4} & 70.1\scriptsize{$\pm$2.6} & 58.7\scriptsize{$\pm$2.7} & 60.6\scriptsize{$\pm$2.7} & 55.5\scriptsize{$\pm$1.0} & 83.0\scriptsize{$\pm$3.1} & 88.4\scriptsize{$\pm$1.5} \\
&DAE       & 2.79M/2.15G & 86.1\scriptsize{$\pm$0.7} & 83.3\scriptsize{$\pm$1.0} & 68.6\scriptsize{$\pm$0.9} & 71.3\scriptsize{$\pm$1.2} & 83.2\scriptsize{$\pm$0.2} & 74.6\scriptsize{$\pm$1.1} & 71.6\scriptsize{$\pm$1.0} & 65.7\scriptsize{$\pm$0.5} & 70.0\scriptsize{$\pm$0.5} & 56.0\scriptsize{$\pm$0.4} & 65.4\scriptsize{$\pm$2.2} & 60.6\scriptsize{$\pm$1.8} & \red{85.9}\scriptsize{$\pm$1.0} & \blue{93.4}\scriptsize{$\pm$0.5} \\
&AE-Grad   & 2.35M/29.9M & 73.7\scriptsize{$\pm$1.0}  & 71.0\scriptsize{$\pm$1.1} & 60.0\scriptsize{$\pm$0.3} & 62.5\scriptsize{$\pm$0.4} & 88.9\scriptsize{$\pm$0.5} & 80.4\scriptsize{$\pm$0.6} & 79.8\scriptsize{$\pm$1.5} & 77.3\scriptsize{$\pm$1.8} & 75.0\scriptsize{$\pm$0.3} & 64.1\scriptsize{$\pm$0.7} & 43.7\scriptsize{$\pm$0.7} & 44.7\scriptsize{$\pm$0.4} & 80.5\scriptsize{$\pm$0.5} & 90.4\scriptsize{$\pm$0.4} \\
&VAE-Grad{\scriptsize rec} & 2.37M/29.9M & 71.6\scriptsize{$\pm$0.8} & 69.3\scriptsize{$\pm$0.9} & 59.2\scriptsize{$\pm$1.1} & 61.9\scriptsize{$\pm$1.3} & 85.3\scriptsize{$\pm$1.8} & 77.2\scriptsize{$\pm$2.1} & 75.8\scriptsize{$\pm$2.6} & 73.0\scriptsize{$\pm$2.4} & 75.3\scriptsize{$\pm$0.1} & 64.6\scriptsize{$\pm$0.3} & 42.0\scriptsize{$\pm$0.3} & 43.3\scriptsize{$\pm$0.2} & 74.2\scriptsize{$\pm$3.6} & 84.7\scriptsize{$\pm$2.1} \\
&VAE-Grad{\scriptsize combi} & 2.37M/29.9M & 67.4\scriptsize{$\pm$0.3} & 67.3\scriptsize{$\pm$0.4} & 53.7\scriptsize{$\pm$1.0} & 56.5\scriptsize{$\pm$1.2} & 86.0\scriptsize{$\pm$0.4} & 77.4\scriptsize{$\pm$0.5} & 80.6\scriptsize{$\pm$0.3} & 78.7\scriptsize{$\pm$0.7} & 73.2\scriptsize{$\pm$0.5} & 63.0\scriptsize{$\pm$0.4} & 35.7\scriptsize{$\pm$1.9} & 39.4\scriptsize{$\pm$0.8} & 75.6\scriptsize{$\pm$0.2} & 87.5\scriptsize{$\pm$0.1} \\ \midrule
\multirow{5}{*}{\rotatebox{90}{feat-rec}}
&FAE-SSIM$^\dag$  & 11.0M/1.80G & \red{91.1}\scriptsize{$\pm$0.1} & \red{89.2}\scriptsize{$\pm$0.1} & 70.2\scriptsize{$\pm$0.6} & 68.4\scriptsize{$\pm$0.6} & 93.2\scriptsize{$\pm$0.1} & 86.3\scriptsize{$\pm$0.0} & \blue{84.6}\scriptsize{$\pm$0.4} & \blue{79.5}\scriptsize{$\pm$0.5} & \blue{77.2}\scriptsize{$\pm$0.3} & \blue{64.9}\scriptsize{$\pm$0.9} & 39.7\scriptsize{$\pm$1.3} & 41.6\scriptsize{$\pm$0.8} & 81.6\scriptsize{$\pm$0.2} & 90.3\scriptsize{$\pm$0.1} \\
&FAE-MSE$^\dag$  & 11.0M/1.80G & 74.2\scriptsize{$\pm$0.2} & 72.4\scriptsize{$\pm$0.2} & 53.1\scriptsize{$\pm$0.1} & 53.2\scriptsize{$\pm$0.1} & 83.0\scriptsize{$\pm$0.1} & 73.4\scriptsize{$\pm$0.1} & 73.7\scriptsize{$\pm$0.2} & 68.6\scriptsize{$\pm$0.2} & 74.7\scriptsize{$\pm$0.1} & 58.3\scriptsize{$\pm$0.1} & 43.7\scriptsize{$\pm$0.3} & 46.3\scriptsize{$\pm$0.2} & 78.6\scriptsize{$\pm$0.1} & 88.4\scriptsize{$\pm$0.1} \\
&RD$^\dag$    & 80.6M/33.9G & 74.7\scriptsize{$\pm$0.3} & 72.7\scriptsize{$\pm$0.5} & 57.8\scriptsize{$\pm$0.2} & 56.0\scriptsize{$\pm$0.3} & 93.9\scriptsize{$\pm$0.3} & 89.1\scriptsize{$\pm$0.9} & 67.5\scriptsize{$\pm$0.4} & 64.6\scriptsize{$\pm$0.5} & 68.3\scriptsize{$\pm$5.2} & 55.3\scriptsize{$\pm$7.3} & 59.5\scriptsize{$\pm$4.0} & 54.6\scriptsize{$\pm$1.0} & 74.8\scriptsize{$\pm$0.4} & 84.5\scriptsize{$\pm$0.4} \\
&RD++$^\dag$  & 166M/48.0G & 78.1\scriptsize{$\pm$0.6} & 75.1\scriptsize{$\pm$0.4} & 57.9\scriptsize{$\pm$0.3} & 55.5\scriptsize{$\pm$0.1} & \red{95.7}\scriptsize{$\pm$0.1} & \red{92.4}\scriptsize{$\pm$0.5} & 73.3\scriptsize{$\pm$0.4} & 68.9\scriptsize{$\pm$0.5} & 74.5\scriptsize{$\pm$1.7} & 61.9\scriptsize{$\pm$0.5} & 66.1\scriptsize{$\pm$0.5} & 59.3\scriptsize{$\pm$0.5} & 78.4\scriptsize{$\pm$1.0} & 87.7\scriptsize{$\pm$1.0} \\
&ReContrast$^\dag$    & 136M/75.3G & 79.5\scriptsize{$\pm$0.2} & 75.2\scriptsize{$\pm$0.2} & 61.4\scriptsize{$\pm$0.7} & 59.7\scriptsize{$\pm$0.6} & 93.5\scriptsize{$\pm$1.2} & 85.8\scriptsize{$\pm$3.3} & 75.2\scriptsize{$\pm$0.1} & 70.2\scriptsize{$\pm$0.3} & \red{79.2}\scriptsize{$\pm$0.6} & \red{67.5}\scriptsize{$\pm$0.3} & 60.9\scriptsize{$\pm$3.0} & 55.7\scriptsize{$\pm$2.3} & 82.4\scriptsize{$\pm$0.2} & 90.2\scriptsize{$\pm$0.2} \\ \midrule \midrule

\multicolumn{3}{l}{\textbf{Self-supervised learning AD}} \\ \midrule
\multirow{7}{*}{\rotatebox{90}{2-stage}}       
& ResNet18-IN$^\dag$  & 11.18M/1.82G & 87.0 & \blue{85.8} & \blue{67.4} & 64.5 & \blue{97.2} & \blue{95.9} & \red{80.5} & \blue{76.7} & 71.7 & 57.9 & \red{73.2} & \blue{63.2} & 72.8 & 84.1 \\
& ResNet152-IN$^\dag$ & 58.14M/11.6G & 86.1 & 84.8 & 64.3 & 61.1 & \blue{97.1} & \red{96.3} & \blue{77.9} & 74.2 & \blue{76.2} & 61.0 & \blue{73.0} & \red{64.7} & \blue{76.3} & \blue{86.4} \\
& PANDA$^\dag$        & 58.14M/11.6G & \blue{88.4}\scriptsize{$\pm$0.0} & \blue{86.6}\scriptsize{$\pm$0.0} & \blue{67.2}\scriptsize{$\pm$0.0} & \blue{64.9}\scriptsize{$\pm$0.0} & \blue{97.1}\scriptsize{$\pm$0.0} & 94.8\scriptsize{$\pm$0.0} & \red{80.5}\scriptsize{$\pm$0.0} & \red{77.5}\scriptsize{$\pm$0.0} & 75.6\scriptsize{$\pm$0.0} & 60.5\scriptsize{$\pm$0.0} & \blue{70.2}\scriptsize{$\pm$0.0} & \blue{61.5}\scriptsize{$\pm$0.0} & \red{78.2}\scriptsize{$\pm$0.0} & \red{87.9}\scriptsize{$\pm$0.0} \\
& MSC$^\dag$           & 11.18M/1.82G & \red{89.1}\scriptsize{$\pm$0.4} & \red{88.6}\scriptsize{$\pm$0.7} & 63.2\scriptsize{$\pm$0.6} & 61.9\scriptsize{$\pm$0.9} & \red{97.3}\scriptsize{$\pm$0.1} & \blue{95.6}\scriptsize{$\pm$0.3} & 77.8\scriptsize{$\pm$1.0} & \blue{75.1}\scriptsize{$\pm$1.1} & 74.2\scriptsize{$\pm$0.3} & 63.0\scriptsize{$\pm$0.2} & 58.0\scriptsize{$\pm$1.2} & 55.4\scriptsize{$\pm$1.1} & 62.5\scriptsize{$\pm$0.4} & 77.3\scriptsize{$\pm$0.4} \\
& CutPaste      & 11.18M/1.82G & 68.2\scriptsize{$\pm$1.9} & 68.0\scriptsize{$\pm$1.2} & 57.0\scriptsize{$\pm$0.8} & 56.0\scriptsize{$\pm$0.9} & 92.6\scriptsize{$\pm$1.3} & 88.1\scriptsize{$\pm$2.7} & 63.0\scriptsize{$\pm$5.8} & 60.5\scriptsize{$\pm$4.6} & 73.3\scriptsize{$\pm$2.0} & 59.4\scriptsize{$\pm$2.5} & 41.3\scriptsize{$\pm$1.3} & 46.1\scriptsize{$\pm$0.8} & 53.4\scriptsize{$\pm$0.1} & 69.7\scriptsize{$\pm$0.3} \\
& CutPaste-3way & 11.18M/1.82G & 76.3\scriptsize{$\pm$0.4} & 74.1\scriptsize{$\pm$0.6} & 60.4\scriptsize{$\pm$1.0} & 59.3\scriptsize{$\pm$1.9} & 75.5\scriptsize{$\pm$1.2} & 62.8\scriptsize{$\pm$1.4} & 60.0\scriptsize{$\pm$2.0} & 56.8\scriptsize{$\pm$1.4} & 71.9\scriptsize{$\pm$0.8} & 58.0\scriptsize{$\pm$1.8} & 42.9\scriptsize{$\pm$0.8} & 47.0\scriptsize{$\pm$0.6} & 53.5\scriptsize{$\pm$0.3} & 69.9\scriptsize{$\pm$0.2} \\
& AnatPaste     & 11.18M/1.82G & \blue{88.4}\scriptsize{$\pm$1.5} & 85.1\scriptsize{$\pm$1.7} & \red{71.0}\scriptsize{$\pm$0.6} & \red{69.3}\scriptsize{$\pm$0.9} & 95.1\scriptsize{$\pm$0.6} & 92.4\scriptsize{$\pm$1.1} & 71.8\scriptsize{$\pm$0.2} & 65.8\scriptsize{$\pm$0.3} & \red{80.7}\scriptsize{$\pm$0.5} & \red{68.1}\scriptsize{$\pm$0.5} & 56.6\scriptsize{$\pm$0.6} & 52.6\scriptsize{$\pm$0.3} & 73.7\scriptsize{$\pm$0.3} & 82.3\scriptsize{$\pm$0.2} \\  \midrule
\multirow{3}{*}{\rotatebox{90}{1-stage-c}}    
& CutPaste      & 11.51M/1.82G & 67.7\scriptsize{$\pm$1.0} & 67.6\scriptsize{$\pm$0.8} & 55.9\scriptsize{$\pm$0.5} & 56.6\scriptsize{$\pm$0.6} & 90.9\scriptsize{$\pm$0.9} & 85.1\scriptsize{$\pm$2.9} & 58.9\scriptsize{$\pm$3.1} & 57.6\scriptsize{$\pm$2.7} & 67.7\scriptsize{$\pm$1.7} & 54.5\scriptsize{$\pm$1.2} & 51.7\scriptsize{$\pm$3.5} & 50.5\scriptsize{$\pm$2.0} & 45.3\scriptsize{$\pm$1.6} & 67.9\scriptsize{$\pm$1.2} \\
& CutPaste-3way & 11.51M/1.82G & 73.0\scriptsize{$\pm$0.2} & 69.2\scriptsize{$\pm$0.5} & 60.3\scriptsize{$\pm$1.1} & 59.0\scriptsize{$\pm$1.6} & 77.0\scriptsize{$\pm$2.4} & 72.2\scriptsize{$\pm$4.3} & 57.8\scriptsize{$\pm$2.3} & 55.9\scriptsize{$\pm$1.6} & 67.3\scriptsize{$\pm$0.7} & 53.3\scriptsize{$\pm$0.9} & 54.7\scriptsize{$\pm$0.5} & 52.6\scriptsize{$\pm$0.3} & 46.1\scriptsize{$\pm$0.8} & 68.3\scriptsize{$\pm$0.3} \\
& AnatPaste     & 11.51M/1.82G & 76.9\scriptsize{$\pm$1.7} & 69.9\scriptsize{$\pm$1.7} & 65.7\scriptsize{$\pm$2.1} & 59.3\scriptsize{$\pm$0.6} & 79.1\scriptsize{$\pm$6.0} & 68.4\scriptsize{$\pm$8.7} & 67.9\scriptsize{$\pm$0.5} & 65.6\scriptsize{$\pm$0.8} & \blue{77.2}\scriptsize{$\pm$0.2} & \blue{64.4}\scriptsize{$\pm$1.2} & 56.8\scriptsize{$\pm$0.3} & 53.2\scriptsize{$\pm$0.4} & \blue{77.7}\scriptsize{$\pm$1.3} & \blue{86.1}\scriptsize{$\pm$0.8} \\  \midrule
\multirow{4}{*}{\rotatebox{90}{1-stage-s}}    
& CutPaste      & 11.49M/1.87G & 57.4\scriptsize{$\pm$2.0} & 59.9\scriptsize{$\pm$2.1} & 56.1\scriptsize{$\pm$0.2} & 56.8\scriptsize{$\pm$0.4} & 72.2\scriptsize{$\pm$2.3} & 67.0\scriptsize{$\pm$3.2} & 50.5\scriptsize{$\pm$0.6} & 52.5\scriptsize{$\pm$0.2} & 56.4\scriptsize{$\pm$1.9} & 46.0\scriptsize{$\pm$3.4} & 56.8\scriptsize{$\pm$1.0} & 55.0\scriptsize{$\pm$0.9} & 38.5\scriptsize{$\pm$0.8} & 62.2\scriptsize{$\pm$0.5} \\
& FPI           & 11.49M/1.87G & 46.3\scriptsize{$\pm$1.4} & 53.9\scriptsize{$\pm$1.0} & 46.6\scriptsize{$\pm$0.5} & 48.4\scriptsize{$\pm$0.6} & 81.5\scriptsize{$\pm$4.3} & 76.1\scriptsize{$\pm$6.2} & 50.6\scriptsize{$\pm$1.2} & 54.2\scriptsize{$\pm$1.4} & 63.4\scriptsize{$\pm$0.8} & 51.3\scriptsize{$\pm$1.0} & 51.7\scriptsize{$\pm$1.8} & 51.8\scriptsize{$\pm$1.1} & 48.8\scriptsize{$\pm$0.5} & 67.8\scriptsize{$\pm$0.1} \\
& PII           & 11.49M/1.87G & 81.9\scriptsize{$\pm$0.5} & 82.3\scriptsize{$\pm$0.4} & 65.7\scriptsize{$\pm$0.5} & \blue{65.3}\scriptsize{$\pm$0.4} & 83.5\scriptsize{$\pm$3.8} & 79.0\scriptsize{$\pm$5.1} & 59.6\scriptsize{$\pm$1.1} & 60.2\scriptsize{$\pm$1.1} & 71.9\scriptsize{$\pm$0.3} & 61.4\scriptsize{$\pm$0.3} & 44.3\scriptsize{$\pm$0.4} & 47.3\scriptsize{$\pm$0.8} & 54.3\scriptsize{$\pm$0.3} & 71.9\scriptsize{$\pm$0.4} \\
& NSA           & 11.49M/1.87G & 82.4\scriptsize{$\pm$0.5} & 83.8\scriptsize{$\pm$0.5} & 62.0\scriptsize{$\pm$0.5} & 63.4\scriptsize{$\pm$0.5} & 82.1\scriptsize{$\pm$1.7} & 78.8\scriptsize{$\pm$2.2} & 66.0\scriptsize{$\pm$0.2} & 66.5\scriptsize{$\pm$0.2} & 73.5\scriptsize{$\pm$0.7} & \blue{63.4}\scriptsize{$\pm$1.1} & 45.6\scriptsize{$\pm$0.7} & 47.2\scriptsize{$\pm$0.5} & 53.5\scriptsize{$\pm$5.3} & 72.3\scriptsize{$\pm$2.8} \\ \midrule
\bottomrule
\end{tabular}
}
\end{table*}

\begin{table}[!t]
\centering
\caption{Performance on pixel-level AnoSeg. In each paradigm, the \red{best} results are highlighted in red, while the \blue{second-} and \blue{third-} best results are highlighted in blue. img-rec: image-reconstruction; feat-rec: feature-reconstruction; 1-stage-s: one-stage segmentation. $^\dag$Methods utilize ImageNet pre-trained weights.}  \label{tab:ano_seg}
\begin{tabular}{llcc}
\toprule
& \multirow{2}{*}{ Method } & \multicolumn{2}{c}{ BraTS2021 } \\
\cmidrule(l){3-4} 
& & $\text{AP}_\text{pix}$ & $\lceil$Dice$\rceil$ \\ \midrule
\rowcolor{gray!20} & Intensity & 63.0 & 58.7  \\ \midrule  \midrule
\multicolumn{3}{l}{\textbf{Reconstruction AD}} \\ \midrule
 \multirow{15}{*}{\rotatebox{90}{img-rec}} & AE          & 33.2\scriptsize{$\pm$2.2} & 39.2\scriptsize{$\pm$1.7}  \\
& AE-$\ell_1$     & 26.5\scriptsize{$\pm$3.7} & 34.4\scriptsize{$\pm$3.1}  \\
& AE-SSIM     & 25.7\scriptsize{$\pm$0.4} & 35.7\scriptsize{$\pm$0.4}  \\
& AE-PL$^\dag$& 44.8\scriptsize{$\pm$0.4} & 45.2\scriptsize{$\pm$0.2}  \\
& AE[$d=4$]   & 33.4\scriptsize{$\pm$5.4} & 40.0\scriptsize{$\pm$4.2}  \\
& VAE         & 44.0\scriptsize{$\pm$5.3} & 47.1\scriptsize{$\pm$3.4}  \\
& Constrained AE & 30.9\scriptsize{$\pm$3.9} & 37.5\scriptsize{$\pm$3.1}  \\
& MemAE       & 22.8\scriptsize{$\pm$5.6} & 30.8\scriptsize{$\pm$4.8}  \\
& CeAE        & 28.6\scriptsize{$\pm$3.2} & 35.8\scriptsize{$\pm$2.4}  \\
& AnoDDPM\tiny{Gaussian}   & 10.0\scriptsize{$\pm$0.1} & 18.1\scriptsize{$\pm$0.2}  \\
& AnoDDPM\tiny{Simplex}    & \blue{53.4}\scriptsize{$\pm$0.1} & 51.5\scriptsize{$\pm$0.2}  \\
& AutoDDPM    & \blue{60.1}\scriptsize{$\pm$3.5} & \blue{58.0}\scriptsize{$\pm$2.5}  \\
& f-AnoGAN    & 23.7\scriptsize{$\pm$1.4} & 28.5\scriptsize{$\pm$2.1}  \\
& AE-U        & 22.2\scriptsize{$\pm$4.0} & 35.9\scriptsize{$\pm$4.4}  \\
& DAE         & \red{75.5}\scriptsize{$\pm$0.7} & \red{71.1}\scriptsize{$\pm$0.6}  \\
& AE-Grad     & 29.9\scriptsize{$\pm$1.0} & 36.6\scriptsize{$\pm$0.5}  \\
& VAE-Grad{\scriptsize rec}   & 35.6\scriptsize{$\pm$1.2} & 42.1\scriptsize{$\pm$0.6}  \\
& VAE-Grad{\scriptsize combi} & 24.3\scriptsize{$\pm$1.2} & 32.9\scriptsize{$\pm$1.5}  \\ \midrule
\multirow{5}{*}{\rotatebox{90}{feat-rec}} 
& FAE-SSIM$^\dag$    & 50.4\scriptsize{$\pm$0.6} & \blue{52.8}\scriptsize{$\pm$0.3}  \\
& FAE-MSE$^\dag$     & 40.5\scriptsize{$\pm$1.6} & 44.8\scriptsize{$\pm$1.3}  \\
& RD$^\dag$          & 28.1\scriptsize{$\pm$0.5} & 35.9\scriptsize{$\pm$0.4}  \\
& RD++$^\dag$        & 28.7\scriptsize{$\pm$0.7} & 36.8\scriptsize{$\pm$0.7}  \\
& ReContrast$^\dag$  & 43.0\scriptsize{$\pm$1.7} & 45.7\scriptsize{$\pm$1.1}  \\ \midrule \midrule

\multicolumn{3}{l}{\textbf{Self-supervised learning AD}} \\ \midrule
\multirow{4}{*}{\rotatebox{90}{1-stage-s}}& CutPaste      & 10.0\scriptsize{$\pm$1.2} & 18.2\scriptsize{$\pm$1.5} \\
                       & FPI           & 6.8\scriptsize{$\pm$0.0} & 15.9\scriptsize{$\pm$0.1}  \\
                       & PII           & 11.3\scriptsize{$\pm$0.5} & 18.2\scriptsize{$\pm$0.4}  \\
                       & NSA           & \red{14.9}\scriptsize{$\pm$4.6} & \red{23.9}\scriptsize{$\pm$5.3} \\ \bottomrule
\end{tabular}
\end{table}

\subsection{Comparison of the state-of-the-arts}
For image-level AnoCls, we observe from Table~\ref{tab:ano_cls} that AE-PL, AE-U, and FAE-SSIM usually achieve SOTA performance among the reconstruction methods. Among the SSL methods, those that utilize or fine-tune ImageNet pre-trained weights are the most competitive. Notably, most of these SOTA methods, except for AE-U, leverage ImageNet pre-trained weights in some way. 
\textbf{These results demonstrate the effectiveness of ImageNet weights for medical AD and highlight their versatility in enhancing AD through various means, including distance measurement (e.g., AE-PL), input data transformation (e.g., feature-reconstruction methods), and direct feature extraction (e.g., two-stage SSL methods).}


\textbf{In the absence of the ImageNet weights, reconstruction methods outperform SSL methods in image-level AnoCls}. When excluding the use of ImageNet weights, the SOTA SSL method is two-stage AnatPaste. However, this method exhibits unstable performance across datasets. It fails to outperform the vanilla AE on LAG and BraTS2021 datasets, indicating a high sensitivity to abnormal patterns. In contrast, AE-U and DAE, which are the SOTA reconstruction methods that do not use ImageNet weights, demonstrate stable and competitive performance on most datasets. 

Notably, \textbf{DDPM-based methods demonstrate especially high performance on histopathology images such as Camelyon16}. As depicted in Table~\ref{tab:ano_cls}, when ImageNet weights are excluded, most AE- or GAN-based methods do not work at all on Camelyon16. Only AE-U and DAE achieve an AUC exceeding 51\%. Conversely, DDPM-based methods like AutoDDPM and AnoDDPM{\tiny Simplex} achieve remarkable AUCs of 80.7\% and 70.6\% on Camelyon16, respectively, surpassing others by a significant margin. This phenomenon can be attributed to the close relationship between anomalies in histopathology images and cellular morphology. Consequently, models with robust reconstruction capabilities (e.g., DAE, AnoDDPM{\tiny Simplex}, and AutoDDPM) or feature extraction prowess (e.g., AE-PL, FAE, and ReContrast) are desirable to repair image-space morphological changes or recognize feature-space deviations. This hypothesis finds support in Fig.~\ref{fig:examples} and Table~\ref{tab:ano_cls}.

\textbf{For pixel-level AnoSeg, reconstruction methods significantly outperform SSL methods}. As shown in Table~\ref{tab:ano_seg}, the vanilla AE beats all SSL one-stage segmentation methods in terms of the segmentation metrics on BraTS2021. This can be attributed to the fact that SSL methods trained through synthetic data segmentation rely on the similarity between synthetic abnormal patterns and real anomalies. They struggle to recognize real abnormal regions when these regions exhibit significant differences in appearance compared to synthetic anomalies. Consequently, their performance deteriorates on such datasets. This point will be further discussed in Section~\ref{sec:ssl_cls}. 

\textbf{The best AnoSeg performance is achieved by DAE, surpassing the second-best method by a margin of over 10\% $\lceil$Dice$\rceil$.} This remarkable performance can be attributed to the synergy between the customized UNet architecture and the proposed self-supervised denoising task. This collaboration enables the meticulous reconstruction of healthy regions while effectively eliminating disease regions. Specifically, the UNet, equipped with skip connections, preserves normal details for high-quality reconstruction. However, skip connections inadvertently introduce abnormal information from the input into the reconstruction, leading to false negatives \citep{mao2020abnormality}. The denoising task in DAE effectively mitigates this drawback of the UNet by optimizing the model to generate a clean version of the corrupted input image. As a result, the model is encouraged to repair corrupted content, such as anomalies, in input images, facilitating the generation of pseudo-healthy reconstructions.

\subsection{Analysis of reconstruction methods}
\subsubsection{Effects of network architectures in AE}
While some papers have conducted analyses on certain aspects of network architectures in AE, such analyses still remain incomplete. For instance, \citet{bercea2023aes} demonstrated that increasing the layer depth of AE tends to learn the data distribution and leads to blurry reconstructions. However, their analysis primarily focuses on reconstruction quality rather than its impact on AD performance. Additionally, the experiments on a single image modality may restrict the generalizability of conclusions. Furthermore, the effects of some essential configurations, such as network width and input size, have not been studied. As a complement to prior studies, we aim to provide a comprehensive analysis of effects of AE architectures on AD performance, based on our extensive experiments on diverse datasets.
Specifically, we conduct experiments using AE with varying input sizes $H \times W$, block depths $D$, and basic widths $C_0$. Their computational cost, represented by the number of network parameters and FLOPs, is summarized in Table~\ref{tab:cost}, while the performance of these architectures on the collected datasets is presented in Tables~\ref{tab:size}, \ref{tab:block_depth}, and \ref{tab:basic_width}. 

\begin{table}[!t]
\centering
\caption{Number of network parameters (\#Params) and FLOPs of AE with different hyperparameter settings.} \label{tab:cost}
\begin{tabular}{llll}
\toprule 
Modification                    & Setting           & \#Params      & FLOPs     \\ \midrule 
\rowcolor{gray!20} -            & Default           & 2.35M         & 29.9M     \\ \midrule 
Input size                      & $128 \times 128$  & 8.65M         & 119.5M    \\ \midrule 
\multirow{2}{*}{Block depth}    & $D=2$             & 2.52M         & 45.5M     \\
                                & $D=3$             & 2.69M         & 61.0M     \\ \midrule 
\multirow{2}{*}{Basic width}    & $C_0=32$          & 5.09M         & 112.2M    \\
                                & $C_0=64$          & 11.84M        & 434.0M    \\ \midrule
\multirow{9}{*}{Latent size}    & $d=128$           & 2.58M         & 30.1M     \\
                                & $d=64$            & 2.45M         & 30.0M     \\
                                & $d=32$            & 2.38M         & 29.9M     \\
                                & $d=8$             & 2.33M         & 29.9M     \\
                                & $d=4$             & 2.33M         & 29.9M     \\
                                & $d=2$             & 2.32M         & 29.9M     \\
                                & $d=1$             & 2.32M         & 29.9M     \\
                                & $\textbf{z} \in \mathbb{R}^{2\times4\times4}$     & 0.22M     & 27.9M     \\
                                & $\textbf{z} \in \mathbb{R}^{1\times4\times4}$     & 0.22M     & 27.9M     \\
                                \bottomrule
\end{tabular}
\end{table}

\begin{table*}[!t]
\centering
\caption{Performance of AE with different input size. $64\times64$ is the baseline. [d] indicates that the latent size is the same as the baseline, while [r] indicates that the compression rate is the same as the baseline.} \label{tab:size}

\resizebox{\linewidth}{!}{
\begin{tabular}{ccccccccccccccccc}
\toprule 
\multirow{2}{*}{\makecell[c]{Input size \\ $H \times W$}} & \multicolumn{2}{c}{ RSNA } & \multicolumn{2}{c}{ VinDr-CXR } & \multicolumn{2}{c}{ Brain Tumor } & \multicolumn{2}{c}{ LAG } & \multicolumn{2}{c}{ISIC2018} & \multicolumn{2}{c}{Camelyon16} & \multicolumn{4}{c}{ BraTS2021 }\\
\cmidrule(l){2-3}\cmidrule(l){4-5} \cmidrule(l){6-7} \cmidrule(l){8-9} \cmidrule(l){10-11} \cmidrule(l){12-13} \cmidrule(l){14-17}
                 & AUC & AP & AUC & AP & AUC & AP & AUC & AP & AUC & AP & AUC & AP & AUC & AP & $\text{AP}_\text{pix}$ & $\lceil$Dice$\rceil$ \\ \midrule
$64 \times 64$   & 67.5\scriptsize{$\pm$0.9} & 66.7\scriptsize{$\pm$0.5} & 56.4\scriptsize{$\pm$0.4} & 60.2\scriptsize{$\pm$0.5} & 85.9\scriptsize{$\pm$0.3} & 76.9\scriptsize{$\pm$0.1} & 79.0\scriptsize{$\pm$1.0} & 75.7\scriptsize{$\pm$1.4} & 74.5\scriptsize{$\pm$0.8} & 63.0\scriptsize{$\pm$0.6} & 36.8\scriptsize{$\pm$1.2} & 40.3\scriptsize{$\pm$0.6} & 82.6\scriptsize{$\pm$0.1} & 92.0\scriptsize{$\pm$0.1} & 33.2\scriptsize{$\pm$2.2} & 39.2\scriptsize{$\pm$1.7}\\
$128 \times 128${\scriptsize [d]} & 66.7\scriptsize{$\pm$0.2} & 65.7\scriptsize{$\pm$0.1} & 54.7\scriptsize{$\pm$0.4} & 58.2\scriptsize{$\pm$0.2} & 85.6\scriptsize{$\pm$0.5} & 75.5\scriptsize{$\pm$0.5} & 78.2\scriptsize{$\pm$2.5} & 75.4\scriptsize{$\pm$2.1} & 74.4\scriptsize{$\pm$0.1} & 62.3\scriptsize{$\pm$0.1} & 35.3\scriptsize{$\pm$0.6} & 39.3\scriptsize{$\pm$0.2} & 79.5\scriptsize{$\pm$0.4} & 90.1\scriptsize{$\pm$0.2} & 26.2\scriptsize{$\pm$1.6} & 35.5\scriptsize{$\pm$1.3} \\
$128 \times 128${\scriptsize [r]} & 60.6\scriptsize{$\pm$0.6} & 60.2\scriptsize{$\pm$0.3} & 48.4\scriptsize{$\pm$0.1} & 51.0\scriptsize{$\pm$0.4} & 85.6\scriptsize{$\pm$0.4} & 75.5\scriptsize{$\pm$0.7} & 71.5\scriptsize{$\pm$1.6} & 67.9\scriptsize{$\pm$1.4} & 75.0\scriptsize{$\pm$0.0} & 61.4\scriptsize{$\pm$0.3} & 30.2\scriptsize{$\pm$0.1} & 37.2\scriptsize{$\pm$0.0} & 78.0\scriptsize{$\pm$0.4} & 88.9\scriptsize{$\pm$0.3} & 19.6\scriptsize{$\pm$1.6} & 29.7\scriptsize{$\pm$1.8} \\

\bottomrule
\end{tabular}
}
\end{table*}

\begin{table*}[!t]
\centering
\caption{Performance of AE with different block depth.} \label{tab:block_depth}

\resizebox{\linewidth}{!}{
\begin{tabular}{ccccccccccccccccc}
\toprule
\multirow{2}{*}{\makecell[c]{Block depth \\ $D$}} & \multicolumn{2}{c}{ RSNA } & \multicolumn{2}{c}{ VinDr-CXR } & \multicolumn{2}{c}{ Brain Tumor } & \multicolumn{2}{c}{ LAG } & \multicolumn{2}{c}{ISIC2018} & \multicolumn{2}{c}{Camelyon16} & \multicolumn{4}{c}{ BraTS2021 } \\
\cmidrule(l){2-3}\cmidrule(l){4-5} \cmidrule(l){6-7} \cmidrule(l){8-9} \cmidrule(l){10-11} \cmidrule(l){12-13} \cmidrule(l){14-17}
  & AUC & AP & AUC & AP & AUC & AP & AUC & AP & AUC & AP & AUC & AP & AUC & AP & $\text{AP}_\text{pix}$ & $\lceil$Dice$\rceil$ \\ \midrule
 1 & 67.5\scriptsize{$\pm$0.9} & 66.7\scriptsize{$\pm$0.5} & 56.4\scriptsize{$\pm$0.4} & 60.2\scriptsize{$\pm$0.5} & 85.9\scriptsize{$\pm$0.3} & 76.9\scriptsize{$\pm$0.1} & 79.0\scriptsize{$\pm$0.9} & 75.7\scriptsize{$\pm$1.4} & 74.5\scriptsize{$\pm$0.8} & 63.0\scriptsize{$\pm$0.6} & 36.8\scriptsize{$\pm$1.2} & 40.3\scriptsize{$\pm$0.6} & 82.6\scriptsize{$\pm$0.1} & 92.0\scriptsize{$\pm$0.1} & 33.2\scriptsize{$\pm$2.2} & 39.2\scriptsize{$\pm$1.7}  \\
 2 & 68.8\scriptsize{$\pm$0.5} & 67.7\scriptsize{$\pm$0.8} & 55.5\scriptsize{$\pm$0.7} & 59.5\scriptsize{$\pm$0.5} & 84.6\scriptsize{$\pm$0.3} & 75.5\scriptsize{$\pm$0.3} & 78.5\scriptsize{$\pm$0.7} & 75.3\scriptsize{$\pm$0.7} & 74.4\scriptsize{$\pm$0.4} & 63.1\scriptsize{$\pm$0.4} & 36.7\scriptsize{$\pm$0.6} & 40.3\scriptsize{$\pm$0.4} & 81.7\scriptsize{$\pm$1.0} & 91.4\scriptsize{$\pm$0.6} & 29.5\scriptsize{$\pm$8.3} & 36.8\scriptsize{$\pm$6.6}  \\
 3 & 67.1\scriptsize{$\pm$0.6} & 66.0\scriptsize{$\pm$0.6} & 56.6\scriptsize{$\pm$0.7} & 60.0\scriptsize{$\pm$0.3} & 84.5\scriptsize{$\pm$0.4} & 75.1\scriptsize{$\pm$0.6} & 82.0\scriptsize{$\pm$1.7} & 78.7\scriptsize{$\pm$2.5} & 74.2\scriptsize{$\pm$0.2} & 63.1\scriptsize{$\pm$0.2} & 36.4\scriptsize{$\pm$0.3} & 40.1\scriptsize{$\pm$0.1} & 81.2\scriptsize{$\pm$0.6} & 91.1\scriptsize{$\pm$0.4} & 28.2\scriptsize{$\pm$3.7} & 36.5\scriptsize{$\pm$3.4}\\
\bottomrule
\end{tabular}

}
\end{table*}

\begin{table*}[!t]
\centering
\caption{Performance of AE with different basic width.} \label{tab:basic_width}
\resizebox{\linewidth}{!}{
\begin{tabular}{ccccccccccccccccc}
\toprule
\multirow{2}{*}{\makecell[c]{Basic width \\ $C_0$}} & \multicolumn{2}{c}{ RSNA } & \multicolumn{2}{c}{ VinDr-CXR } & \multicolumn{2}{c}{ Brain Tumor } & \multicolumn{2}{c}{ LAG } & \multicolumn{2}{c}{ISIC2018} & \multicolumn{2}{c}{Camelyon16} & \multicolumn{4}{c}{ BraTS2021 } \\
\cmidrule(l){2-3}\cmidrule(l){4-5} \cmidrule(l){6-7} \cmidrule(l){8-9} \cmidrule(l){10-11} \cmidrule(l){12-13} \cmidrule(l){14-17}
                 & AUC & AP & AUC & AP & AUC & AP & AUC & AP & AUC & AP & AUC & AP & AUC & AP & $\text{AP}_\text{pix}$ & $\lceil$Dice$\rceil$ \\ \midrule
 16 & 67.5\scriptsize{$\pm$0.9} & 66.7\scriptsize{$\pm$0.5} & 56.4\scriptsize{$\pm$0.4} & 60.2\scriptsize{$\pm$0.5} & 85.9\scriptsize{$\pm$0.3} & 76.9\scriptsize{$\pm$0.1} & 79.0\scriptsize{$\pm$1.0} & 75.7\scriptsize{$\pm$1.4} & 74.5\scriptsize{$\pm$0.8} & 63.0\scriptsize{$\pm$0.6} & 36.8\scriptsize{$\pm$1.2} & 40.3\scriptsize{$\pm$0.6} & 82.6\scriptsize{$\pm$0.1} & 92.0\scriptsize{$\pm$0.1} & 33.2\scriptsize{$\pm$2.2} & 39.2\scriptsize{$\pm$1.7}  \\
 32 & 68.9\scriptsize{$\pm$0.6} & 67.7\scriptsize{$\pm$0.8} & 56.9\scriptsize{$\pm$0.4} & 60.5\scriptsize{$\pm$0.6} & 88.8\scriptsize{$\pm$0.2} & 79.3\scriptsize{$\pm$0.3} & 79.5\scriptsize{$\pm$1.6} & 76.4\scriptsize{$\pm$1.3} & 74.7\scriptsize{$\pm$0.2} & 63.4\scriptsize{$\pm$0.2} & 36.5\scriptsize{$\pm$0.4} & 40.1\scriptsize{$\pm$0.2} & 82.4\scriptsize{$\pm$0.1} & 91.8\scriptsize{$\pm$0.1} & 29.8\scriptsize{$\pm$2.3} & 36.9\scriptsize{$\pm$1.7} \\
 64 & 68.1\scriptsize{$\pm$0.5} & 66.8\scriptsize{$\pm$0.5} & 56.4\scriptsize{$\pm$0.2} & 60.3\scriptsize{$\pm$0.2} & 89.2\scriptsize{$\pm$0.2} & 79.6\scriptsize{$\pm$0.3} & 77.1\scriptsize{$\pm$2.7} & 73.5\scriptsize{$\pm$2.7} & 74.4\scriptsize{$\pm$0.1} & 63.2\scriptsize{$\pm$0.3} & 36.0\scriptsize{$\pm$0.6} & 39.7\scriptsize{$\pm$0.2} & 82.3\scriptsize{$\pm$0.4} & 91.8\scriptsize{$\pm$0.2} & 34.7\scriptsize{$\pm$5.9} & 40.7\scriptsize{$\pm$4.4} \\
\bottomrule
\end{tabular}

}
\end{table*}

\begin{table*}[!t]
\centering
\caption{Performance of AE with different latent size. The best two results are marked in bold and underlined.} \label{tab:latent}

\resizebox{\linewidth}{!}{
\begin{tabular}{ccccccccccccccccc}
\toprule
\multirow{2}{*}{\makecell[c]{Latent size \\ $d$}} & \multicolumn{2}{c}{ RSNA } & \multicolumn{2}{c}{ VinDr-CXR } & \multicolumn{2}{c}{ Brain Tumor } & \multicolumn{2}{c}{ LAG } & \multicolumn{2}{c}{ISIC2018} & \multicolumn{2}{c}{Camelyon16} & \multicolumn{4}{c}{ BraTS2021 } \\
\cmidrule(l){2-3}\cmidrule(l){4-5} \cmidrule(l){6-7} \cmidrule(l){8-9} \cmidrule(l){10-11} \cmidrule(l){12-13} \cmidrule(l){14-17}
                 & AUC & AP & AUC & AP & AUC & AP & AUC & AP & AUC & AP & AUC & AP & AUC & AP & $\text{AP}_\text{pix}$ & $\lceil$Dice$\rceil$ \\ \midrule
 128 & 60.5\scriptsize{$\pm$0.6} & 60.3\scriptsize{$\pm$0.4} & 48.6\scriptsize{$\pm$0.2} & 51.9\scriptsize{$\pm$0.3} & 86.5\scriptsize{$\pm$0.3} & 76.9\scriptsize{$\pm$0.4} & 76.0\scriptsize{$\pm$1.2} & 72.0\scriptsize{$\pm$0.4} & \underline{75.8}\scriptsize{$\pm$0.4} & 62.8\scriptsize{$\pm$0.4} & 30.1\scriptsize{$\pm$0.2} & 37.2\scriptsize{$\pm$0.1} & 80.5\scriptsize{$\pm$0.9} & 90.5\scriptsize{$\pm$0.7} & 25.6\scriptsize{$\pm$2.9} & 32.7\scriptsize{$\pm$2.3}     \\
 64  & 60.7\scriptsize{$\pm$0.6} & 60.7\scriptsize{$\pm$0.6} & 49.0\scriptsize{$\pm$0.2} & 52.0\scriptsize{$\pm$0.3} & \underline{86.8}\scriptsize{$\pm$0.1} & \underline{77.2}\scriptsize{$\pm$0.2} & 73.5\scriptsize{$\pm$1.4} & 69.9\scriptsize{$\pm$1.2} & 75.4\scriptsize{$\pm$0.1} & 62.7\scriptsize{$\pm$0.1} & 31.7\scriptsize{$\pm$0.4} & 37.8\scriptsize{$\pm$0.2} & 81.3\scriptsize{$\pm$0.2} & 91.0\scriptsize{$\pm$0.2} & 20.8\scriptsize{$\pm$3.1} & 28.8\scriptsize{$\pm$2.8}  \\
 32  & 62.8\scriptsize{$\pm$0.4} & 62.5\scriptsize{$\pm$0.2} & 52.5\scriptsize{$\pm$0.4} & 55.6\scriptsize{$\pm$0.5} & \textbf{87.1}\scriptsize{$\pm$0.2} & \textbf{77.8}\scriptsize{$\pm$0.2} & 75.1\scriptsize{$\pm$2.7} & 71.0\scriptsize{$\pm$3.3} & 75.5\scriptsize{$\pm$0.3} & 63.1\scriptsize{$\pm$0.3} & 34.6\scriptsize{$\pm$0.4} & 39.1\scriptsize{$\pm$0.1} & \underline{82.1}\scriptsize{$\pm$0.3} & \underline{91.6}\scriptsize{$\pm$0.1} & 25.2\scriptsize{$\pm$3.8} & 32.7\scriptsize{$\pm$3.3}  \\
 16  & 67.5\scriptsize{$\pm$0.9} & 66.7\scriptsize{$\pm$0.5} & 56.4\scriptsize{$\pm$0.4} & 60.2\scriptsize{$\pm$0.5} & 85.9\scriptsize{$\pm$0.3} & 76.9\scriptsize{$\pm$0.1} & 79.0\scriptsize{$\pm$1.0} & 75.7\scriptsize{$\pm$1.4} & 74.5\scriptsize{$\pm$0.8} & 63.0\scriptsize{$\pm$0.6} & 36.8\scriptsize{$\pm$1.2} & 40.3\scriptsize{$\pm$0.6} & \textbf{82.6}\scriptsize{$\pm$0.1} & \textbf{92.0}\scriptsize{$\pm$0.1} & 33.2\scriptsize{$\pm$2.2} & 39.2\scriptsize{$\pm$1.7}  \\
 8   & \underline{69.4}\scriptsize{$\pm$1.2} & \underline{68.3}\scriptsize{$\pm$0.8} & 57.4\scriptsize{$\pm$0.3} & 61.7\scriptsize{$\pm$0.3} & 84.1\scriptsize{$\pm$0.6} & 75.5\scriptsize{$\pm$0.8} & 80.8\scriptsize{$\pm$2.2} & 77.1\scriptsize{$\pm$2.5} & 74.5\scriptsize{$\pm$0.3} & \underline{63.6}\scriptsize{$\pm$0.5} & 36.1\scriptsize{$\pm$0.4} & 40.3\scriptsize{$\pm$0.2} & 81.0\scriptsize{$\pm$0.5} & 90.9\scriptsize{$\pm$0.3} & \textbf{36.1}\scriptsize{$\pm$7.9} & \textbf{42.3}\scriptsize{$\pm$5.9} \\
 4   & \textbf{72.9}\scriptsize{$\pm$2.1} & \textbf{70.3}\scriptsize{$\pm$0.9} & \textbf{61.3}\scriptsize{$\pm$0.6} & \textbf{64.1}\scriptsize{$\pm$0.8} & 70.8\scriptsize{$\pm$1.4} & 64.1\scriptsize{$\pm$1.0} & \textbf{83.3}\scriptsize{$\pm$1.0} & \textbf{78.9}\scriptsize{$\pm$0.7} & 67.8\scriptsize{$\pm$0.2} & 57.1\scriptsize{$\pm$0.5} & 34.5\scriptsize{$\pm$0.1} & 39.7\scriptsize{$\pm$0.1} & 79.4\scriptsize{$\pm$0.1} & 89.6\scriptsize{$\pm$0.3} & 33.4\scriptsize{$\pm$5.4} & 40.0\scriptsize{$\pm$4.2} \\
 2   & 67.5\scriptsize{$\pm$0.8} & 65.4\scriptsize{$\pm$1.0} & \underline{60.5}\scriptsize{$\pm$0.3} & \underline{59.7}\scriptsize{$\pm$0.2} & 47.3\scriptsize{$\pm$1.7} & 48.3\scriptsize{$\pm$1.0} & \underline{81.3}\scriptsize{$\pm$2.6} & \underline{77.5}\scriptsize{$\pm$3.2} & 64.0\scriptsize{$\pm$0.1} & 53.9\scriptsize{$\pm$0.5} & 34.5\scriptsize{$\pm$0.4} & 40.0\scriptsize{$\pm$0.2} & 77.2\scriptsize{$\pm$0.3} & 87.5\scriptsize{$\pm$0.1} & 30.0\scriptsize{$\pm$2.5} & 38.0\scriptsize{$\pm$1.7} \\
 1   & 66.5\scriptsize{$\pm$0.4} & 63.6\scriptsize{$\pm$0.9} & 59.4\scriptsize{$\pm$1.6} & 58.6\scriptsize{$\pm$1.6} & 36.6\scriptsize{$\pm$0.3} & 42.4\scriptsize{$\pm$0.6} & 68.4\scriptsize{$\pm$1.5} & 63.6\scriptsize{$\pm$1.5} & 61.0\scriptsize{$\pm$0.5} & 51.2\scriptsize{$\pm$0.3} & 33.9\scriptsize{$\pm$0.7} & 40.3\scriptsize{$\pm$0.2} & 72.8\scriptsize{$\pm$0.8} & 84.5\scriptsize{$\pm$0.7} & \underline{35.9}\scriptsize{$\pm$5.1} & \underline{40.8}\scriptsize{$\pm$3.4}  \vspace{4pt} \\

 $\mathbf{z} \in \mathbb{R}^{2\times4\times4}$   & 54.8\scriptsize{$\pm$1.4} & 56.3\scriptsize{$\pm$0.9} & 49.1\scriptsize{$\pm$0.2} & 51.8\scriptsize{$\pm$0.0} & 64.1\scriptsize{$\pm$0.7} & 58.3\scriptsize{$\pm$0.5} & 69.8\scriptsize{$\pm$0.2} & 67.3\scriptsize{$\pm$0.9} & \textbf{76.6}\scriptsize{$\pm$0.2} & \textbf{64.0}\scriptsize{$\pm$0.4} & 31.7\scriptsize{$\pm$0.1} & 37.8\scriptsize{$\pm$0.0} & 81.3\scriptsize{$\pm$0.4} & 91.2\scriptsize{$\pm$0.2} & 27.5\scriptsize{$\pm$4.2} & 34.6\scriptsize{$\pm$3.1}  \\
 $\mathbf{z} \in \mathbb{R}^{1\times4\times4}$   & 68.1\scriptsize{$\pm$1.7} & 67.7\scriptsize{$\pm$1.1} & 56.9\scriptsize{$\pm$2.3} & 57.7\scriptsize{$\pm$1.7} & 55.1\scriptsize{$\pm$1.0} & 52.6\scriptsize{$\pm$1.1} & 73.6\scriptsize{$\pm$2.9} & 68.0\scriptsize{$\pm$2.1} & 73.8\scriptsize{$\pm$0.7} & 62.6\scriptsize{$\pm$0.7} & 31.6\scriptsize{$\pm$0.3} & 37.9\scriptsize{$\pm$0.1} & 81.0\scriptsize{$\pm$0.1} & 90.8\scriptsize{$\pm$0.0} & 30.1\scriptsize{$\pm$2.9} & 36.8\scriptsize{$\pm$2.3} \\
\bottomrule
\end{tabular}
}
\end{table*}

Surprisingly, we observe from Table~\ref{tab:size} that using an input size of $64 \times 64$ achieves comparable performance to using a larger input size, such as $128 \times 128$. These results suggest that current AE architectures for AD may be insufficient in exploiting intricate details present in high-resolution images, even when the latent size is increased to maintain a consistent compression rate. Consequently, employing higher resolutions for current AE architectures, as commonly done in prevailing methods \citep{baur2021autoencoders}, may be unnecessary. Exploring methods to enhance the capability of AEs in capturing finer details for AD becomes a problem that warrants careful consideration.


Overall, these experimental results demonstrate that \textbf{current AE architectures for medical AD does not benefit from large resolutions or large networks}. This perspective on reconstruction-based AD suggests that increasing the complexity of the model may be unnecessary. It naturally raises a question of how the reconstruction methods effectively work for AD. This question will be further discussed in the following Section~\ref{subsec:latent}.

\subsubsection{Effects of latent space restrictions in AE} \label{subsec:latent}

Reconstruction AE-based AD methods map the input data into a condensed latent space and then reconstruct the input accordingly. The latent space acts as an information bottleneck, potentially impeding the propagation of abnormal information and, consequently, obstructing the reconstruction of anomalies. 
While previous methods have incorporated various forms of latent space restriction, such as KL-divergence in VAE \citep{baur2021autoencoders} or memory modules in MemAE \citep{gong2019memorizing}, none of these studies systematically compare and analyze the effects of these strategies. Furthermore, the most straightforward approach for latent space restriction, namely latent dimension reduction, remains unexplored. To address these issues, we explicitly investigate effects of different restrictions imposed on the latent space using our curated datasets. This investigation aims to examine their effects and shed light on the underlying principles of AE-based methods.

To begin, we discuss latent dimension reduction. Table~\ref{tab:latent} shows that \textbf{a remarkably compact latent size (e.g., 4--32) yields a substantial performance boost when applied to datasets containing local regional anomalies} (i.e., RSNA, VinDr-CXR, Brain Tumor, LAG, and BraTS2021 datasets, as mentioned in Section~\ref{subsec:data_summary}). For example, decreasing the latent size from 128 to 4 improves the evaluation metrics by 12.4\% AUC and 10.0\% AP on RSNA dataset, 12.7\% AUC and 12.2\% AP on VinDr-CXR dataset, and 7.3\% AUC and 6.9\% AP on LAG dataset. In the case of brain MRI modalities, specifically the Brain Tumor and BraTS2021 datasets, a similar trend is observed, albeit with a larger optimal latent size of approximately 32 and 16, respectively. 

Based on these results, we make two main conclusions. \textbf{Firstly, a narrow enough bottleneck in reconstruction networks is preferred to prevent anomaly reconstruction in the detection of local anomalies}. As mentioned in Section~\ref{subsec:data_summary}, abnormal samples with local anomalies exhibit unexpected changes in some areas compared to normal samples while still retaining most normal regions. An appropriately compact latent size allows AE to represent the normal variations and restrict its generalization ability, rendering it incapable of representing abnormal changes. If the latent size is too large, the AE could exhibit undesired generalization on unseen abnormal changes, leading to false negatives.

\textbf{Secondly, the optimal latent size differs across datasets, with datasets in which the normal data exhibits more variations (i.e., is more informative) tending to prefer a larger latent size.} As mentioned above, the optimal latent size of brain MRI datasets is larger compared to other datasets. This disparity can be attributed to the fact that MRIs offer more information content compared to 2D scans. MRIs are volumetric scans that capture detailed tissue information, exhibiting greater variations among healthy subjects and encompassing variations among axial slices. These characteristics enable MRI datasets to surpass the information content of other datasets, necessitating a larger latent size to adequately represent them. 

One unexpected trend in the results is the performance on ISIC2018 dataset. As shown in Table~\ref{tab:latent}, contrary to the aforementioned datasets that favor relatively small latent sizes, ISIC2018 demonstrates improved performance with increasing latent sizes. The potential reason of this phenomenon is the fact that ISIC2018 contains global semantic anomalies that encompass entirely different categories compared to the normal data. Unlike images with local anomalies, which primarily involve regional unexpected variations, abnormal images in ISIC2018 introduce entirely new subject categories that do not overlap with the normal ones. These samples are relatively far out-of-distribution (OOD). Consequently, increasing the latent size may be ineffective for enabling the network to generalize well on these far OOD samples, while still facilitating the reconstruction of normal samples. This outcome makes ISIC2018 dataset benefit from large latent sizes. However, Camelyon16, another dataset in our experiments that contains global anomalies, is so difficult that AEs with any latent sizes fail on this dataset. Consequently, we cannot derive conclusive evidence from Camelyon16 to validate our hypothesis based on experiments on ISIC2018 dataset. The property of global semantic anomalies still requires further exploration.

Furthermore, we explore the performance of the spatial latent size, where the bottleneck comprises $1 \times 1$ convolutional layers instead of flattening and fully-connected layers. This allows the latent code to retain spatial shapes, such as $\mathbf{z} \in \mathbb{R}^{1\times4\times4}$ or $\mathbf{z} \in \mathbb{R}^{2\times4\times4}$. \textbf{While the spatial AE significantly reduces the number of parameters compared to the dense AE, the performance may decrease on certain datasets.} Specifically, both the dense AE with $\mathbf{z} \in \mathbb{R}^{16}$ and the spatial AE with $\mathbf{z} \in \mathbb{R}^{1\times4\times4}$ have the same number of latent variables. However, the dense AE ($\mathbf{z} \in \mathbb{R}^{16}$) has 2.35M parameters, whereas the spatial AE ($\mathbf{z} \in \mathbb{R}^{1\times4\times4}$) only has 0.22M parameters. Regarding performance, the spatial version suffers from a significant drop on Brain Tumor and LAG datasets. This phenomenon can be attributed to the fact that the spatial bottleneck compresses different regions to a consistent degree, regardless of the varying amounts of information present in these regions.

\begin{figure*}[!t]
\centering

\subfigure[RSNA - 0.48/0.53]{
\begin{minipage}[t]{0.2\linewidth}
\centering
\includegraphics[width=\linewidth]{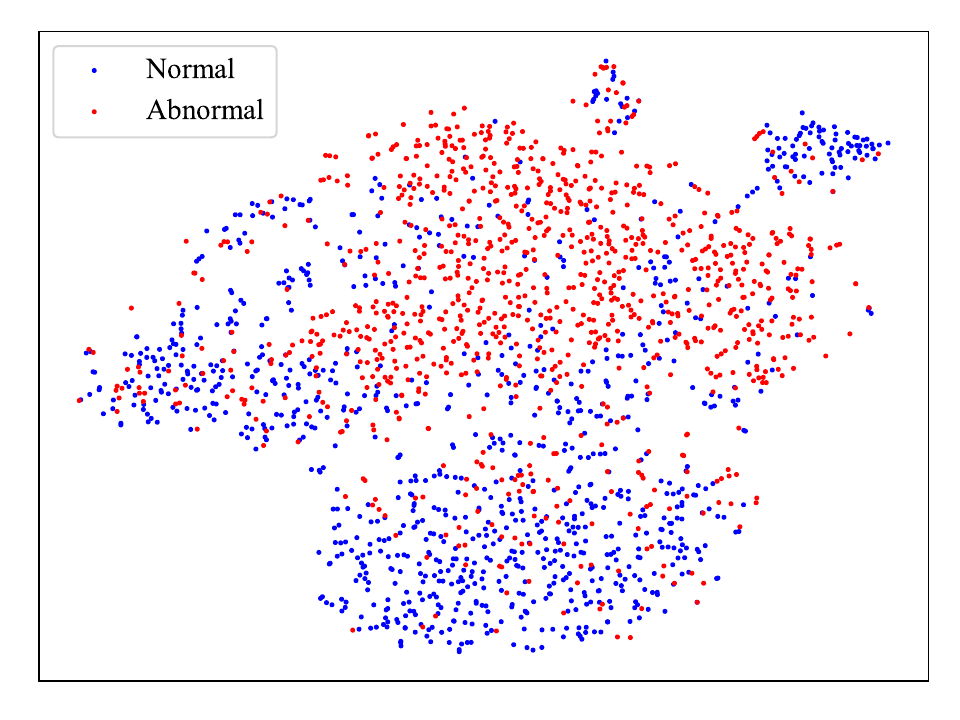}
\end{minipage}
}
\subfigure[VinDrCXR - 0.40/0.47]{
\begin{minipage}[t]{0.2\linewidth}
\centering
\includegraphics[width=\linewidth]{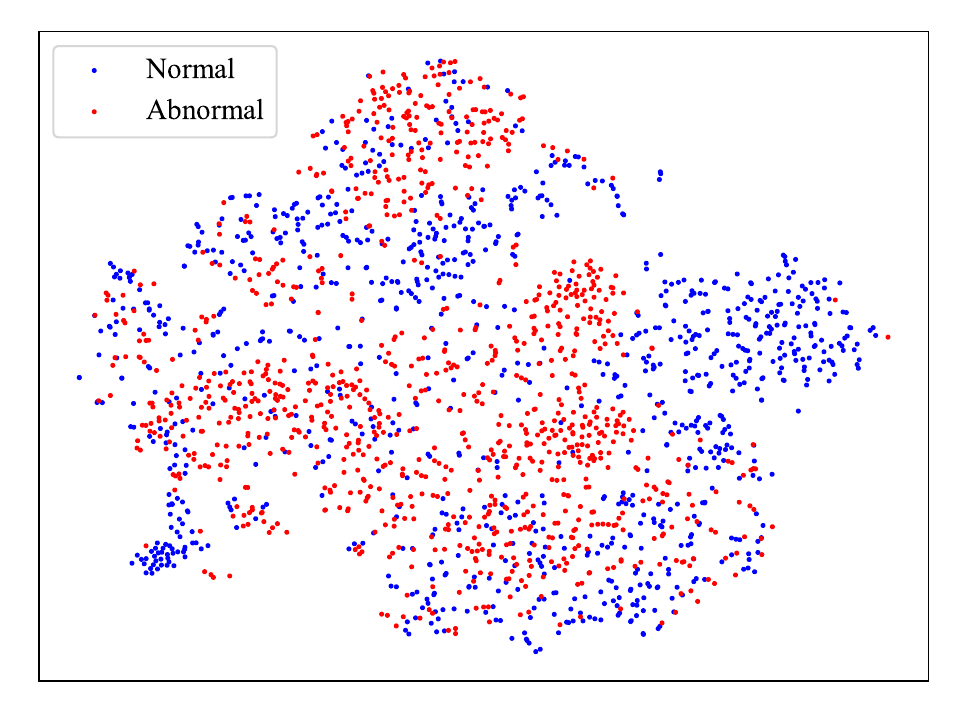}
\end{minipage}
}
\subfigure[Brain Tumor - 0.22/0.35]{
\begin{minipage}[t]{0.2\linewidth}
\centering
\includegraphics[width=\linewidth]{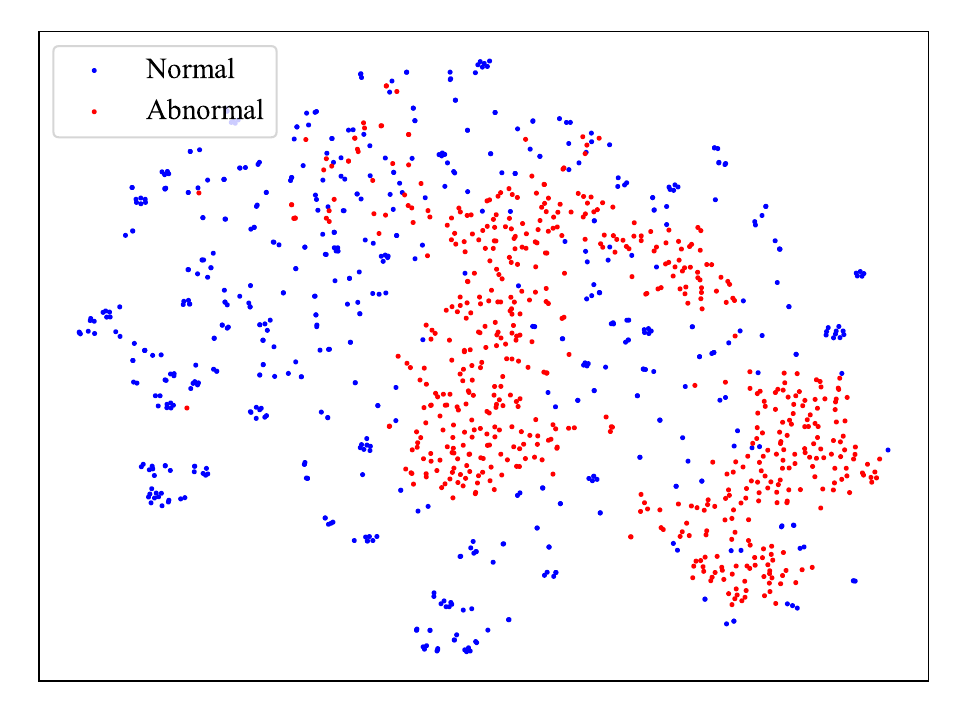}
\end{minipage}
}
\subfigure[LAG - 0.64/0.60]{
\begin{minipage}[t]{0.2\linewidth}
\centering
\includegraphics[width=\linewidth]{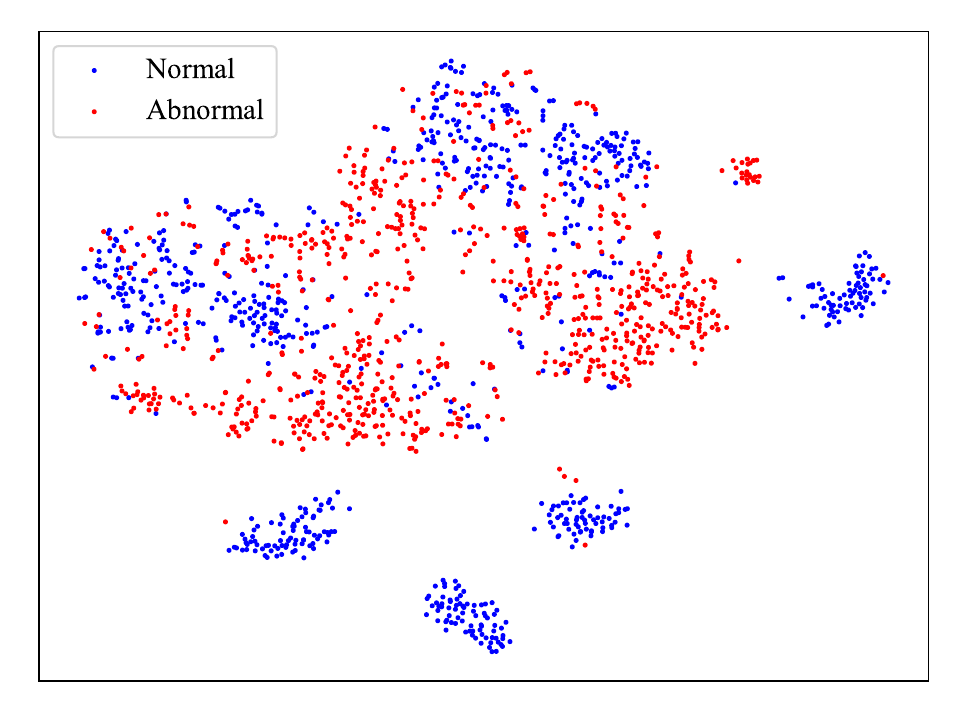}
\end{minipage}
}

\subfigure[ISIC2018 - 0.60/0.51]{
\begin{minipage}[t]{0.2\linewidth}
\centering
\includegraphics[width=\linewidth]{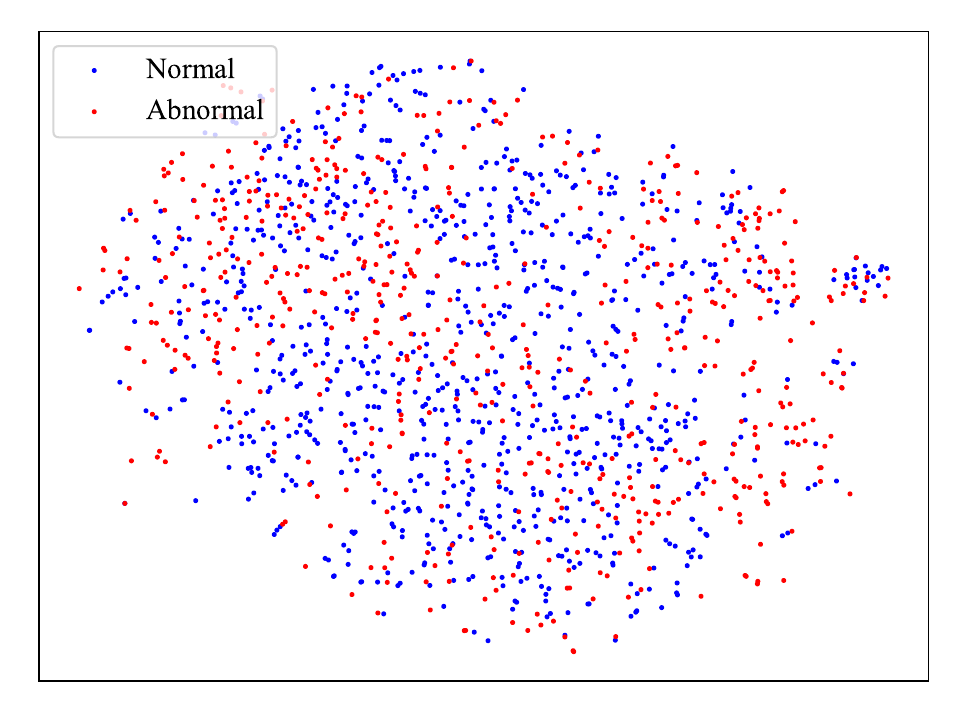}
\end{minipage}
}
\subfigure[Camelyon16 - 0.49/0.50]{
\begin{minipage}[t]{0.2\linewidth}
\centering
\includegraphics[width=\linewidth]{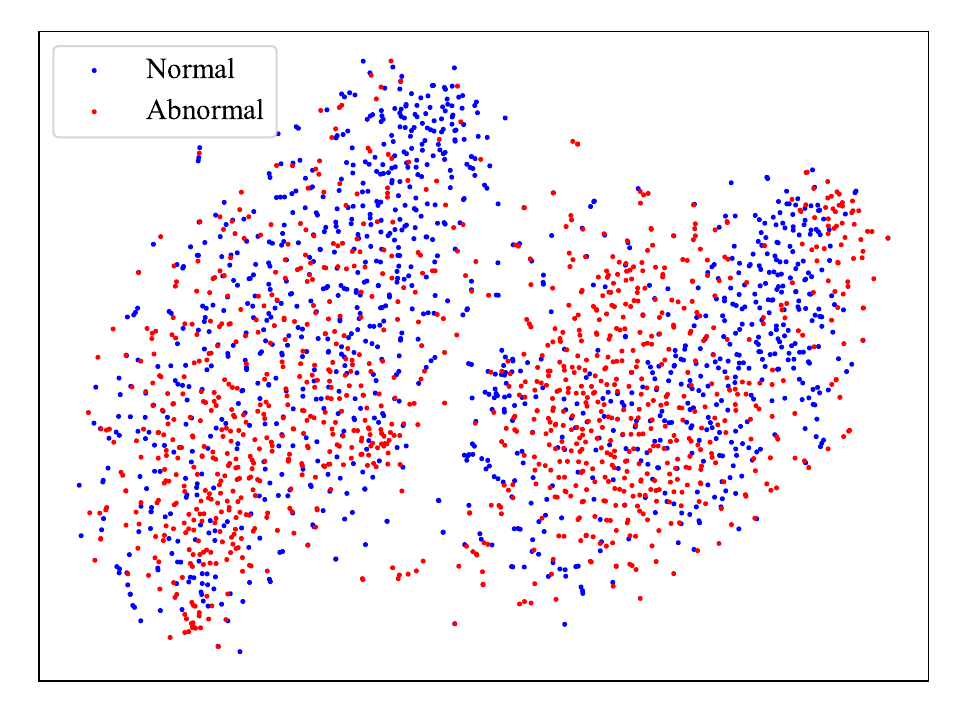}
\end{minipage}
}
\subfigure[BraTS2021 - 0.66/0.82]{
\begin{minipage}[t]{0.2\linewidth}
\centering
\includegraphics[width=\linewidth]{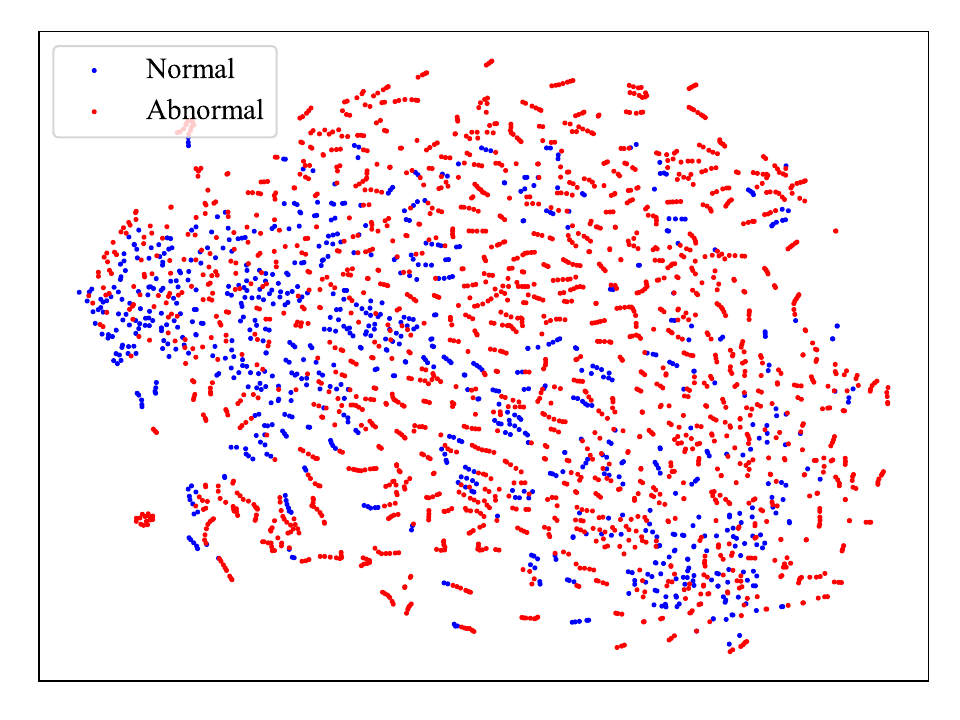}
\end{minipage}
}
\caption{T-SNE visualization for the latent representations of AE on the seven datasets. The performance of OC-SVM built on the latent representations is presented in the format AUC/AP.} \label{fig:tsne_latent}
\end{figure*}

In addition to the simplest latent dimension reduction, there exist other popular designs that impose latent space restrictions in various ways, such as VAE \citep{kingma2013auto}, Constrained AE \citep{chen2018unsupervised}, and MemAE \citep{gong2019memorizing}. However, Table~\ref{tab:ano_cls} shows that \textbf{none of the existing approaches for restricting the latent space demonstrate superior performance compared to latent dimension reduction}. Potential reasons are that VAE samples the latent code from the learnt distribution instead of using a deterministic vector, making the reconstruction indeterministic, while Constrained AE does not explicitly restrict the latent space.
The unexpected failure of MemAE \citep{gong2019memorizing}, despite its success on various natural images, raises questions. MemAE introduced a memory bank to store prototypes of normal latent features during training and utilized a combination of these prototypes to replace the latent feature of the input. It hypothesized a distinct difference between normal and abnormal features, aiming to render abnormal inputs unreconstructible using the combination of normal prototypes. However, this hypothesis is not necessarily valid due to the inherent nature of AE \citep{bercea2023aes,cai2024rethinking}. Consequently, \textbf{we attribute the failure of MemAE to the inseparability of latent features of normal and abnormal medical images}. To substantiate our argument, we visualize the latent features of normal and abnormal medical images obtained from AE and present the performance of OC-SVM built on these features, in Fig.~\ref{fig:tsne_latent}. The results illustrate that these features are mixed together and inseparable by OC-SVM. This finding contradicts the underlying hypothesis of MemAE and provides support for our argument, ultimately explaining the failure of MemAE in medical scenarios.

\subsubsection{Effects of distance functions for the reconstruction error in AE} \label{subsec:distance}
In addition to the simplest $\ell_2$ and $\ell_1$ distances \citep{baur2021autoencoders}, various distance functions have been explored to measure the reconstruction error for anomaly detection \citep{bergmann2018improving,behrendt2024diffusion,shvetsova2021anomaly,bercea2023generalizing}. For example, \citet{bergmann2018improving,behrendt2024diffusion} employed Structural Similarity Index Measure (SSIM) \citep{wang2004image}, while \citet{shvetsova2021anomaly,bercea2023generalizing} utilized perceptual loss (PL) \citep{johnson2016perceptual}. However, these studies combine specific distance functions with other methodological designs, and they do not explicitly compare the effects of different distance functions on various types of images. Consequently, the characteristics of different distance functions and the principles for selecting the most suitable ones for reconstruction-based AD remain unclear. To address these questions, we explicitly investigate the effects of these distance functions by employing the same AE architecture but with different reconstruction error functions. The performance of AE ($\ell_2$), AE-$\ell_1$, AE-SSIM and AE-PL is presented in Tables~\ref{tab:ano_cls} and \ref{tab:ano_seg}.

Surprisingly, we observe a significant impact of the distance function on the performance of reconstruction-based AD. While the standard $\ell_2$ and $\ell_1$ losses exhibit comparable performance on the majority of datasets, SSIM and PL losses present remarkable outcomes. Specifically, replacing the default $\ell_2$ loss with SSIM results in an improvement of over 10\% AUC on RSNA dataset, slightly worse performance on VinDr-CXR, and a drop of more than 10\% AUC on LAG dataset. Notably, AE-PL demonstrates SOTA or comparable performance on six datasets: RSNA, VinDr-CXR, Brain Tumor, LAG, Camelyon16, and BraTS2021, surpassing the $\ell_2$ version by over 20\% AUC on certain datasets. However, it performs worse than the $\ell_2$ loss on ISIC2018. These observations suggest that \textbf{distance functions for measuring the reconstruction error play a substantial role in reconstruction-based AD, and their performance is closely related to the abnormal patterns present in the datasets}. Therefore, developing a suitable distance function to measure the reconstruction error of anomalies in different scenarios is a promising direction.

\subsubsection{Effects of the gradient strategy in AE}
When comparing methods that use gradient-based anomaly scores in AE-based backbones, such as AE-Grad and VAE-$\text{Grad}_\text{rec}$, with those using original anomaly score, namely AE and VAE, we observe that \textbf{gradient-based scores usually exhibit slightly better performance than the original reconstruction error}. This finding could inspire us to aggregate the global information into scalars and use the gradient w.r.t. the input as prediction maps. However, VAE-Grad$_{combi}$, which incorporates the KL-term, performs worse than VAE-Grad$_{rec}$. This raises a question regarding how to properly exploit the extra information as a complement to reconstruction errors through the gradient strategy.

\subsubsection{Effects of noise types in DDPM}
Current DDPM-based AD methods rely on the addition of noise during the diffusion process to corrupt abnormal regions, thereby facilitating pseudo-healthy reconstructions. \citet{Wyatt_2022_CVPR} demonstrated the superiority of their proposed Simplex noise over the original Gaussian noise used in DDPM for corrupting abnormal regions, showcasing better performance on brain MRIs. We evaluate the performance of AnoDDPM \citep{Wyatt_2022_CVPR} equipped with these two noise types on a wider range of datasets. Our findings align with their conclusions and provide additional insights into the limitations of DDPM-based AD methods stemming from their reliance on noise.

Table~\ref{tab:ano_seg} presents that AnoDDPM{\tiny Simplex} outperforms its Gaussian noise counterpart on most datasets, particularly on BraTS2021, where it achieves a remarkable margin of over 30\% $\lceil$Dice$\rceil$. Visual evidence provided in Fig.~\ref{fig:visualization} demonstrates that AnoDDPM{\tiny Simplex} successfully repairs most abnormal regions in chest X-rays and brain MRIs, while AnoDDPM{\tiny Gaussian} reconstructs numerous undesirable lesions. These findings establish the superiority of Simplex noise in corrupting abnormal regions across various image modalities.

However, the performance of DDPM-based methods on LAG and ISIC2018 datasets reveals limitations inherent to these methods. In both cases, the vanilla AE outperforms DDPM-based methods. The potential reason behind this observation is that \textbf{DDPM-based methods heavily rely on the similarity between the noise type employed and the abnormal patterns to ensure the corruption of abnormal regions during the diffusion process}. This characteristic introduces a strong bias that renders these methods unsuitable for anomalies that differ from the noise type. Despite attempts to integrate healthy context into the repair process, such as in AutoDDPM \citep{bercea2023mask}, the performance on LAG and ISIC2018 datasets still falls short compared to the performance of the vanilla AE. Consequently, this unresolved issue restricts the generalizability of DDPM-based methods for AD, making them less versatile than AE-based methods.

\subsubsection{Reconstruction quality and AD performance}
\noindent
\textbf{Correlation.} Previously, \citet{baur2021autoencoders} demonstrated a relatively weak correlation between reconstruction quality and AnoSeg performance on brain MRIs. Instead, it is the distinguishable residual histograms of normal and anomalous intensities that matter most. They observed that in certain cases, as the reconstruction error increases (i.e., the reconstruction quality decreases), the AnoSeg performance actually improves. Building upon their findings and conducting experiments on a wider range of datasets and methods, we extend their conclusion to encompass additional data modalities and provide further insights to certain datasets. 

Expanding on the conclusion drawn by \citet{baur2021autoencoders}, we show that the relatively weak correlation between reconstruction quality and AD performance is also applicable to other modalities such as chest X-rays, retinal fundus images, and dermatoscopic images. A prominent example supporting this conclusion is the comparison between AnoDDPM{\tiny Gaussian} and AE-PL. As depicted in Fig.~\ref{fig:visualization}, AnoDDPM{\tiny Gaussian} achieves the highest reconstruction quality on all datasets, effectively preserving both normal and abnormal regions. However, as highlighted in Tables~\ref{tab:ano_cls} and \ref{tab:ano_seg}, AnoDDPM{\tiny Gaussian} exhibits the poorest overall performance. This can be attributed to the ineffectiveness of Gaussian noise in corrupting low-frequency abnormal regions, leading to their well-reconstruction alongside normal regions. Conversely, despite generating visually ``faulty" reconstructions, AE-PL demonstrates competitive performance across most datasets. The reason is that perceptual loss performs anomaly detection in the feature space, prioritizing high-level semantic information over intensity reconstruction.

It is noteworthy that some recent works \citep{bercea2023mask,liang2024itermask} achieve both high reconstruction quality and competitive AD performance. These studies concentrate on enhancing the reconstruction quality of normal areas, rather than the entire image, and employ strategies such as masking to degrade the reconstruction quality of abnormal regions. Therefore, they still support the conclusion that distinguishable residual histograms of normal and anomalous intensities are more crucial than the overall reconstruction quality.

Additionally, we reveal a complementary insight concerning the detection of specific abnormal patterns, such as tumor cells in histopathology images (Camelyon16), where high reconstruction quality can yield benefits. Combining Table~\ref{tab:ano_cls} and Fig.~\ref{fig:visualization}, we observe that, excluding methods equipped with ImageNet weights, the top three performing methods on Camelyon16 -- AutoDDPM, AnoDDPM{\tiny Simplex}, and DAE -- consistently produce high-fidelity reconstructions. We attribute this observation to the distinctive abnormal patterns present in histopathology images, where tumor cells manifest morphological deviations compared to normal cells. High reconstruction quality is beneficial for recognizing these anomalies related to cellular morphology.

\begin{figure*}[!t]
\centering
\includegraphics[width=\linewidth]{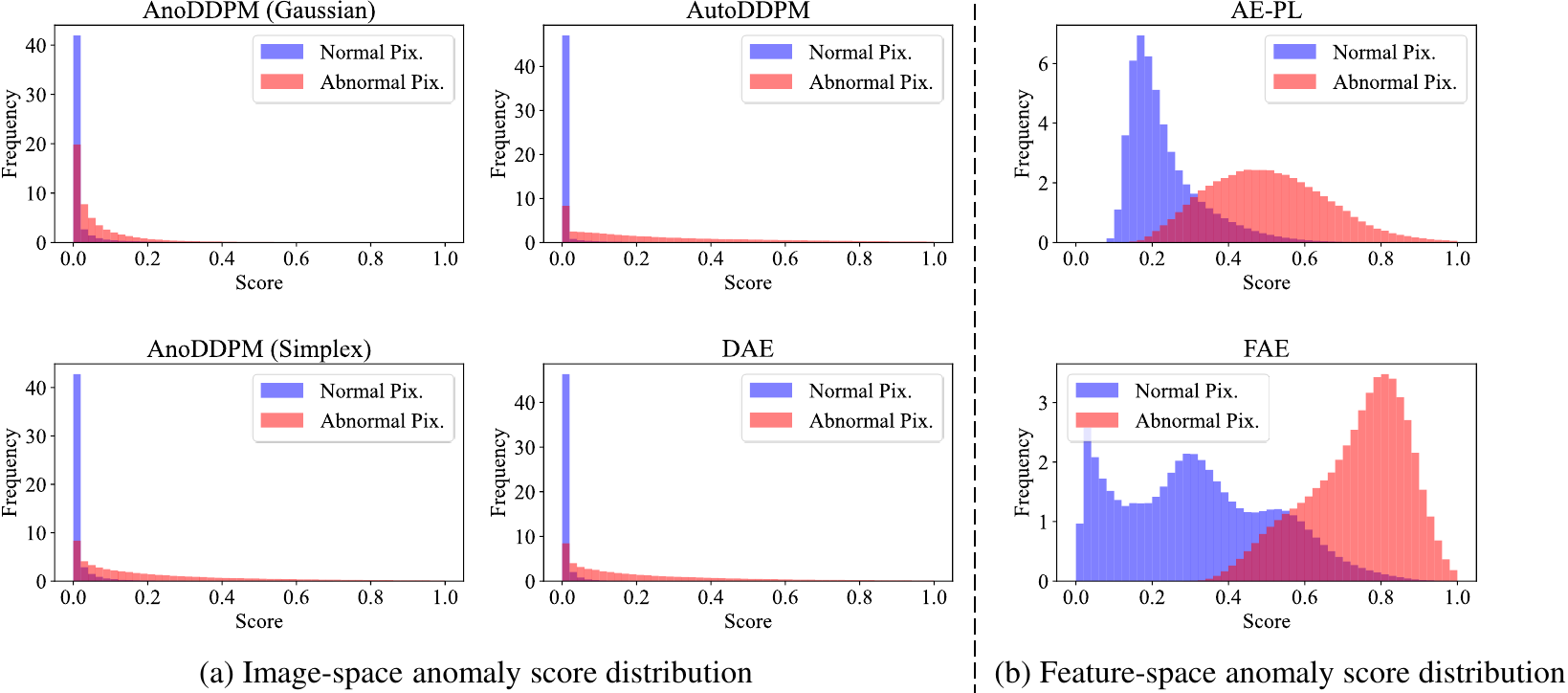}
\caption{Distribution of pixel-level anomaly scores calculated in image-space and feature-space. Scores are normalized through division by the maximum value. Note the substantial difference in the vertical scale of (a) and (b).}
\label{fig:histogram}
\end{figure*}

\noindent
\textbf{Characteristics of relevant methods.} To investigate the disparity between reconstruction quality and performance metrics, we began from analyzing the binary prediction maps depicted in Fig.~\ref{fig:visualization} (the third column of BraTS2021 dataset). The binary prediction maps illustrate that methods such as DAE and autoDDPM, known for their high-quality reconstruction, excel in segmenting small tumors. However, their prediction maps contain scattered noisy points on normal regions. In contrast, methods such as AE-PL, which generate coarse score maps from feature-space, tend to produce larger segmentation regions than actual abnormal areas but exhibit improved robustness in normal regions without introducing noisy points. These characteristics could be relevant to the calculation space (i.e., image-space and feature-space) of their anomaly scores. 

For further examination, we visualized the distribution of pixel-level anomaly scores for typical methods in Fig.~\ref{fig:histogram}. Initially, Fig.~\ref{fig:histogram} shows that image-space anomaly scores are concentrated around zero, reflecting high reconstruction quality of corresponding methods. Improvements in these methods should prioritize amplifying errors in abnormal regions to reduce indistinguishable anomalies. For instance, autoDDPM exhibits notably lower frequencies of abnormal histogram overlapping with normal ones near zero compared to AnoDDPM (Gaussian), consistent with its superior performance metrics. In contrast to the lower values of image-space anomaly scores, feature-space scores from AE-PL and FAE exhibit values distant from zero on both normal and abnormal pixels, with distinguishable scores between the two, explaining their coarse score maps yet high performance metrics.

Analyzing these characteristics is crucial for clinical practice. However, existing quantitative metrics may not fully capture their distinctions to reflect clinical utility. For instance, widely used metrics like Dice score favor overconfident predictions, thus benefiting methods inclined towards over-segmentation like AE-PL and FAE. Furthermore, while feature-space anomaly scores from FAE and AE-PL distinguish between normal and abnormal pixels and achieve satisfactory performance metrics, they pose a challenge in clinical scenarios where discerning intensity differences in non-zero positive scores without proper calibration can be difficult for human experts. Future efforts should develop metrics and calibration strategies to effectively evaluate and enhance the clinical utility of anomaly segmentation.

\subsection{Analysis of SSL methods}

The performance of SSL-based methods on AnoCls and AnoSeg is presented in Tables~\ref{tab:ano_cls} and \ref{tab:ano_seg}, respectively. To facilitate a clear discussion, we categorize these methods based on whether they utilize ImageNet weights. Methods that utilize ImageNet weights are denoted with ``$^\dag$" in the tables.

\subsubsection{Methods without ImageNet weights}  \label{sec:ssl_cls}
SSL methods without ImageNet weights mainly rely on training on synthetic data. From their results, we can draw two significant conclusions as follows. 

Firstly, comparing methods with the same number of stages, we conclude that \textbf{more realistic synthetic anomalies usually lead to better performance in SSL-based methods}. For example, AnatPaste \citep{sato2023anatomy} is specifically designed for chest X-rays and generates more realistic lung lesions than CutPaste. 
As a result, in the two-stage paradigm, AnatPaste outperforms CutPaste on the two chest X-ray datasets by a large margin, 20.2\% AUC and 17.1\% AP on RSNA dataset, and 14.0\% AUC and 13.3\% AP on VinDr-CXR dataset. A similar trend is also observed in the one-stage classification paradigm, where AnatPaste outperforms CutPaste on both RSNA and VirDr-CXR datasets. Additionally, in the one-stage segmentation mode, NSA \citep{schluter2022natural} and PII \citep{tan2021detecting} synthesize more realistic anomalies using Poisson image editing, and thereby consistently outperform CutPaste and FPI on all datasets. 

Secondly, comparing the one-stage and two-stage paradigms using the same anomaly-synthesis method, we conclude that \textbf{in SSL-based AD, the two-stage paradigm usually outperforms the one-stage paradigm}. Specifically, anomaly-synthesis approaches can be employed for both paradigms. While these two paradigms share a training procedure that classifies normal and synthetic data, the two-stage paradigm has an extra procedure to remove the projection head and build a Gaussian density estimator on learned representations. 
Results show that when using AnatPaste for self-supervised training, the two-stage paradigm outperforms the one-stage classification paradigm on four datasets: by 11.5\% AUC and 15.2\% AP on RSNA dataset, 5.3\% AUC and 10\% AP on VinDr-CXR dataset, 16.0\% AUC and 24.0\% AP on Brain Tumor dataset, and 3.9\% AUC and 0.2\% AP on LAG dataset. A consistent trend also appears when using CutPaste or CutPaste-3way. 


\begin{figure}[!b]
\centering
\includegraphics[width=\linewidth]{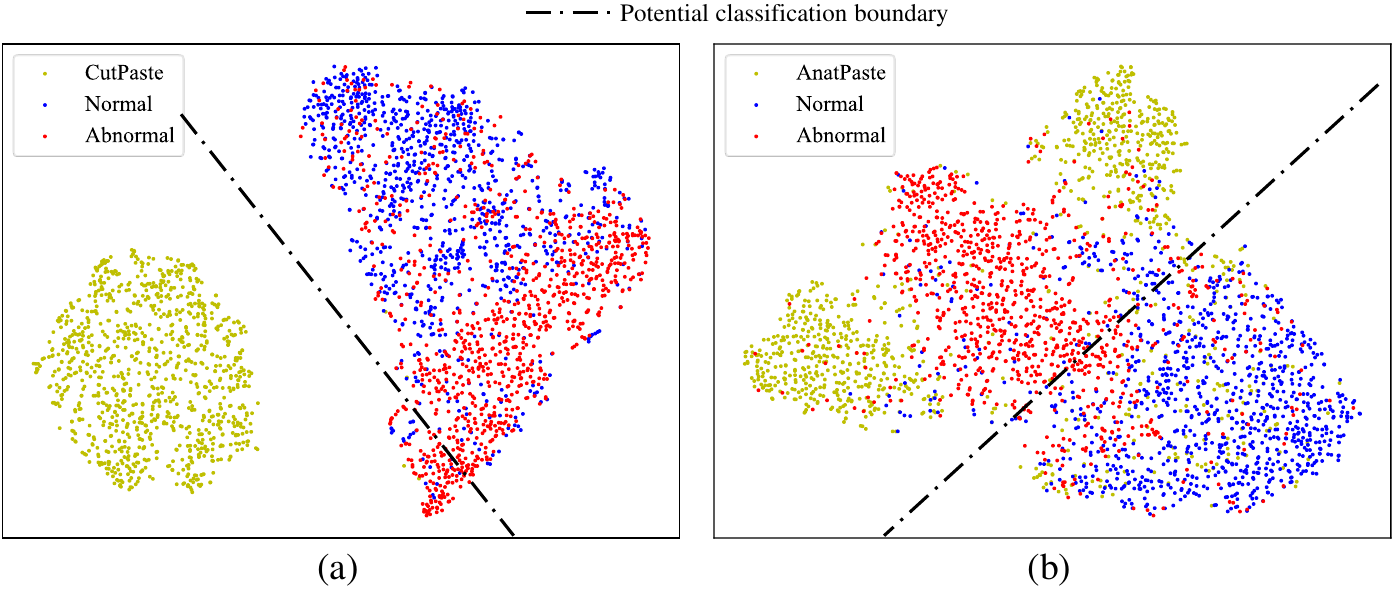}
\caption{T-SNE visualization of representations learned through synthetic data classification. (a) CutPaste; (b) AnatPaste.}
\label{fig:tsne_ssl}
\end{figure}

To elucidate the rationale behind these conclusions, we conduct t-SNE visualization of representations learned through AnatPaste and CutPaste. For the first conclusion (the efficacy of realistic synthesis), we compare the visualization of the two typical synthesis approaches. As depicted in Fig.~\ref{fig:tsne_ssl}(a), a model trained through CutPaste can easily differentiate between the normal data and the unrealistic CutPaste samples. However, due to the significant discrepancy between real anomalies and synthetic ones, this model struggles to distinguish real anomalies from normal data. In contrast, Fig.~\ref{fig:tsne_ssl}(b) illustrates that the model trained on realistic synthetic data can naturally recognize real anomalies by leveraging the classification training. This observation underscores the importance of the realism of synthetic data. For the second conclusion (the superiority of the two-stage paradigm), let us consider the example provided in Fig.~\ref{fig:tsne_ssl}(a). The visualization highlights that although the classifier trained on unrealistic synthetic data fails to recognize real anomalies, the training process enables the model to effectively distinguish between the features of normal and abnormal samples. As a result, the two-stage paradigm, which removes the classification head and utilizes the learned representations, outperforms the one-stage paradigm, which directly employs the subpar classification head.

\subsubsection{Methods with ImageNet weights}
Most SSL-based AD methods that utilize ImageNet weights were initially developed for industrial scenarios \citep{li2021cutpaste,reiss2021panda,reiss2023mean}. While some studies have leveraged ImageNet weights for medical AD \citep{sato2023anatomy,lagogiannis2023unsupervised}, they primarily utilized these weights for model initialization only. Consequently, the performance of ImageNet weights themselves in medical AD remains unclear, leading to potential misconceptions regarding their utility. To explicitly evaluate the performance of ImageNet weights in medical AD and assess the effectiveness of relevant SSL methods, we conducted evaluations not only on the existing methods but also on the vanilla ImageNet-pretrained ResNet18 and ResNet152 feature extractors without special designs. This allows us to uncover the true impact of ImageNet weights on medical AD and shed light on the effectiveness of existing methodologies.

Surprisingly, we observe from Table~\ref{tab:ano_cls} that \textbf{directly utilizing a fixed ImageNet pre-trained network to extract representations for one-class classifiers outperforms most prevailing SSL methods that incorporate more complex training processes}. Specifically, ImageNet pre-trained ResNet18, employed under the two-stage mode, achieves top-3 performance on nearly all datasets. This finding highlights the effectiveness of ImageNet weights in medical AD, which has not received sufficient attention and exploration. While some approaches, such as PANDA \citep{reiss2021panda} and MSC \citep{reiss2023mean}, attempt to adapt ImageNet weights for AD, the experimental results suggest that they do not significantly improve performance compared to directly utilizing these weights. Therefore, adapting the powerful ImageNet weights for medical AD remains a promising but unresolved problem.

\section{Challenges and future directions}  \label{sec:discussion}

\begin{figure}[!b]
\centering
\includegraphics[width=0.7\linewidth]{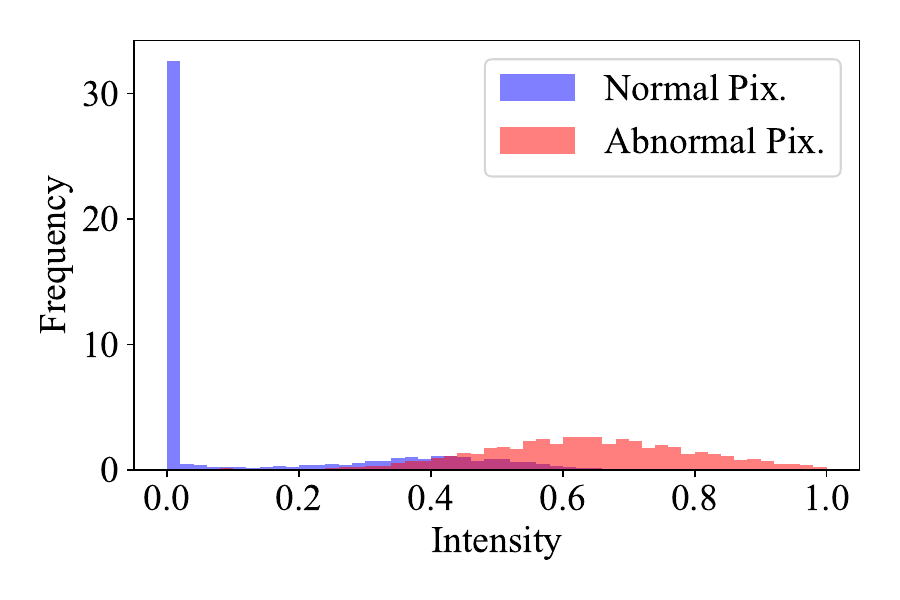}
\caption{Intensity distribution of normal and abnormal pixels in BraTS2021 dataset.}
\label{fig:intensity}
\end{figure}

\subsection{Anomalies with pronounced intensity differences}
\citet{meissen2021challenging} demonstrated that simple intensity thresholding outperforms several deep learning methods on FLAIR MR scans. To further investigate this phenomenon, we analyze the intensity distributions of normal and abnormal pixels in BraTS2021 dataset and incorporate intensity thresholding into our benchmark. Surprisingly, as illustrated in Fig.~\ref{fig:intensity}, normal and abnormal pixels in BraTS2021 dataset exhibit distinguishable intensity distributions. Quantitative results in Table~\ref{tab:ano_seg} reveal that intensity thresholding achieves the second-best segmentation performance (measured by $\lceil$Dice$\rceil$), even surpassing advanced methods based on DDPM. These findings suggest that FLAIR scans effectively highlight most tumor regions through intensity variations, while many sophisticated deep learning methods underperform compared to this simple approach in identifying tumor regions. Consequently, for anomalies with pronounced intensity differences, such as tumors in FLAIR scans, it is crucial to benchmark deep learning-based methods against intensity thresholding to ensure their clinical relevance and utility.

The current analysis of pixel-level AnoSeg remains limited due to the scarcity of annotations and the narrow range of anomaly types, primarily focusing on prominent tumors in brain MRIs. Future work should expand the evaluation to include diverse and subtle anomalies, which could better reveal the potential of deep learning methods and provide deeper insights into the capabilities and limitations of AD methods in real-world clinical scenarios.

\subsection{Latent space configuration of AE}
Section~\ref{subsec:latent} demonstrates that the latent space is the core component in reconstruction-based AD, with a simple adjustment of the latent dimensions proving to be the most effective means of controlling the latent space and achieving satisfactory performance. Unfortunately, existing approaches still rely on testing results to find the optimal configuration, leaving the question of how to adjust the latent space to its optimal state for diverse datasets unresolved. Recently, \citet{cai2024rethinking} leveraged information theory to reveal that the entropy of an optimal latent space should align with that of normal data, but they did not develop a method to explicitly measure this entropy. To tackle this challenge, a promising direction would be to quantify the information entropy of normal training data and develop self-adaptive methods that dynamically constrain the latent space to approach the entropy of normal data on different datasets. This approach would enable the determination of the theoretically optimal solution of AE-based AD methods.

\subsection{Distance function for measuring reconstruction error}
Section~\ref{subsec:distance} highlights the significance of distance functions in measuring reconstruction error for reconstruction-based AD. While some distance functions, such as perceptual loss (AE-PL), exhibit remarkable performance, none surpass the baseline AE-$\ell_2$ on all seven datasets. For instance, the powerful AE-PL performs 6.1\% worse in AUC than AE-$\ell_2$ on ISIC2018, despite showing over 20\% better AUC on datasets like RSNA and Camelyon16. These observations indicate that designing an effective distance function for reconstruction error remains a challenging problem, which has not been adequately investigated in the literature.

Driven by these findings, a fundamental question arises: what should be a good distance function for measuring reconstruction error? To answer this question, we compare different distance functions based on their reconstruction fidelity in Fig.~\ref{fig:visualization} and their metrics in Tables~\ref{tab:ano_cls} and \ref{tab:ano_seg}. From a visual perspective, AE-SSIM generates high-quality reconstructions that preserve more fine-grained details compared to other methods. In contrast, AE-PL produces reconstructions that appear significantly distorted. However, AE-PL outperforms AE-SSIM by a considerable margin in terms of the metrics on six datasets, for both AnoCls and AnoSeg. This suggests that what is a good distance function can be counterintuitive. Therefore, it is preferable to learn a distance function automatically through deep learning rather than relying on manual design.

The concept of learning distance functions (similarity metrics) through deep learning was originally proposed for multimodal medical image registration. Traditional metrics often struggle to measure the similarity between different imaging modalities due to the high variability of tissue appearance. \citet{simonovsky2016deep} addressed this challenge by training a network to measure similarity using constructed image pairs. Analogously, in anomaly detection, certain reconstruction discrepancies may not necessarily correspond to actual anomalies but rather reflect inherent limitations in model capacity. Ideally, such discrepancies should be systematically excluded when computing the distance metric between input and reconstructed images, with emphasis placed solely on highlighting abnormal regions. Drawing inspiration from deep similarity metrics in registration, we believe it is possible to train a similarity network to automatically measure anomaly-related reconstruction errors. This network can be trained on synthesized image pairs generated from the normal training set, taking into account the data characteristics. Pursuing this direction may lead to the development of a robust distance function capable of effectively measuring reconstruction errors across various datasets with different image modalities.

\subsection{Utilization of ImageNet pre-trained weights}
As depicted in Tables~\ref{tab:ano_cls} and \ref{tab:ano_seg}, the SOTA methods among both reconstruction-based and SSL-based methods extensively utilize ImageNet pre-trained weights. While these methods demonstrate competitive performance, the utilization of ImageNet weights takes different forms, including \textbf{distance measurement} (e.g., AE-PL), \textbf{input data transformation} (e.g., feature-reconstruction methods), and \textbf{direct feature extraction} (e.g., two-stage SSL methods). These successful applications highlight the remarkable capability of ImageNet weights in discerning normal and abnormal patterns, providing inspiration to advance AD from various perspectives.

The effective utilization of ImageNet weights in distance measurement suggests that in reconstruction-based AD, it is desirable to measure semantic differences rather than low-level intensity differences in reconstruction errors. This aligns with the findings of \citet{meissen2022pitfalls}, which indicate that residual errors alone cannot reliably identify abnormal regions with moderate pixel intensity due to intrinsic errors in normal regions. The application of ImageNet weights in input data transformation indicates the feasibility of enhancing the saliency of abnormal regions using feature maps. As revealed in Fig.~\ref{fig:examples}, medical abnormal regions often exhibit subtle and seamless characteristics, which may explain the relatively lower performance of AD in medical scenarios compared to industrial scenarios. Enhancing the saliency of these anomalies becomes crucial in such cases. Lastly, the explicit use of ImageNet weights for feature extraction underscores their discriminative capability in AD, motivating further adaptation to improve performance.

While ImageNet weights have been effectively applied in various contexts, it is important to highlight that their application in feature reference-based methods, such as knowledge distillation \citep{zhang2023destseg} and feature modeling \citep{lee2022cfa} approaches, which are particularly popular in the industrial domain, is not suitable for medical AD. The performance of these methods in medical images, as shown in Table~\ref{tab:feat-reference}, is inferior to that of other strategies utilizing ImageNet weights, as indicated in Table~\ref{tab:ano_cls}. This observation prompts us to reconsider how we can enhance these commonly used strategies in industrial AD and adapt them effectively to the medical domain.

\begin{table*}[!t]
\centering
\caption{Performance of SOTA feature reference-based methods.} \label{tab:feat-reference}
\resizebox{\linewidth}{!}{
\begin{tabular}{lcccccccccccccccc}
\toprule
\multirow{2}{*}{Method} & \multicolumn{2}{c}{ RSNA } & \multicolumn{2}{c}{ VinDr-CXR } & \multicolumn{2}{c}{ Brain Tumor } & \multicolumn{2}{c}{ LAG } & \multicolumn{2}{c}{ISIC2018} & \multicolumn{2}{c}{Camelyon16} & \multicolumn{4}{c}{ BraTS2021 }\\
\cmidrule(l){2-3}\cmidrule(l){4-5} \cmidrule(l){6-7} \cmidrule(l){8-9} \cmidrule(l){10-11} \cmidrule(l){12-13} \cmidrule(l){14-17}
                 & AUC & AP & AUC & AP & AUC & AP & AUC & AP & AUC & AP & AUC & AP & AUC & AP & $\text{AP}_\text{pix}$ & $\lceil$Dice$\rceil$ \\ \midrule
DeSTSeg   & 79.6\scriptsize{$\pm$1.3} & 76.8\scriptsize{$\pm$2.9} & 67.7\scriptsize{$\pm$3.3} & 66.7\scriptsize{$\pm$2.6} & 96.5\scriptsize{$\pm$1.5} & 94.7\scriptsize{$\pm$2.7} & 66.2\scriptsize{$\pm$1.9} & 65.2\scriptsize{$\pm$0.5} & 72.7\scriptsize{$\pm$1.8} & 59.3\scriptsize{$\pm$2.3} & 48.0\scriptsize{$\pm$2.0} & 48.3\scriptsize{$\pm$1.7} & 73.9\scriptsize{$\pm$1.9} & 86.2\scriptsize{$\pm$2.0} & 34.5\scriptsize{$\pm$2.0} & 39.8\scriptsize{$\pm$1.1}\\
CFA & 65.1\scriptsize{$\pm$0.7} & 60.9\scriptsize{$\pm$0.5} & 59.4\scriptsize{$\pm$1.1} & 54.7\scriptsize{
    0.6} & 82.6\scriptsize{$\pm$1.9} & 72.1\scriptsize{$\pm$3.2} & 69.2\scriptsize{$\pm$0.6} & 63.2\scriptsize{$\pm$1.1} & 68.6\scriptsize{$\pm$2.0} & 52.4\scriptsize{$\pm$1.8} & 54.2\scriptsize{$\pm$0.8} & 53.2\scriptsize{$\pm$0.9} & 73.5\scriptsize{$\pm$5.2} & 88.5\scriptsize{$\pm$2.0} & 41.4\scriptsize{$\pm$3.7} & 44.3\scriptsize{$\pm$1.3} \\
\bottomrule
\end{tabular}
}
\end{table*}

\begin{table*}[!t]
\centering
\caption{Performance of VLM-based methods \citep{zhou2023anomalyclip} under the zero-shot setting. *The performance of the officially released weights.} \label{tab:vlm}
\resizebox{\linewidth}{!}{
\begin{tabular}{lcccccccccccccccc}
\toprule
\multirow{2}{*}{Method} & \multicolumn{2}{c}{ RSNA } & \multicolumn{2}{c}{ VinDr-CXR } & \multicolumn{2}{c}{ Brain Tumor } & \multicolumn{2}{c}{ LAG } & \multicolumn{2}{c}{ISIC2018} & \multicolumn{2}{c}{Camelyon16} & \multicolumn{4}{c}{ BraTS2021 }\\
\cmidrule(l){2-3}\cmidrule(l){4-5} \cmidrule(l){6-7} \cmidrule(l){8-9} \cmidrule(l){10-11} \cmidrule(l){12-13} \cmidrule(l){14-17}
                 & AUC & AP & AUC & AP & AUC & AP & AUC & AP & AUC & AP & AUC & AP & AUC & AP & $\text{AP}_\text{pix}$ & $\lceil$Dice$\rceil$ \\ \midrule
AnomalyCLIP*     & 66.4 & 61.3 & 51.8 & 53.7 & 95.8 & 95.3 & 70.6 & 68.9 & 58.7 & 48.3 & 59.7 & 59.4 & 84.9 & 93.3 & 60.7 & 57.7 \\
AnomalyCLIP      & 75.2\scriptsize{$\pm$0.4} & 72.1\scriptsize{$\pm$0.6} & 49.9\scriptsize{$\pm$0.1} & 52.3\scriptsize{$\pm$0.1} & 88.4\scriptsize{$\pm$1.3} & 87.5\scriptsize{$\pm$1.4} & 76.0\scriptsize{$\pm$0.3} & 73.5\scriptsize{$\pm$0.3} & 61.9\scriptsize{$\pm$0.9} & 52.9\scriptsize{$\pm$0.6} & 67.4\scriptsize{$\pm$1.0} & 60.0\scriptsize{$\pm$0.8} & 85.3\scriptsize{$\pm$0.1} & 93.3\scriptsize{$\pm$0.1} & 62.2\scriptsize{$\pm$0.1} & 58.8\scriptsize{$\pm$0.2} \\
\bottomrule
\end{tabular}
}
\end{table*}

Moreover, the ImageNet pre-trained weights are definitely sub-optimal due to the gap between natural images and medical images, making fine-tuning on target datasets a promising direction. Although \citet{reiss2021panda} and \citet{reiss2023mean} tried to investigate this point with PANDA and MSC respectively, the experiments in Tables~\ref{tab:ano_cls} and \ref{tab:ano_seg} show that these methods do not achieve satisfactory improvement. Therefore, how to fine-tune the pre-trained features on the target datasets remains an unresolved problem. 

Meanwhile, vision language models (VLMs) \citep{radford2021learning} have witnessed substantial advancements recently. Trained on vast amounts of image-text data, these models exhibit exceptional generalizability and robustness across various downstream tasks. Naturally, exploring the utilization of these powerful models from the aforementioned perspectives is a worthwhile endeavor \citep{sun2024position}.

\subsection{Special settings for anomaly detection}
In addition to the traditional OCC setting that involves only normal training data for AD, recent works have increasingly focused on more challenging and meaningful AD settings that closely resemble real-world scenarios. 


\textbf{One-class semi-supervised setting} performs training using both normal data and unlabeled data, which may contain some anomalies. In this setting, a typical idea is to employ two modules to learn the data distribution: one is trained exclusively on normal data, while the other is trained on both normal and unlabeled data \citep{cai2022dual,cai2023dual,bozorgtabar2023amae}. These methods utilize the discrepancy in reconstruction between the two modules to identify anomalies. Other approaches \citep{siddiquee2024brainomaly,zhang2024spatial} train a GAN-based model to eliminate potential diseases in the unlabeled data, generating healthy reconstruction for both normal and unlabeled inputs. The reconstruction error is then applied as an anomaly score.

It is noteworthy that the presence of various rare diseases poses a challenge in incorporating all types of anomalies into the unlabeled training set. In other words, the testing set may include numerous unseen anomaly types that never appeared in the unlabeled training set. This mismatch has been shown by \citet{oliver2018realistic} to lead to significant performance degradation in traditional semi-supervised learning approaches. 
To address this challenge, various approaches have been proposed in this setting. For instance, \citet{cai2022dual,cai2023dual} employed an additional module to capture the distribution of both normal and unlabeled training data as a complement to the normative module. This module serves as a complement, enhancing the ability to identify seen anomalies present in the unlabeled training data without negative impact on recognizing unseen anomalies. \citet{siddiquee2024brainomaly} utilized the unlabeled training data to facilitate the model in generating pseudo-healthy reconstructions rather than recognizing seen anomalies, thereby enhancing generalization to unseen anomalies.

While methods in this setting exploit readily available unlabeled data to assist in AD, a sufficient number of normal samples are required to train the model. Therefore, exploring techniques to train an effective model with a smaller amount of data would further enhance its applicability in clinical practice.

\textbf{Few-/Zero-shot setting} trains AD methods with little to no samples from the target datasets. This setting has gained significant attention in industrial defect detection \citep{huang2022registration,xie2023pushing,chen2023zero}. Recently, with the rapid advancement of VLMs, these models have been widely adopted for the few-/zero-shot setting \citep{jeong2023winclip,huang2024adapting,zhou2023anomalyclip,li2024musc}. Compared to the one-class and one-class semi-supervised settings, this setting reduces the requirement for a large amount of normal training data, enabling rapid deployment of models in entirely new scenarios. However, the few-/zero-shot setting has not been extensively explored in the medical domain, and existing methods have not exhibited satisfactory performance on medical data, as indicated in Table~\ref{tab:vlm}. Medical images, with more complex variations than natural industrial images, pose additional challenges for developing AD approaches under this setting. Addressing these challenges remains a crucial area for future research.

\subsection{3D anomaly detection}
Similar to most prevailing methods for AD in medical images, our analysis relies on leveraging 2D information. Even in the case of 3D volumetric scans, such as MRIs, the common approach is to extract axial slices and analyze them as 2D images. However, this strategy overlooks crucial information in the axial direction, leading to sub-optimal performance in AD for volumetric medical images.

Several previous works have explored AD on volumetric medical images. One common strategy is to directly adapt 2D methods into the 3D domain by replacing 2D layers in networks with their 3D counterparts. For example, \citet{simarro2020unsupervised} developed a 3D version of f-AnoGAN. \citet{cho2021self} and \citet{marimont2023achieving} adapted the one-stage SSL-based AD methods for volumetric CT scans or MRIs. Recognizing that 3D networks often encounter depth limitations due to memory constraints, \citet{kang2022joint} proposed joint embedding to combine the advantages of both 2D and 3D networks by aligning their representations in the embedding space. Additionally, \citet{naval2021implicit} introduced implicit field learning approaches \citep{park2019deepsdf} to reconstruct voxel intensities based on the latent variables of the volume and the spatial coordinates of the voxels.

While the exploration of 3D AD on medical images presents a new avenue, it remains an area that requires further investigation. The lack of medical datasets that support both 2D and 3D AD poses challenges for comparing the performance of 3D and 2D methods. Moreover, memory limitations in 3D networks restrict their network designs for effective feature extraction. Future research should prioritize addressing these issues and developing solutions to advance the field.

\section{Conclusion}   \label{sec:conclusion}
This paper presents a comprehensive benchmark for medical anomaly detection, incorporating seven datasets and a comparison of thirty typical methods. Our extensive evaluation and analysis reveal several key findings and challenges to guide future research in this field.

Firstly, in the absence of pre-training, reconstruction-based methods demonstrate greater robustness compared to SSL-based methods. Among the reconstruction-based methods, the simplest AE serves as a good baseline, delivering satisfactory performance across various datasets and near-perfect metrics on simple datasets like Hyper-Kvasir and OCT2017. Therefore, we strongly recommend researchers to include AE as a reference in their comparisons.
Moreover, our results highlight the importance of latent space configuration and reconstruction error measurement in reconstruction methods, which display a substantial impact on performance. Regarding latent space configuration, datasets with near OOD (e.g., local anomalies) tend to benefit from very small latent sizes. Among these datasets, the more complex ones exhibit increasing optimal values of latent size. Conversely, datasets with far OOD (e.g., global semantic anomalies) benefit from large latent sizes. In terms of reconstruction error measurement, it is crucial to employ a distance function that captures the anomaly-related semantic difference rather than only low-level intensity difference. This is why perceptual loss outperforms $\ell_2$ loss by a significant margin on most datasets. However, the current strategies for latent space configuration and reconstruction error measurement are still sub-optimal, presenting a promising avenue for further investigation.

Additionally, we observe that ImageNet pre-trained weights exhibit high effectiveness and potency in medical AD. They are successfully employed in various ways, including distance measurement, input data transformation, and direct feature extraction. Fine-tuning these weights on task-specific datasets to enhance performance remains an unresolved challenge. Furthermore, the recent advancements in vision-language models (VLMs) offer new possibilities for leveraging these powerful pre-trained models in AD.

Lastly, we identify several special settings for AD, such as the one-class semi-supervised setting and the zero-/few-shot setting, which closely align with real-world scenarios and warrant further exploration. Particularly, the significant progress in VLMs facilitates the development of zero-/few-shot methods for AD, which should attract more attention in the future.

Overall, our benchmark establishes a foundation for researchers in medical AD, providing insights into the typical methods. We hope that this work will contribute to the development of more effective and robust AD methods, ultimately benefiting the medical domain in recognizing rare diseases and improving health screening.

\section*{Acknowledgments}
This work was partially supported by the Research Grants Council of the Hong Kong Special Administrative Region, China (Project Nos. T45-401/22-N and R6003-22), Shenzhen Science and Technology Innovation Committee Fund (Project No. SGDX20210823103201011), and National Natural Science Foundation of China/HKSAR Research Grants Council Joint Research Scheme under Grant N\_HKUST627/20.

\bibliographystyle{model2-names.bst}\biboptions{authoryear}
\bibliography{refs}

\end{document}